\ifdefined\pdfoutput\pdfoutput=1\fi %
\RequirePackage{fix-cm} %
\documentclass{article}

\usepackage{arxiv}
\usepackage[numbers, compress]{natbib}

\usepackage{standalone}
\usepackage{tikz}
\usetikzlibrary{positioning,arrows.meta,calc,fit,backgrounds,math}
\usepackage{pgfplots}
\pgfplotsset{compat=1.18}
\usepgfplotslibrary{groupplots}

\usepackage{wrapfig}
\usepackage{amsmath,amssymb,amsthm}
\usepackage{float}
\usepackage{multirow}
\usepackage{array} %
\usepackage{graphicx}

\usepackage{iftex}
\ifpdftex
  \usepackage[utf8]{inputenc} %
  \usepackage[T1]{fontenc}    %
\else
  \usepackage{fontspec}
\fi
\usepackage{hyperref}       %
\usepackage{url}            %
\usepackage{booktabs}       %
\usepackage{amsfonts}       %
\usepackage{nicefrac}       %
\usepackage{microtype}      %
\usepackage{xcolor}         %
\usepackage{pgf}
\providecommand{\mathdefault}[1]{#1}
\makeatletter
\newcommand{\UnderscoreCommands}{\do\citep \do\citet \do\NAT@testdef \do\bibliographystyle}
\makeatother
\usepackage[strings]{underscore}
\usepackage[font=small,skip=4pt]{caption} %

\title{\ourNO: Training Neural Operators with Noisy Monte Carlo Estimates for Particle Transport Problems}
\renewcommand{\shorttitle}{\ourNO: Neural Operators from Noisy Monte Carlo Labels}

\usepackage{authblk}

\author[1]{Yubo Cao\thanks{Equal contribution.}}
\author[1]{Xi Deng\protect\footnotemark[1]}
\author[2]{Mengqi Xia}
\author[1,3]{Vignesh Gopakumar}
\author[4]{Ander Gray}
\author[1]{Anima Anandkumar}
\affil[1]{California Institute of Technology}
\affil[2]{Yale University}
\affil[3]{UK Atomic Energy Authority}
\affil[4]{LIX, CNRS, \'Ecole polytechnique}

\graphicspath{{./fig/}}

\usepackage{amssymb}
\usepackage{enumitem}
\usepackage{xspace}

\NewDocumentCommand{\ourNO}{}{PTNO\xspace}
\NewDocumentCommand{\ourNOlong}{}{Particle Transport Neural Operator\xspace}

\newcommand{\Density}{\psi} %
\newcommand{\totCS}{\mathbf{\sigma}}

\newcommand{\Sphere}{\mathbb{S}}
\newcommand{\Source}{Q}
\newcommand{\Dir}{\omega}
\newcommand{\sPos}{\mathbf{x}}
\newcommand{\dif}{\mathrm{d}}
\newcommand{\ScatteringFunction}{\mathbf{f}_s}

\newcommand{\Energy}{E}
\newcommand{\NO}{\mathcal{G}_\theta}
\newcommand{\scene}{\xi}
\newcommand{\sol}{\mathcal{U}}

\newcommand{\Loss}{\mathcal{L}}

\newcommand{\filter}{\chi} %
\newcommand{\MC}{\hat{\sol}}

\newcommand{\param}{a}

\newcommand{\expect}[2]{\mathbb{E}_{#1}[#2]}
\newcommand{\error}{\zeta} %
\newcommand{\SCNO}{\mathcal{B}_\theta}
\newcommand{\LossC}{\Loss_C}
\newcommand{\sg}{\operatorname{sg}}
\newcommand{\softplus}{\operatorname{softplus}}
\newcommand{\loss}{\ensuremath{\mathrm{PRel}L_{2}}\xspace}
\newcommand{\DENO}{\mathcal{S}}

\begin{document}

\maketitle

\begin{abstract}
Particle transport under multiple scattering is central to radiative transfer and plasma physics, yet high-fidelity Monte Carlo (MC) simulations must trace prohibitively many particles.
Learning-based surrogates can amortize this cost, but their training typically requires expensive, well-converged MC solutions.
We propose the \textbf{\ourNOlong (\ourNO)}, a neural operator framework that learns particle transport surrogates directly from noisy, low-cost MC labels.
Such labels pose two challenges: (1) high variance, which destabilizes standard supervised learning, and (2) a high dynamic range (HDR) spanning many orders of magnitude.
For the first, we learn the solution operator from noisy labels of many configurations, amortizing MC cost and generalizing to unseen configurations.
Because MC labels are unbiased, we show that the squared loss on them shares its minimizer with the loss on converged solutions, and our budget-allocation study over training scenes $M$, Monte Carlo samples per render $N$, and independent renders per scene $K$ shows that many noisy scenes beat fewer converged ones.
For the second, a nonlinear transform such as the logarithm biases noisy supervision.
Instead, \ourNO keeps labels in physical space and enforces positivity with a softplus output layer that represents small values effectively.
We further train with a pointwise relative $L_2$ loss (\loss), the stop-gradient relative loss of HDR denoising and neural rendering, which normalizes each residual by the stop-gradient prediction instead of the noisy label.
We demonstrate \ourNO on neutron transport in fusion reactors and radiative transfer in participating media, yielding accurate surrogates in both domains.
On the two neutronics tasks, \ourNO is $10^4$--$10^5{\times}$ faster than converged MC on the same CPU and $10^3$--$10^5{\times}$ cheaper than MC at matched accuracy; on the two radiative-transfer tasks, MC at matched accuracy costs $0.8$--$11{\times}$ as much as \ourNO.
\end{abstract}


\section{Introduction}

Particle transport governs the spatial, angular, and energy distribution of particles as they propagate and interact with matter.
It underpins a wide range of applications: light transport in participating media~\citep{Mitsuba}, neutron transport in shielding and reactor design~\citep{OpenMC, MCNP}, and neutral particle transport in plasma physics~\citep{EIRENE, SOLPS}.
Though these domains differ in their scattering laws, energy dependence, and material modeling, they share a common foundation in the Boltzmann transport equation.

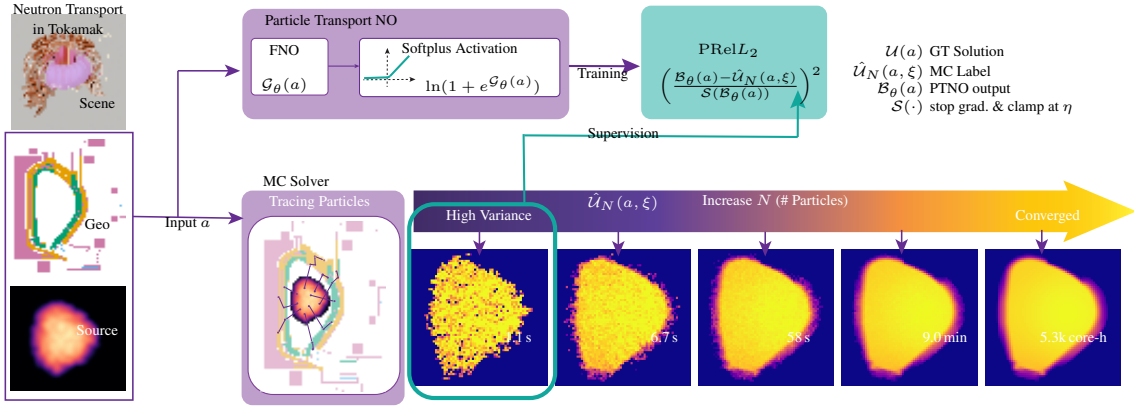
\begin{figure*}[t]
    \centering
    \resizebox{0.95\linewidth}{!}{%
    \begin{tikzpicture}[x=\textwidth,every text node part/.style={align=left}]
    \node at (0, 0){\includegraphics[width=\linewidth,page=1]{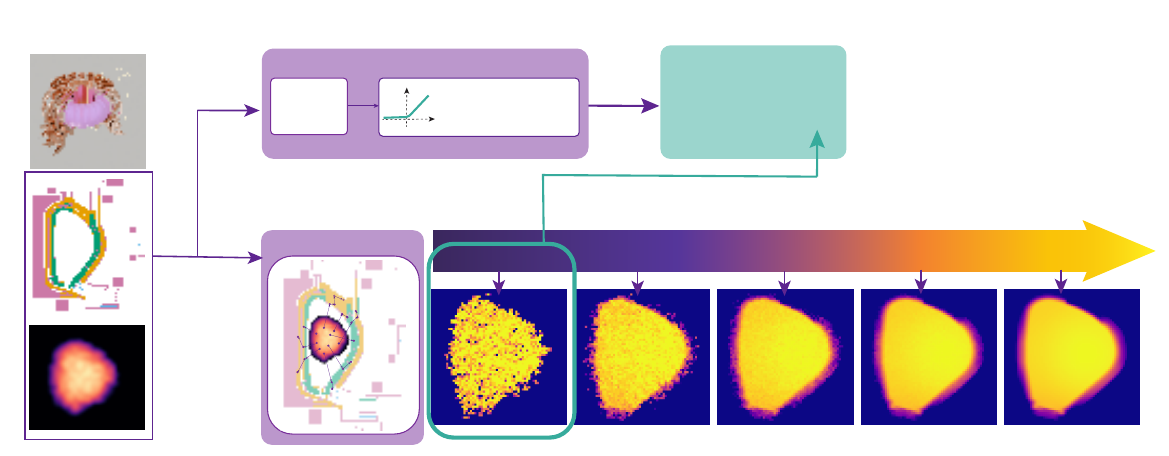}};
\node[anchor=north west, fill=white, fill opacity=0.7, text opacity=1,
      inner sep=3pt, rounded corners=1pt]
  at (0.235, 2.15) {%
  \renewcommand{\arraystretch}{0.65}
  \begin{tabular}{@{}r@{~}l@{}}
    \tiny$\sol(\param)$             & \tiny GT Solution \\
    \tiny$\MC_N(\param,\xi)$        & \tiny MC Label \\
    \tiny$\SCNO(\param)$            & \tiny \ourNO output \\
    \tiny$\DENO(\cdot)$             & \tiny stop grad.\ \& clamp at $\eta$ \\
  \end{tabular}%
};
    \node at (-0.2, 2.35) {\tiny{Particle Transport NO}};
\node at (-0.24, 2.0-0.05) {\tiny{FNO}};
    \node at (-0.09, 2.0-0.05) {\tiny{Softplus Activation}};
    \node at (-.07, 1.45) {\tiny{$\ln(1 + e^{\NO(\param)})$}};
    \node at (0.05, 0.75) {\tiny{Supervision}};
    \node at (0.0325, 1.6) {\tiny{Training}};
    \node at (0.14, 2.0-0.05) {\tiny{\loss}};
    \node at (0.15, 1.45) {\tiny{$\bigg(\frac{\SCNO(\param) - \MC_N(\param,\xi)}{\DENO(\SCNO(\param))}\bigg)^{2}$}};
    \node at (-0.24, 1.45) {\tiny{$\NO(\param)$}};
    \node at (-0.3225, -0.5) {\textcolor{black}{\tiny{Input $a$}}};
    \node at (-0.23, 0.10) {\textcolor{black}{\tiny{MC Solver}}};
    \node at (-0.21, -0.2) {\textcolor{white}{\tiny{Tracing Particles}}};
    \node at (-0.065, -0.4) {\textcolor{white}{\tiny{High Variance}}};
   \node at (0.05, -0.19) {\textcolor{white}{\tiny{$\MC_N(a,\xi)$}}};
   \node at (0.18, -0.19) {\textcolor{white}{\tiny{Increase $N$ (\# Particles)}}};
   \node at (0.413, -0.4) {\textcolor{white}{\tiny{Converged}}};
   \node at (-0.04, -2.1) {\textcolor{white}{\tiny{1.1\,s}}};
   \node at (-0.04+0.125, -2.1) {\textcolor{white}{\tiny{6.7\,s}}};
   \node at (-0.04+0.24, -2.1) {\textcolor{white}{\tiny{58\,s}}};
   \node at (-0.04+0.365, -2.1) {\textcolor{white}{\tiny{9.0\,min}}};
   \node at (-0.04+0.475, -2.1) {\textcolor{white}{\tiny{5.3k\,core-h}}};
    \node at (-0.4, -2.0) {\textcolor{white}{\tiny{Source}}};
    \node at (-0.4, -0.5) {\textcolor{black}{\tiny{Geo}}};
    \node at (-0.4, 1.2) {\textcolor{black}{\tiny{Scene}}};
    \node at (-0.425, 2.5) {\textcolor{black}{\tiny{Neutron Transport}}};
    \node at (-0.425, 2.25) {\textcolor{black}{\tiny{in Tokamak}}};
    \end{tikzpicture}}
    \caption{\textbf{\ourNO learns from extremely high-variance MC labels and, at inference, predicts the well-converged solution.} The \ourNOlong is trained on MC supervision (left green box, columns 1--2) across varying configurations of the input $\param$.
    Bottom row shows the EU-DEMO 1/16 wedge neutron-transport task: columns 1--4 are analog OpenMC runs (no variance reduction) of one held-out configuration (the scene of \autoref{fig:intro-demo}) with $N{=}10^{3},10^{4},10^{5},10^{6}$ histories, i.e., simulated neutron trajectories, on one 24-thread Xeon Platinum 8352Y CPU, labeled with their measured transport wall-clock (library initialization, $35$--$41$\,s, excluded).
    Column 5 is the converged reference of the same configuration, computed with weight windows (a variance reduction that splits or randomly terminates particles to keep their statistical weights within a target band in each region) at a cost of $5{,}280$ core-hours.
    \ourNO keeps positivity and a wide dynamic range with a softplus output activation, and our \loss loss keeps MC labels in physical space.}
    \vspace{-10pt}
    \label{fig:illustration}
\end{figure*}

Monte Carlo (MC) methods remain the standard approach for solving particle transport problems in complex geometries and high-dimensional state spaces, due to their generality and physical fidelity.
Rather than solving the Boltzmann equation directly, MC methods simulate particle propagation by sampling from probability distributions over possible transport events—free flight, scattering, absorption, and boundary interactions—and estimate the solution as an expectation over many particle histories, the simulated trajectories of individual particles.
We distinguish between \textbf{low-variance MC estimates}, obtained with many particles and therefore close to the converged solution, and \textbf{high-variance MC estimates}, obtained with few particles and therefore much noisier but significantly cheaper.

Obtaining a smooth, low-noise solution requires tracing a large number of particles, prohibitively costly when repeated solves are needed across varying geometries, material configurations, or boundary conditions, as in inverse design, optimization, and multiphysics coupling.
Recent learning-based approaches reduce this cost via learned sampling strategies~\citep{muller2019neural, NeuralPGuiding}, learned caching and variance reduction~\citep{woscache, radiancecaching}, or denoising of high-variance MC estimates~\citep{ResidualDenoising, KernelDenoise, MCDenoising}.
However, these methods either retain MC simulation at inference time or operate as post-processing modules, so prediction still depends on an external solver.
This motivates a different goal: a standalone surrogate that directly maps a transport environment to its solution field.

Neural operators~\citep{li2021fno} provide a natural framework by learning mappings from input functions—geometry, material coefficients, sources, boundary conditions—to output solution fields, enabling fast inference across varying environments without running a solver.
However, training typically requires near-converged MC labels, making dataset construction expensive.
Recent work has shown this requirement can be relaxed: Walk-on-Spheres Neural Operator (WoSNO)~\citep{viswanath2026operatorlearningusingweak} trains neural operators from noisy walk-on-spheres estimates for elliptic PDEs, demonstrating operator learning under weak stochastic supervision.
This result is limited to elliptic PDEs and does not address the high-dimensional, multiple-scattering transport problems we consider here.

Following a similar philosophy, our \textbf{\ourNOlong (\ourNO)} trains neural operators directly from \textbf{high-variance MC estimates}.
The key insight is that unbiased stochastic estimates provide sufficient supervision to learn the underlying operator across a distribution of transport scenarios, without requiring a converged solution for each input.

An additional challenge is that particle transport solutions are positive and often span many orders of magnitude (about $10$ to $17$ in our 3D fluence and neutronics datasets), making direct regression poorly conditioned.
A common remedy is to normalize errors by the target magnitude—e.g., through the reciprocal of the supervision label.
However, when supervision is a noisy MC estimate, such nonlinear operations introduce bias, since nonlinear functions do not commute with expectation.
We show that existing bias-correction strategies~\citep{misso2022unbiased} do not fully resolve this issue in our training dynamics.

Rather than correcting transformed noisy targets, \ourNO avoids transforming the MC labels altogether.
It applies a softplus output parameterization to enforce positivity and evaluates errors in the original value space.
For the loss, the standard relative $L_2$—which places the noisy label in the denominator—leads to instability; instead we normalize by the \emph{predicted} value at each point.
This stop-gradient relative $L_2$, which we call the \textbf{pointwise relative $L_2$ loss} (\loss), is established for training on noisy HDR targets in image denoising and neural rendering~\citep{lehtinen2018noise2noise, muller2020ncv, radiancecaching, mildenhall2022rawnerf, tinits2025nonlinear}; to our knowledge it has not been used to train neural operators from noisy MC transport labels, where it provides a stable objective for high-variance supervision.

Under a fixed simulation budget, this creates a fundamental allocation tradeoff: spending many particles on few configurations yields low-variance labels, while spending fewer particles per configuration covers many more training environments.
In our experiments, a converged MC label costs up to $10^5\times$ more than the noisy labels we use.
Cheaper labels, therefore, allow broader coverage of geometries and material configurations, relying on the unbiasedness of MC estimates and operator learning across environments to recover the underlying solution map.

Our framework is domain-agnostic: we formulate radiative transfer and neutron transport as instances of a unified particle transport problem, and show that a single neural-operator formulation can be applied across both domains despite differences in their physical parameters and interaction models.
In summary, we make the following contributions:

\begin{itemize}[leftmargin=10pt]
\item \ourNO, a neural-operator framework that learns particle-transport surrogates directly from high-variance, unbiased MC labels, applicable to both radiative transfer and neutron transport under multiple scattering.
\item A justification and a budget study for noisy supervision: unbiased labels leave the population minimizer of the squared loss unchanged, and a controlled allocation study over the number of training scenes, Monte Carlo samples per render, and independent renders per scene shows that, until scene coverage saturates, many noisy scenes outperform fewer converged ones, in line with walk-on-spheres operator learning and Monte Carlo surrogate models~\citep{viswanath2026operatorlearningusingweak, pratt2026moment}.
What the allocation buys depends on the loss: from $10^5$ to $10^6$ far-field scenes, \ourNO gains $0.14$ in $\log_{10}$ SSIM while $L_2$ and log MSE gain $0.02$ and $0.01$, respectively.
\item An analysis of the Jensen bias introduced by nonlinear transformations of noisy MC labels, and a recipe that avoids it for neural operators on radiative and neutron transport: a softplus output head that enforces positivity in the original physical space, the physical-space $L_2$ residual that anchors the minimizer to the label mean, and the stop-gradient relative loss \loss of prior denoising and rendering work~\citep{lehtinen2018noise2noise, muller2020ncv, radiancecaching, mildenhall2022rawnerf, tinits2025nonlinear}, together with an analysis of why the combination is stable.
\item Demonstrations of prediction on held-out geometries and material configurations, achieving $3.4$--$8.2\%$ $\log_{10}$ relative error on the radiative-transfer benchmarks and $4.5$--$8.4\%$ on the two tokamak neutron-transport benchmarks (with the historical rows qualified in Appendix~\ref{app:eval-protocol}), while providing $10^4$--$10^5\times$ faster prediction than converged MC on the same hardware for neutron transport.
\end{itemize}

\begin{figure*}[t]
    \centering
  \begingroup%
  \catcode`\_=8\relax%
  \edef\pgfincludename{\detokenize{fig_intro_demo}}%
  \graphicspath{{fig/\pgfincludename/}{fig/}}%
  \setkeys{Gin}{draft=false}%
  \resizebox{\linewidth}{!}{\input{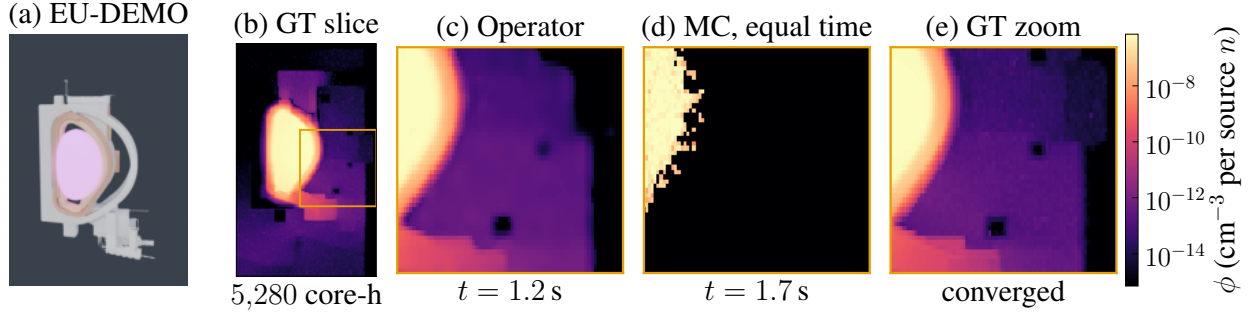}}%
  \endgroup%

\caption{
  \textbf{\ourNO predicts the EU-DEMO neutron flux in about a second, closer to the converged reference than Monte Carlo run for the same time.}
  One held-out source configuration of the EU-DEMO 1/16 wedge.
  \emph{(a)}~Rendering of the DEMO tokamak; the wireframe marks the inspected poloidal cross-section~\cite{DEMO_diagnostics}.
  \emph{(b)}~Converged reference flux on the full slice (OpenMC with weight windows at $700\times365\times730$, block-averaged to the $140\times73\times146$ training mesh; $5{,}280$ core-hours for this scene, median $6{,}130$ over the $24$ held-out scenes); the orange box marks the region zoomed in panels~(c)--(e).
  \emph{(c)}~Neural operator (NO) prediction on the zoomed region ($1.2$\,s on a 24-thread Xeon Platinum 8352Y CPU).
  \emph{(d)}~Analog Monte Carlo (MC, no variance reduction) matched to that wall-clock time: $N{=}2\times10^{3}$ histories on the same CPU ($1.7$\,s of transport), still visibly noisy.
  \emph{(e)}~Converged reference on the zoomed region, for direct comparison with (c)--(d).
  \ourNO uses about $6\times10^5\times$ less CPU time than the converged reference on the training mesh ($1.2\times10^4\times$ on the reference grid, \autoref{par:exp-demo}), and its $\log_{10}$ relative $L_2$ is $3.2\times$ lower than that of MC at the same wall-clock time (per-seed median over the $24$ held-out scenes; $3.18\pm0.02$, mean $\pm$ std over five seeds, range $3.16$--$3.20\times$), scored only on cells where the reference is measured (Appendix~\ref{app:eval-protocol}).
}
    \label{fig:intro-demo}
\end{figure*}
\section{Related Work}
Learning-based methods for radiative and neutron transport fall into two categories: replacing specific components of the transport pipeline with neural counterparts, or learning end-to-end surrogates that predict the final physical quantity.

\paragraph{Modular neural components.} This approach integrates neural networks into traditional Monte Carlo (MC) pipelines to accelerate convergence or reduce variance without discarding the underlying physical framework.
\begin{itemize}[leftmargin=10pt]
\item \textit{Computer Graphics:} In rendering, neural networks are widely used for denoising, mapping noisy renders with few samples per pixel (SPP) to clean images by leveraging geometric features like normals and albedo as auxiliary inputs~\cite{kalantari2015machine, KernelDenoise, chen2024temporally}.
Beyond post-processing, path-guiding reduces variance by learning incident radiance distributions~\cite{vorba2014line, herholz2016product, muller2017practical, muller2019neural}.
Neural radiance caching replaces expensive recursive path tracing with learned regression models that approximate the radiance field for real-time global illumination~\cite{radiancecaching}.
Training such models on noisy HDR targets commonly uses a relative $L_2$ whose normalizer is the stop-gradient prediction, in Noise2Noise denoising~\cite{lehtinen2018noise2noise}, neural control variates~\cite{muller2020ncv}, radiance caching~\cite{radiancecaching}, and RawNeRF~\cite{mildenhall2022rawnerf}; \citet{tinits2025nonlinear} analyze which nonlinear target transforms keep Noise2Noise training nearly unbiased.
\loss is this loss; we apply it to neural operators on transport problems.

\item \textit{Other Transport Domains:} In nuclear engineering, MC neutron-transport simulations can be inefficient because few particle histories reach a detector or a deeply shielded region, so the tallies there, the accumulated particle scores that estimate the flux, converge slowly.
Classical variance reduction relies on ``importance'' maps such as weight windows, which split particles entering important regions and randomly terminate particles in unimportant ones so that statistical weights stay within a target band; CADIS/ADVANTG automate their construction via deterministic adjoint transport~\cite{culbertson1999assessment, mosher2013advantg}.
Recent work explores neural alternatives for source biasing, flux prediction, or surrogate transport models~\cite{martinez2024neural, moloko2023prediction, kobayashi2024deep}.
\end{itemize}

\paragraph{End-to-end surrogates and neural operators.} An alternative line of work replaces the entire transport solver with a learned proxy.
Earlier efforts used point-wise surrogates such as Physics-Informed Neural Networks (PINNs), enforcing the transport equation at discrete coordinates, but they generalize poorly across geometries and source terms~\cite{prantikos2023physics}.
Neural operators such as Fourier Neural Operators (FNO) instead learn mappings between function spaces, enabling generalization across environments~\cite{li2023fourier}.
Surrogates have been applied to plasma modeling~\cite{FNO_Plasma_2024} and Tritium Breeding Ratio estimation in tokamaks and stellarators~\cite{Mánek_2023,TBR_stellerator}, but not yet for predicting full neutron-flux fields.
Existing neural operators for physics typically assume access to high-quality simulation data or rely on PDE residuals that are hard to optimize for the stiff, high-dynamic-range equations of particle transport.
We target this gap by training neural operators directly from low-sample MC estimates.
WoS-NO trains neural operators for elliptic PDEs from walk-on-spheres estimates~\cite{viswanath2026operatorlearningusingweak}, and \citet{pratt2026moment} show for Monte Carlo surrogates of Markov-chain moments that mean estimates benefit most from parameter-space coverage, while nonlinear targets such as covariances also need replication to control bias; our allocation study adds how this trade-off interacts with the loss.

\paragraph{Bias correction.} Bias enters MC pipelines through many channels.
Classical debiasing typically targets biased estimators arising from finite approximations (step size, kernel radius, path-length truncation) and removes them by constructing sequences of increasingly accurate estimators.
Randomized debiasing and multilevel Monte Carlo can convert convergent biased approximations into unbiased ones, but remain expensive~\cite{mcleish2010general,rhee2015unbiased,blanchet2015unbiased}.
In computer graphics, \citet{misso2022unbiased} unify this view via Taylor- and telescoping-series corrections for transmittance, photon mapping, and finite-difference derivatives.
Our setting differs: the supervision targets are unbiased in the original physical scale, but a nonlinear log transformation introduces Jensen bias.
Series-expansion corrections require sufficiently many samples for the estimator to be near the true value, which is impractical here.
We instead keep the noisy target in unbiased linear form while using a positive model output head and prediction-normalized residual for stability under high dynamic range.

\section{Background}
\label{sec:background}
We formulate particle transport as an operator-learning problem with noisy Monte Carlo supervision.
For each input configuration $a$, including geometry, material properties, and source parameters, there is an unknown transport response $\sol(a)$ that we would like to predict.
Monte Carlo solvers provide unbiased but noisy estimates $\MC_N(a,\xi)$ of $\sol(a)$ from $N$ particles, and our goal is to train a neural operator directly from the high-variance estimates.
Appendix~\ref{app:notation} lists the notation.

Particle transport describes the evolution of particles undergoing scattering and absorption within a medium. %
The solution is expressed as an integral over particle trajectories.
Let $\xi \in \Omega$ denote a particle path consisting of a sequence of free-flight segments and scattering events, and let $\Omega = \cup_{k\ge1} \Omega_k$ be the space of all paths grouped by the number of scattering events $k$.
The measurable response can be written as
\begin{align}
    \sol(a) = \int_{\Omega} f(\xi, a) \dif \nu(\xi),
\end{align}
where $\nu$ is the underlying path-space measure induced by the transport process.
The contribution function $f(\xi, \param)$ encodes the accumulated weight of a particle trajectory $\xi$, including emission at the source, attenuation along free-flight segments, and scattering events along the path (see \autoref{app:pt} for details).
This formulation is equivalent to the integral form of the Boltzmann equation.

MC methods approximate this integral by sampling $N$ paths $\xi_i \sim p(\cdot \mid a)$ from a proposal distribution $p$ and collecting them in $\xi=(\xi_1,\ldots,\xi_N)$.
The estimator is given by
\begin{align}
    \MC_N(a, \xi) = \frac{1}{N}\sum_{i=1}^{N} \frac{f(\xi_i, a)}{p(\xi_i \mid a)}.
\end{align}
This estimator is unbiased but often exhibits high variance, particularly in multiple-scattering regimes where the path length $k$ can be large, the contribution function $f$ is only implicitly defined (i.e., not available in closed form), and it is difficult to construct a sampling distribution $p$ proportional to $f$.
As a result, obtaining low-noise estimates of $\sol(\param)$ requires a large number of particle paths, making data generation expensive.

\subsection{Neural Operator}

Neural operators learn mappings between function spaces and provide a natural framework for amortizing the solution of parametric PDEs and integral equations.
For particle transport, we view the transport environment as an input function or configuration $a$, which may include spatially varying material coefficients, boundary conditions, source parameters, and geometry.
The corresponding output is the measurable response or solution field $\sol(a)$ defined by the transport equation.
The goal is to learn the solution operator
\begin{align}
    \NO: a \mapsto \sol(a),
\end{align}
from training examples drawn from a distribution of environments.
In contrast to standard neural networks that approximate a finite-dimensional input-output map tied to a particular discretization, neural operators are designed to approximate the underlying continuous operator and can be evaluated across different resolutions or discretizations.
This property is especially useful for particle transport, where one often needs to solve the same physical model repeatedly across many geometries, material fields, or source conditions.
After training, the neural operator can predict $\sol(a)$ for a new configuration $a$ at a fraction of the cost of a full MC solve.
\section{Method}
\label{sec:method}
We formulate particle transport surrogate learning as operator learning from unbiased but high-variance stochastic simulation labels.
For each input configuration $a \sim \mathcal{P}_a$, including geometry, material properties, and source parameters, let $\sol(a)$ denote the converged transport response.
An MC solver provides a noisy estimate
\begin{equation}
    \MC_N(a,\xi) = \sol(a) + \error_N(a,\xi), \label{eq:MC}
\end{equation}
where $\xi$ denotes the random particle histories used by the estimator.
We assume that the MC estimator is unbiased in the original physical space:
\begin{equation}
    \expect{\xi}{\MC_N(a,\xi) \mid a} = \sol(a),
    \qquad
    \expect{\xi}{\error_N(a,\xi) \mid a} = 0. \label{eq:MCError}
\end{equation}
Our goal is to learn a neural operator that maps $a$ to $\sol(a)$ without requiring well-converged MC labels for every training configuration.

\subsection{Learning from Unbiased Stochastic Labels}
The ideal supervised objective assumes access to the converged solution:
\begin{equation}
    \LossC(\NO)
    =
    \expect{a}{
        \|\NO(a)-\sol(a)\|_2^2
    }.
\end{equation}
In practice, $\sol(a)$ is unavailable, and we instead observe noisy MC labels $\MC_N(a,\xi)$.
Training with these labels yields the surrogate objective
\begin{align}
    \Loss_N(\NO) = \expect{a,\xi}{\| \NO(a) - \MC_N(a, \xi)\|^2}.
\end{align}
Substituting the decomposition of $\MC_N(\param, \xi)$ from \autoref{eq:MC} gives 
\begin{align*}
    \Loss_N(\NO) &= \expect{a,\xi}{\|\NO(a) - \sol(a) - \error_N(a, \xi)\|^2}\\
    &= \LossC(\NO) + \expect{a,\xi}{\|\error_N(a, \xi)\|^2} - 2\expect{a}{\langle\NO(a) - \sol(a), \expect{\xi}{\error_N(a, \xi)\mid a}\rangle}. 
\end{align*}

The cross term vanishes by the unbiasedness of $\MC_N$ (\autoref{eq:MCError}), and the variance term $\expect{a, \xi}{\|\error_N(a, \xi)\|_2^2}$ is independent of $\NO$.
Hence the two objectives share the same population minimizers:
\begin{align}
    \arg \min_{\NO} \Loss_N(\NO) = \arg \min_{\NO} \LossC(\NO).
\end{align}
Unbiased Monte Carlo labels therefore provide statistically valid supervision in the original linear physical space, even when individual labels have high variance.
The variance term inflates the loss value but leaves the optimum unchanged.

\subsection{Jensen Bias from Nonlinear MC-Label Transforms}
Although the squared-loss argument justifies training from noisy labels in the linear physical space, particle-transport solutions often span many orders of magnitude (about $10$ to $17$ in our 3D fluence and neutronics references, \autoref{tab:dynamic-range}).
Direct regression in value space can therefore be poorly conditioned: High-flux regions dominate the loss, while low-flux regions may be ignored.
A natural remedy is the pointwise relative error, which normalizes each residual by the local field value before squaring and averaging.
With access to the clean solution, the corresponding objective is
\begin{align*}
    \Loss_{\mathrm{rel}}(\NO) = \expect{a}{\| (\NO(a) - \sol(a)) / (\sol(a)+\eta)\|^2}.
\end{align*}
If we instead substitute a noisy MC label inside the reciprocal, the objective becomes
\begin{align}
    \Loss_{N,\mathrm{rel}}(\NO) = \expect{a,\xi}{\|(\NO(a) - \MC_N(a,\xi)) / (\MC_N(a, \xi)+\eta) \|^2}. \label{eq:biased-rel}
\end{align}
This objective is biased because the reciprocal does not commute with the expectation over MC noise:
\begin{align}
   \expect{\xi}{\frac{1}{\MC_N(a,\xi) + \eta}\mid\param} \neq \frac{1}{\sol(a)+\eta} = \frac{1}{\expect{\xi}{\MC_N(a,\xi)\mid\param}+\eta}.\label{eq:bias}
\end{align}
Applying a nonlinear transformation to noisy labels, therefore, changes the expected training objective and shifts its minimizer away from $\sol$.
We refer to this effect as \emph{Jensen bias}.
This motivates a key design choice in our method: keep MC labels in the original physical space and avoid passing them through nonlinear transformations such as logarithms or reciprocal normalizers.

\subsection{\texorpdfstring{\ourNOlong}{Particle Transport Neural Operator}}
To handle high dynamic range (HDR) without transforming noisy labels, we modify the surrogate architecture rather than the target.
Let $\NO(\param)$ be a neural operator backbone (e.g., an FNO) that produces a latent field.
We map this latent field to the physical solution space through a softplus activation:
\begin{equation}
    \SCNO(a)
    =
    \softplus(\NO(a))
    =
    \ln(1 + e^{\NO(a)}).
\end{equation}
From here on, $\NO$ denotes this latent backbone and $\SCNO$ the physical prediction of \ourNO.
 For large negative latent values, the softplus behaves approximately exponentially, allowing negative activations to represent very small positive physical values.
 For large positive values, it behaves approximately linearly, avoiding the severe gradient growth of a pure exponential parameterization.
 This makes softplus a useful positive parameterization for high-dynamic-range nonnegative fields.
 Crucially, the prediction $\SCNO(a)$ is compared directly against the noisy MC label in the original physical scale.

\paragraph{Pointwise relative $L_2$.}
We avoid applying a nonlinear reciprocal to the MC label by using the prediction as the local normalization scale.
Specifically, we define
$
\DENO(x)=\max(\sg(x),\eta),
$
where $\sg(\cdot)$ denotes stop-gradient and $\eta>0$ prevents division by very small values.
\begin{align}
    \Loss_{N,\mathrm{PRel}}(\NO) = \expect{a, \xi}{\| (\SCNO(a) - \MC_N(a, \xi)) / \DENO(\SCNO(a))\|^2}.
\end{align}
The denominator provides a local scale estimate, but it is treated as constant during backpropagation.
Therefore, gradients are taken through the linear residual in the numerator, and the MC label enters the loss only in the original physical space.
While the exact minimizer-equivalence argument applies to the unweighted squared loss, this objective preserves the key requirement that noisy MC labels are not passed through a nonlinear transformation.
Appendix~\ref{app:prel2-analysis} shows that, in expectation over MC noise, its gradient has no stationary point other than $\SCNO(a)=\sol(a)$ wherever the output Jacobian has full row rank.
The formula is not new: the same stop-gradient relative $L_2$ trains HDR denoisers on noisy targets~\citep{lehtinen2018noise2noise, tinits2025nonlinear}, neural control variates and radiance caches~\citep{muller2020ncv, radiancecaching}, and NeRF on noisy raw images~\citep{mildenhall2022rawnerf}.
What we add is its use for neural-operator learning from noisy MC transport labels, paired with the softplus head, and the stationary-point analysis above for that combination.

\paragraph{Fixed points.}
For a fixed input $a$, let $B_\theta=\SCNO(a)$ denote the physical prediction, $U=\expect{\xi}{\MC_N(a,\xi)\mid a}$ the mean of the noisy label, $D(B)=\operatorname{diag}(\max(B_i,\eta)^2)$ the diagonal matrix of squared normalizers, and $J=\partial B_\theta/\partial\theta$ the output Jacobian.
Because the pointwise denominator is detached, the expected parameter gradient of the per-sample loss $\ell$, up to a positive constant, is
\begin{align}
   \mathbb{E}_\xi[\nabla_\theta\ell\mid a]
=J^\top D(B_\theta)^{-1}(B_\theta-U). 
\end{align}
Thus, a stationary point satisfies $J^\top D(B_\theta)^{-1}(B_\theta-U)=0$.
Since $D(B_\theta)$ is positive definite, the normalization introduces no zero directions.
Under the standard local assumption that $J$ has full row rank ($JJ^\top\succ0$), $J^\top$ is injective and stationarity forces $B_\theta=U$.
\loss therefore introduces no additional biased fixed point in output space.
A rank-deficient Jacobian can create parameter-space stationary points, but that limitation is shared by ordinary squared loss.

\paragraph{Optimization dynamics.} The stop-gradient update is the gradient of the output-space potential
\begin{align}
  \Phi(B_\theta)
=\sum_i\int_{U_i}^{B_{\theta,i}}
\frac{s-U_i}{\max(s,\eta)^2}\,\dif s,  
\end{align}
because $\nabla_B\Phi=D(B_\theta)^{-1}(B_\theta-U)$.
Under continuous-time gradient descent, $\dot\theta=-J^\top\nabla_B\Phi$, so
\begin{align}
    \frac{\dif\Phi}{\dif t}
=-\nabla_B\Phi^\top JJ^\top\nabla_B\Phi
\leq0.
\end{align}
The expected dynamics therefore decrease $\Phi$ monotonically.
The prediction-dependent denominator rescales the per-voxel update but cannot reverse its direction or create another output-space minimum.
Under $JJ^\top\succ0$, the dynamics stop only at $B_\theta=U$; when $U_i=0$ exactly, softplus approaches this solution as a boundary limit rather than attaining it at a finite latent value.
This is a population-gradient argument; finite steps, stochastic-gradient variance, and rank-deficient parameterizations remain practical considerations.

\paragraph{Sample allocation.}
In finite training we draw $M$ training scenes and, for each scene, $K$ independent renders of $N$ Monte Carlo samples each.
The empirical objective is
\begin{align}
    \widehat{\mathcal L}_{M,K,N}(\theta)
    =
    \frac{1}{MK}
    \sum_{i=1}^{M}
    \sum_{j=1}^{K}
    \| (\SCNO(a_i) - \MC_N(a_i,\xi_{i,j})) / \DENO(\SCNO(a_i))\|^2,
\end{align}
where $a_i$ is the $i$-th scene and $\xi_{i,j}$ the samples of its $j$-th render.
$M$ controls coverage of the scene distribution, while $N$ and $K$ control the noise of the supervision.
Under a fixed particle budget $N_{\mathrm{tot}}=MKN$, reducing $N$ frees samples for more scenes $M$ or more renders $K$; \autoref{sub:exp2} studies this trade-off empirically.
Two exact facts and one standard bound organize the trade-off (Appendix~\ref{app:finite-sample}).
First, because the normalizer is detached, for fixed normalizers the $K$ renders of a scene enter the objective and its gradient only through their mean, whose noise variance is $\sigma^2/(NK)$ for single-sample variance $\sigma^2$; a full pass over $K$ renders of $N$ samples is therefore equivalent to one render of $NK$ samples.
Second, for the unweighted squared loss with unbiased labels, the standard least-squares bound for a realizable class $\mathcal F$ of effective dimension $d_{\mathrm{eff}}$ gives
\begin{align}
\mathbb{E}\,\mathcal{R}(\hat{\theta})-\inf_{\theta\in\Theta}\mathcal{R}(\theta)
\lesssim
\mathcal{E}_M(\mathcal{F})
+\frac{\sigma^2 d_{\mathrm{eff}}}{MNK},
\label{eq:alloc-bound}
\end{align}
where $\ell$ is the per-sample loss, $\mathcal R(\theta)=\mathbb E[\ell(\theta)]$ its population risk, $\hat\theta$ the minimizer of $\widehat{\mathcal L}_{M,K,N}$ over the parameter set $\Theta$, and $\mathcal E_M(\mathcal F)$ the excess risk the same class would reach from $M$ converged labels.
Scene coverage enters through $\mathcal E_M$ and the $1/M$ of the noise term, label noise only through $1/(MNK)$, and there is no term in $N$ or $K$ alone.
Third, a nonlinear label transform such as the logarithm moves the population minimizer from $\log_{10}\sol$ to $\mathbb E[\log_{10}\MC_N]=\log_{10}\sol+O(1/N)$, or $O(1/(NK))$ when the $K$ renders are averaged before the transform; this adds a risk floor of order $1/N^2$ (respectively $1/(NK)^2$) that no number of scenes removes.
For \loss, Appendix~\ref{app:prel2-analysis} shows that the expected gradient $\mathbb{E}_\xi[\nabla_\theta\ell\mid a]=J^\top D(B_\theta)^{-1}(B_\theta-U)$ has no output-space stationary point other than the label mean, so this Jensen shift does not arise; the bound \eqref{eq:alloc-bound} itself is proved only for the unweighted squared loss.

\newcommand{\E}{\mathbb{E}}
\newcommand{\Var}{\operatorname{Var}}

\section{Experiments}
\label{sec:experiments}

We evaluate \ourNO on four datasets spanning two physical regimes of particle transport---photon transport and neutron transport---comparing against the NO baseline under matched architectures and varying loss formulations.
The \textbf{far-field radiance task} models light propagating through a heterogeneous participating medium bounded by a ball (with spatially varying extinction and albedo defined in spherical coordinates) and predicts the far-field radiance; this is a canonical problem in atmospheric and volumetric rendering, where multiple scattering renders analytic solutions intractable.
Its references span only about $3$ orders of magnitude (\autoref{tab:dynamic-range}), so it is not a high-dynamic-range case; because converged references are affordable for thousands of its scenes, we use it as a controlled testbed for label noise and particle-budget allocation.
The \textbf{3D fluence task} extends this setting and defines the spatially varying material parameters and emission functions in 3D Cartesian coordinates, predicting the full 3D fluence field, whose references span about $10$ orders of magnitude.
The \textbf{spherical-tokamak neutron transport task} moves to nuclear fusion: a parametric tokamak geometry with a varying D--T plasma source produces $14.1$\,MeV fusion neutrons that scatter and attenuate through shield, first-wall, and blanket materials.
The resulting neutron flux distribution~\cite{shimwell2021paramak} is the key quantity for estimating material activation (radioactivity induced in structures by neutron capture), nuclear heating, and tritium breeding.
Finally, the \textbf{EU-DEMO task} targets a reactor-scale engineering model of the European demonstration fusion reactor~\cite{DEMO_diagnostics,lu2020serpent_mcnp_demo}, where predicting the neutron flux across the full poloidal cross-section under varying plasma-source conditions is essential for shielding design, diagnostic placement, and blanket performance assessment.
Full dataset details are in Appendix~\ref{sec:data}.

Our experiments show that (1) \ourNO outperforms alternatives and (2) high-variance MC labels are more sample-efficient than converged references.
The main text reports $\log_{10}$ relative $L_2$ and $\log_{10}$ SSIM, both computed on the output field itself (the angular map or the 3D volume) after $\log_{10}(\max(\cdot, \epsilon))$ with a per-dataset floor $\epsilon$, so that accuracy counts across the entire dynamic range; appendix tables add linear-space \%~RMSE, PSNR, and SSIM, and Appendix~\ref{app:eval-protocol} defines every metric.
All validation and test metrics are computed on unseen input configurations held out from training.
Unless noted otherwise, entries are the mean $\pm$ sample standard deviation over five training seeds, scored against converged MC references.
Far-field radiance and EU-DEMO now use the final checkpoint of a fixed training schedule; the 3D fluence and spherical-tokamak rows remain historical checkpoint-selected results pending replacement (Appendix~\ref{app:eval-protocol}).
$(M,N,K)$ denotes the number of training scenes, the Monte Carlo samples per render, and the independent renders per scene (\autoref{sec:method}); on the radiative-transfer tasks $N$ is counted in samples per pixel or per voxel (SPP), and on the neutron-transport tasks in histories, a render being one OpenMC run.

\subsection{Jensen Bias and Loss Comparison}
\label{sec:bias-variance}

Passing noisy MC labels through a nonlinear map moves the minimizer of the training objective away from the physical solution; this is the Jensen bias of \autoref{sec:method}, illustrated in \autoref{fig:jensen}.

We study this effect extensively in Appendix~\ref{app:exp2-debias}.
On the far-field radiance dataset, \autoref{fig:inverse-mc-bias} shows that training with the biased objective of \autoref{eq:biased-rel} collapses the prediction by several orders of magnitude.
\ourNO avoids this failure mode by normalizing with a stop-gradient prediction rather than the noisy MC label.

In \autoref{tab:inverse-mc-v2} (Appendix~\ref{app:inverse-mc}), we evaluate the effect of Jensen bias on training with the far-field radiance dataset.
We find that bias correction does not fully close the performance gap, whereas \ourNO trained with \loss achieves stable performance by avoiding nonlinear transformations of noisy MC labels, and therefore avoiding Jensen bias.
Appendix~\ref{app:denominator-ablation} isolates the mechanism with a full denominator~$\times$~activation ablation on the 3D fluence dataset and a matching $2{\times}2$ recipe ablation on EU-DEMO: every variant whose residual scale is the noisy MC label, or the prediction with gradients flowing through it, either collapses or inflates its output scale, while only the stop-gradient pointwise residual paired with the softplus head remains stable.

\subsection{Particle-Budget Allocation and Noise Stability}
\label{sub:exp2}

\autoref{tab:exp2} varies the Monte Carlo samples per render, $N\in\{4,64,128\}$, for three losses on the far-field radiance testbed and reports $\log_{10}$~SSIM.
$L_2$ changes least with $N$ but stays weak in dim regions (\autoref{tab:exp1} in Appendix~\ref{app:exp2-debias}); log MSE degrades sharply at lower $N$, consistent with a Jensen bias that grows with label noise; \ourNO + \loss keeps $0.96\times$ of its $N{=}128$ score at $N{=}4$.

\begin{table}[!htb]
\caption{\textbf{\ourNO + \loss loses less accuracy than log MSE as labels get noisier.}
Far-field radiance with $M{=}10^6$ training scenes and $N$ samples per pixel per label; all other settings shared.
Entries are $\log_{10}$~SSIM on the $100$ held-out scenes against the converged reference, mean $\pm$ std over five seeds (higher is better); parenthesized ratios compare each lower-$N$ column against $N{=}128$.}
\label{tab:exp2}
\centering
\footnotesize
\begin{tabular*}{\textwidth}{@{\extracolsep{\fill}}llccc@{}}
    \toprule
    Arch. & Loss & $N{=}128$ $\uparrow$ & $N{=}64$  (vs.\ $N{=}128$) $\uparrow$ & $N{=}4$ (vs.\ $N{=}128$) $\uparrow$ \\
    \midrule
    NO      & $L_2$       & $0.660\pm0.026$ & $0.662\pm0.022$ ($1.00{\times}$) & $0.652\pm0.020$ ($0.99{\times}$) \\ %
    NO      & log MSE      & $0.853\pm0.003$ & $0.812\pm0.002$ ($0.95{\times}$) & $0.470\pm0.001$ ($0.55{\times}$) \\ %
    \ourNO & \loss (ours)  & $\mathbf{0.952\pm0.002}$ & $\mathbf{0.949\pm0.002}$ ($\mathbf{1.00{\times}}$) & $\mathbf{0.917\pm0.004}$ ($\mathbf{0.96{\times}}$) \\ %
    \bottomrule
\end{tabular*}
\end{table}

\begin{figure}[htb]
    \centering
    \resizebox{\linewidth}{!}{\input{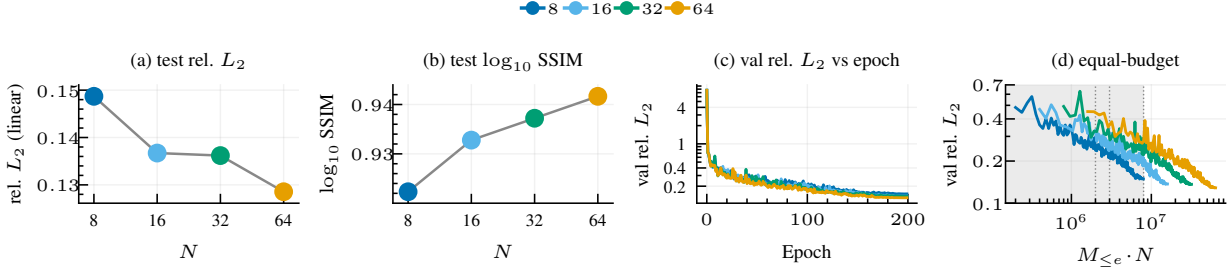}}
    \caption{\textbf{Under \ourNO + \loss, more scenes outperform more particles per scene.}
    Far-field radiance; all runs use the same loss with $M{=}10^6$ and a 200-epoch cosine schedule, one seed per $N$, scored on the $100$ held-out scenes against the earlier $10^5$-SPP reference (Appendix~\ref{subsec:data-spherical}).
    (a,b)~End-of-training test linear rel.~$L_2$ and $\log_{10}$~SSIM vs.\ training $N$ at fixed $M{=}10^6$, i.e.\ with a budget that grows with $N$.
    (c)~Per-epoch validation linear rel.~$L_2$ (log scale).
    (d)~Per-epoch validation against cumulative budget $M_{\le e}{\cdot}N$, where $M_{\le e}$ is the number of distinct scenes seen by epoch $e$ and reaches $M{=}10^6$ at the final epoch; the shaded band marks the budgets at which every run has data (${\le}8{\times}10^6$ particles), and gray vertical lines mark equal-budget slices.
    At equal budget, lower-$N$ runs win.}
    \vspace{-1em}
    \label{fig:exp-spp-sweep}
\end{figure}

At equal particle budget (\autoref{fig:exp-spp-sweep}), we fix the objective to \ourNO + \loss and sweep $N\in\{8,16,32,64\}$ on the same $M{=}10^6$ training set.
For the same total number of simulated particles, lower-$N$ runs reach lower error, so the particle budget is better spent on covering more scenes than on reducing the noise of each label; this sweep has one seed per $N$, and the multi-seed allocation study of Appendix~\ref{app:controlled-ablation} finds the same trend until scene coverage saturates, as reported for other operator and surrogate models trained on Monte Carlo labels~\citep{viswanath2026operatorlearningusingweak, pratt2026moment}.

\subsection{Multi-Domain and 3D Validation}
\label{sec:cross-domain}

\begin{figure*}[t]
    \centering
    \includegraphics[width=\linewidth]{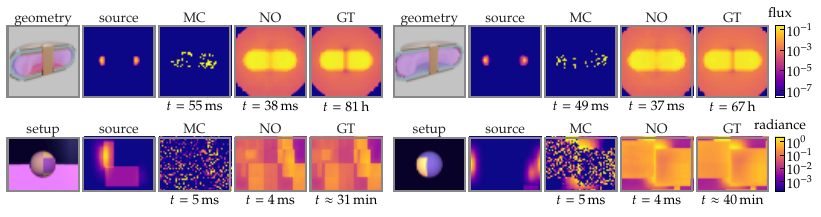}
\caption{
    \textbf{At matched wall-clock time, MC is still dominated by noise while the NO prediction is close to the converged reference.} Each block shows five panels left-to-right: geometry, source distribution, low-budget MC, NO prediction, and converged-MC ground truth (GT), for the two held-out configurations closest to the median per-scene $\log_{10}$~SSIM of the seed-42 model of \autoref{tab:main-results}.
    All timings are measured on one node with an NVIDIA H200 GPU and a 24-thread AMD EPYC 9554 CPU; the MC budget is the smallest whose wall-clock comes closest to one NO forward pass.
    \emph{Top---Spherical tokamak neutron flux:} NO inference takes $37$--$38$\,ms on the GPU; OpenMC on 24 CPU threads needs $49$--$55$\,ms for $N{=}37$--$43$ histories, the smallest budgets it runs near that time.
    The GT pools $10^9$ histories from $100$ independent $10^7$-history runs ($67$--$81$\,h of summed wall-clock, each run on $4$ CPU cores, that is $270$--$320$ core-hours).
    \emph{Bottom---Spherical-slab far-field radiance $L$:} NO inference takes $3.6$\,ms and a Mitsuba render with one sample per pixel $4.8$\,ms on the same GPU; the GT accumulated $10^{7}$\,SPP in $31$--$40$\,min on one H200.
}
    \label{fig:paper-setup-compare}
\end{figure*}

We replicate the loss comparison on the 3D fluence dataset, on spherical-tokamak neutron transport, and on the reactor-scale EU-DEMO benchmark, confirming the findings transfer across physics domains and unseen inputs.
Table~\ref{tab:main-results} reports five-seed results on all four datasets.

\paragraph{3D fluence (historical results).}
We train on a $64^3$ fluence dataset (Appendix~\ref{subsec:data-spatial}) at $(M,N,K){=}(3{\times}10^5,4,2)$; the reference fluence of a scene spans about $10$ orders of magnitude.
A 3D~\ourNO (modes $20^3$, width $64$, $6$ layers) is trained for $120$ epochs ($1.5{\times}10^5$ optimizer updates) and validated on 100 unseen held-out scenes.
Linear-output $L_2$ fails in dim regions, and naive log-target MSE trails \ourNO + \loss by $0.054$ in $\log_{10}$~SSIM, consistent with Jensen bias.
\ourNO + \loss achieves the best performance in both log-space metrics.

\paragraph{Spherical-tokamak neutron transport (historical results).}
We train on $(M,N,K){=}(4{\times}10^5,10^5,1)$, drawn without repetition from a pool of $1{,}024{,}000$ configurations of a parametric D--T spherical-tokamak distribution varying both geometry and plasma source~\citep{shimwell2021paramak}; the reference flux of a scene spans about $10$ orders of magnitude per energy group.
A 3D~NO (modes $24^3$, width $32$, $4$ layers) is trained for 40 epochs ($10^5$ optimizer updates) at batch~4, with validation on 50 unseen configurations against a $10^9$-history reference.
Naive log-target MSE reaches only $\log_{10}$~SSIM $0.560$, while \ourNO + \loss retains $\log_{10}$~SSIM $0.863$ and $\log_{10}$~rel.~$L_2$ $0.045$ on the unseen validation split.

\paragraph{EU-DEMO.}
\label{par:exp-demo}

Building on the EU-DEMO accuracy reported in Table~\ref{tab:main-results}, we now quantify the practical speed--accuracy tradeoff against the converged MC reference.
We train on $(M,N,K){=}(10^5,5{\times}10^5,1)$ on a $140{\times}73{\times}146$ mesh and validate on 24 unseen configurations against references computed with FW-CADIS and MAGIC weight windows (median $6{,}130$ CPU core-hours per configuration).
Weight windows are a variance reduction that splits or randomly terminates particles to keep their statistical weights within a target band in each region; FW-CADIS derives them from an approximate deterministic adjoint solution, and MAGIC refines them over successive MC runs.
On one 24-thread CPU, the converged reference costs $1.2{\times}10^4\times$ the operator's forward pass on the reference grid.
With an FW-CADIS weight window, MC needs $1.6{\times}10^3\times$ the operator's cost, window generation included, to match its $\log_{10}$~rel.~$L_2$, and on none of the 24 configurations does it reach the operator's $\log_{10}$~SSIM within $10^7$ histories.
Appendix~\ref{app:cost} reports the same-hardware and matched-accuracy comparison for all four datasets.

\begin{table}[!htb]
\caption{\textbf{\ourNO achieves superior performance across all four datasets.} Each block compares loss formulations under matched architectures, data, and optimizer-update budgets; bolded values are the best per dataset.
EU-DEMO replaces the linear-$L_2$ loss with relative $L_2$ because the flux peak sits near the Adam $\epsilon$ scale.
Mean $\pm$ standard deviation over five seeds on held-out configurations against converged references ($100$ far-field scenes, $100$ 3D fluence scenes, $50$ spherical-tokamak configurations, $24$ EU-DEMO configurations).
Far-field and EU-DEMO use final checkpoints; the 3D fluence and spherical-tokamak blocks are provisional historical results selected on their evaluation configurations.
EU-DEMO is scored only on cells where the converged reference is measured ($45.4\%$ of the mesh; Appendix~\ref{app:eval-protocol}).}
\label{tab:main-results}
\centering
\scriptsize
\renewcommand{\arraystretch}{0.85}
\begin{tabular*}{\textwidth}{@{\extracolsep{\fill}}llcc@{}}
    \toprule
    Dataset & Method & $\log_{10}$~rel.~$L_2$ $\downarrow$ & $\log_{10}$~SSIM $\uparrow$ \\
    \midrule
    \multirow{3}{*}{Far-field radiance}
      & NO + $L_2$            & $0.4782\pm0.0408$ & $0.6524\pm0.0196$ \\ %
    & NO + log MSE          & $1.3994\pm0.0013$ & $0.4699\pm0.0007$ \\ %
    & \ourNO + \loss (ours) & $\mathbf{0.0825\pm0.0022}$ & $\mathbf{0.9165\pm0.0040}$ \\ %
    \midrule
    \multirow{3}{*}{3D fluence}
      & NO + $L_2$            & $0.2824\pm0.0006$ & $0.7243\pm0.0004$ \\ %
    & NO + log MSE          & $0.0843\pm0.0045$ & $0.8977\pm0.0042$ \\ %
    & \ourNO + \loss (ours) & $\mathbf{0.0338\pm0.0005}$ & $\mathbf{0.9518\pm0.0011}$ \\ %
    \midrule
    \multirow{3}{*}{Sph.~tokamak}
      & NO + $L_2$            & $0.4062\pm0.0058$ & $0.2021\pm0.0227$ \\ %
    & NO + log MSE          & $0.3364\pm0.0027$ & $0.5604\pm0.0286$ \\ %
    & \ourNO + \loss (ours) & $\mathbf{0.0453\pm0.0011}$ & $\mathbf{0.8627\pm0.0024}$ \\ %
    \midrule
    \multirow{3}{*}{EU-DEMO}
      & NO + rel.~$L_2$       & $0.2310\pm0.0050$ & $0.1455\pm0.0066$ \\ %
    & NO + log MSE          & $0.2388\pm0.0002$ & $0.5846\pm0.0001$ \\ %
    & \ourNO + \loss (ours) & $\mathbf{0.0839\pm0.0006}$ & $\mathbf{0.7923\pm0.0016}$ \\ %
    \bottomrule
\end{tabular*}
\end{table}

\section{Discussion}

\ourNO offers several advantages for particle-transport surrogate learning.
It learns directly from high-variance MC labels, avoiding the need to generate expensive converged labels for every training configuration.
Rather than treating MC noise as a preprocessing problem, it leverages noisy but unbiased estimates as stochastic supervision: high-variance labels provide an unbiased estimate of the converged solution up to an additive noise term independent of the model.
To handle high dynamic range, \ourNO avoids Jensen bias by applying a softplus activation to the model prediction and using the prediction, rather than the noisy MC estimate, as the denominator in the pointwise relative loss.
This keeps MC targets in the original physical scale.
Under a fixed simulation budget, this regime is especially valuable for multiple-scattering transport, where low-sample MC labels enable broader coverage of input configurations than fully converged labels.

The method has important limitations.
\ourNO assumes an unbiased MC estimator in linear physical space; bias from path truncation, approximate physics, or a biased Russian roulette (random termination of low-weight particles without the compensating weight increase) weakens the risk-equivalence argument underpinning the training scheme.
Effectiveness is also constrained by the capacity of the underlying neural operator: while the approach generalizes well across in-distribution variations, far out-of-distribution configurations can lead to failure modes not covered by the training distribution.
Our analysis of \loss concerns the expected gradient under continuous-time, full-batch dynamics with a full-row-rank output Jacobian, and does not establish convergence for Adam or finite stochastic steps (Appendix~\ref{app:prel2-analysis}).
All evaluations are simulation-based and test interpolation within parameterized families; generalization to measured data or to new geometry families is not established (Appendix~\ref{app:scope}).
Safety-critical applications would still require validation against high-fidelity MC references before the surrogate could be trusted as a replacement rather than an accelerator.

Future work includes applying \ourNO to full-resolution tokamak models, extending it to neutral transport in plasma-edge simulation, and integrating it into coupled multiphysics, inverse design, and differentiable optimization pipelines.

\newpage

\section*{Acknowledgments}
Anima Anandkumar is supported in part by Bren endowed chair, ONR (MURI grant N00014-23-1-2654), and the AI2050 senior fellow program at Schmidt Sciences.
Ander Gray is partially funded by the Academic and Research Chaire ``Architecture des Systèmes Complexes'' Dassault Aviation, Naval Group, Dassault Systèmes, KNDS France, Agence de l'Innovation de Défense, Institut Polytechnique de Paris.
The Fusion Futures Programme partially funds Vignesh Gopakumar.
As announced by the UK Government in October 2023, Fusion Futures aims to provide holistic support for the development of the fusion sector.
The authors thank Modal for providing part of the compute credits.

\bibliographystyle{unsrtnat}
\bibliography{ref}

\newpage
\appendix

\section{Notation}
\label{app:notation}

\autoref{tab:notation} lists the symbols used across the paper, each with one meaning; symbols local to a single table (the sampling boxes of \autoref{tab:sptok-sampling} and \autoref{tab:demo-sampling}) are defined there.

\begin{table}[!htb]
\caption{\textbf{Each symbol keeps one meaning throughout the paper.} Symbols grouped by where they first appear.}
\label{tab:notation}
\centering
\scriptsize
\setlength{\tabcolsep}{3pt}
\renewcommand{\arraystretch}{1.1}
\begin{tabular}{@{}>{\raggedright\arraybackslash}p{0.2\textwidth}>{\raggedright\arraybackslash}p{0.56\textwidth}>{\raggedright\arraybackslash}p{0.19\textwidth}@{}}
    \toprule
    Symbol & Meaning & Defined in \\
    \midrule
    \multicolumn{3}{@{}l}{\textit{Problem and Monte Carlo labels}} \\
    $a$, $\mathcal{P}_a$ & input configuration (geometry, materials, source) and its distribution & Secs.~\ref{sec:background}, \ref{sec:method} \\
    $\sol(a)$ & converged transport response & \autoref{sec:background} \\
    $\xi\in\Omega$, $\nu$, $k$ & particle path, path space, path-space measure, number of scattering events & \autoref{sec:background} \\
    $f$, $p$ & path contribution function and proposal distribution & \autoref{sec:background} \\
    $\MC_N(a,\xi)$, $\error_N(a,\xi)$ & $N$-particle MC estimate from the paths $\xi=(\xi_1,\ldots,\xi_N)$ and its zero-mean noise & Secs.~\ref{sec:background}, \ref{sec:method} \\
    $M$, $N$, $K$ & training scenes, Monte Carlo samples per render (histories or SPP), independent renders per scene & \autoref{sec:method} \\
    $N_{\mathrm{tot}}$ & particle budget $MKN$ & \autoref{sec:method} \\
    \midrule
    \multicolumn{3}{@{}l}{\textit{Model and losses}} \\
    $\theta$, $\Theta$, $\hat\theta$ & network parameters, parameter set, empirical-risk minimizer & \autoref{sec:method} \\
    $\NO$ & neural operator; from \autoref{sec:method} on, the latent backbone & \autoref{sec:method} \\
    $\SCNO=\softplus(\NO)$, $B_\theta$ & physical prediction of \ourNO; $B_\theta=\SCNO(a)$ at a fixed input & \autoref{sec:method} \\
    $\Loss_C$, $\Loss_N$, $\Loss_{\mathrm{rel}}$, $\Loss_{N,\mathrm{rel}}$, $\Loss_{N,\mathrm{PRel}}$ & population objectives: clean, noisy, clean relative, noisy relative, \loss & \autoref{sec:method} \\
    $\widehat{\mathcal L}_{M,K,N}$, $\ell$, $\mathcal R$ & empirical objective, per-sample loss, population risk & \autoref{sec:method} \\
    $\sg$, $\DENO(x)=\max(\sg(x),\eta)$ & stop-gradient and the detached \loss normalizer & \autoref{sec:method} \\
    $\eta$ & denominator floor of \loss and of the relative losses & \autoref{sec:method} \\
    $U$, $J$, $D(B)$, $\Phi$ & label mean, output Jacobian, squared-normalizer matrix, output-space potential & \autoref{sec:method} \\
    $\mathcal F$, $d_{\mathrm{eff}}$, $\mathcal E_M(\mathcal F)$ & hypothesis class, its effective dimension, its excess risk from $M$ converged labels & \autoref{sec:method} \\
    $\sigma$ & single-sample standard deviation of a Monte Carlo label & \autoref{sec:method} \\
    \midrule
    \multicolumn{3}{@{}l}{\textit{Evaluation}} \\
    $\epsilon$ & floor of the $\log_{10}$ metrics, per dataset & App.~\ref{app:eval-protocol} \\
    $\phi$, $\hat\phi$ & reference and predicted field (flux or radiance) & App.~\ref{app:eval-protocol} \\
    $u$, $\hat u$, $\mathcal C$, $w_i$ & scored field ($\phi$ or $\log_{10}\max(\phi,\epsilon)$), scored cells, cell weights & App.~\ref{app:eval-protocol} \\
    $R$, $\mu$, $\sigma$, $C_1$, $C_2$ & PSNR/SSIM data range, SSIM local mean and standard deviation, SSIM constants & App.~\ref{app:eval-protocol} \\
    \midrule
    \multicolumn{3}{@{}l}{\textit{Transport and datasets}} \\
    $\Density(\sPos,\Dir,\Energy)$ & angular flux at position $\sPos$, direction $\Dir$, energy $\Energy$ & App.~\ref{app:pt} \\
    $\totCS_t$, $\totCS_s$, $\totCS_a$, $\ScatteringFunction$ & total, scattering, absorption cross sections (extinction); scattering function & App.~\ref{app:pt} \\
    $\Source$, $\filter$, $\tau$ & source, detector response, transmittance & App.~\ref{app:pt} \\
    $c$, $g$ & single-scattering albedo, Henyey--Greenstein asymmetry & App.~\ref{sec:data} \\
    $(\vartheta,\varphi)$ & polar and azimuthal angle of the far-field sphere & App.~\ref{subsec:data-spherical} \\
    $R_0$, $a_{\mathrm{p}}$, $\kappa$, $\delta$ & plasma major radius, minor radius, elongation, triangularity & App.~\ref{subsec:data-sptok} \\
    $n$, $c_{\mathrm{m}}$, $H$, $h$ & refractive index, matte albedo, cube half-width, voxel width & App.~\ref{app:refract} \\
    \midrule
    \multicolumn{3}{@{}l}{\textit{Debiasing estimators}} \\
    $Y_i$, $\bar Y_K$, $\hat\mu$, $r_j$ & repeated renders of one pixel, their mean, expansion point, relative residual & App.~\ref{app:exp2-debias} \\
    $A_i$, $A'_j$, $w(x)$, $\widehat W$, $q_m$ & anchor and correction renders, inverse-square weight and its estimate, continuation probability & App.~\ref{app:inverse-mc} \\
    \bottomrule
\end{tabular}
\end{table}

\section{Particle Transport}
\label{app:pt}
Particle transport describes the evolution of particles undergoing scattering and absorption within a medium.
The spatial, angular, and energy distribution of particles is governed by the steady-state linear Boltzmann transport equation.
Let $\Density(\sPos, \Dir, \Energy)$ denote the angular particle density at position $\sPos$, direction $\Dir \in \Sphere$, and energy $\Energy$.
The transport equation reads

\begin{align*}
    \Dir \cdot \nabla_\sPos \Density(\sPos, \Dir, \Energy) + \totCS_t(\sPos, \Energy) \Density(\sPos, \Dir, \Energy) &= \Source(\sPos, 
     \Dir, \Energy) \\ &+ \int_{0}^{\infty} \int_\Sphere \totCS_s\ScatteringFunction(\sPos, \Energy'\rightarrow\Energy, \Dir'\rightarrow\Dir)\Density(\sPos, \Dir', \Energy') \dif \Dir' \dif \Energy'
\end{align*}
Here $\Source$ is the source term, $\totCS_t$ is the total cross section (the probability per unit path length that a particle interacts), $\totCS_s$ is the scattering cross section (the same probability for scattering alone), and $\ScatteringFunction$ is the scattering function; all three are properties of the material the particle is interacting with.
Applications usually require a lower-dimensional response, such as detector flux, surface current, or image radiance, rather than the full angular density $\Density$.
We denote this response by $\sol(\param)$ and express it as a linear functional of $\Density$.
We write
\begin{align}
    \sol(\param) = \int \dif \Energy \int \dif \Dir \int   \dif \sPos \,\, \filter(\sPos,\Dir,\Energy) \Density(\sPos, \Dir, \Energy)
\end{align}
where $\filter(\sPos,\Dir,\Energy)$ is a test function encoding the measurement process (e.g., detector response, surface current, or volume flux), and $\param$ denotes the input configuration, including geometry, material properties, and source parameters.

The same response can be viewed as an expectation over simulated particle trajectories, each contributing to image radiance or a neutron-flux tally.
The path-space integral below expresses this average over trajectories.
Let $\xi \in \Omega$ denote a particle path consisting of a sequence of free-flight segments and scattering events, and let $\Omega = \cup_{k\ge1} \Omega_k$ be the space of all paths grouped by the number of scattering events $k$.
The measurable response can be written as
\begin{align}
    \sol(a) = \int_{\Omega} f(\xi, a) \dif \nu(\xi),
\end{align}
where $\nu$ is the underlying path-space measure induced by the transport process.
The contribution function $f(\xi, \param)$ encodes the accumulated weight of a particle trajectory $\xi$, including emission at the source, attenuation along free-flight segments, and scattering events along the path.
More concretely, for a path $\xi = (\sPos_0 \to \sPos_1 \to \cdots \to \sPos_k)$ with $k$ scattering events, $f$ takes the form of a product of source, transmittance, and scattering terms derived from the transport equation:
\begin{align}
f(\xi, \param)
\propto
\Source(\sPos_0\rightarrow\sPos_1)
[\prod_{j=0}^{k-1}
\tau(\sPos_j \rightarrow \sPos_{j+1})]\prod_{j=1}^{k-1}\left[
\, \totCS_s(\sPos_{j+1})
\, \ScatteringFunction(\sPos_{j-1} \rightarrow \sPos_j \rightarrow \sPos_{j+1})
\,
\right]
\, \filter(\sPos_k),
\end{align}
where $\tau$ denotes the transmittance along each segment, which depends on the total cross section $\totCS_t$, and $\filter$ encodes the measurement response.
This formulation is equivalent to the integral form of the Boltzmann equation.

Monte Carlo methods approximate this integral by sampling $N$ paths $\xi_i \sim p(\cdot \mid a)$ from a proposal distribution $p$ and collecting them in $\xi=(\xi_1,\ldots,\xi_N)$.
The estimator is given by
\begin{align}
    \MC_N(a, \xi) = \frac{1}{N}\sum_{i=1}^{N} \frac{f(\xi_i, a)}{p(\xi_i \mid a)}.
\end{align}
With a proposal that covers all contributing paths and the corresponding importance weights $f/p$, this estimator is unbiased.
Its variance can be high in multiple-scattering or shielding regimes because only a small fraction of sampled histories contribute appreciably to the response.
Sampling paths in proportion to their contribution would reduce this variance, but constructing such a distribution requires knowing which complete trajectories contribute most before tracing them.
As a result, obtaining low-noise estimates of $\sol(\param)$ requires a large number of samples, making data generation expensive.
\section{Loss Definitions and Engineering Stabilizers}
\label{app:losses}

\paragraph{Output head.}
The \ourNO output head $\SCNO(a)=\softplus(\NO(a))$ is the default everywhere in the main paper; we use $\softplus(s)=\ln(1+e^{s})$, which is smooth, monotone, and nonnegative, asymptotically linear for $s\!\gg\!0$ and decays smoothly to zero for $s\!\ll\!0$.
Its derivative $\partial_s\softplus(s)=1/(1+e^{-s})\le1$ already bounds the per-voxel gradient.

\paragraph{Floors.}
We use small positive floors in two places: a stop-gradient denominator floor $\eta$ in the training residual (\autoref{sec:method}), and a log-metric floor $\epsilon$ when reporting $\log_{10}$ metrics.
The paper reports $\log_{10}(\max(\cdot,\epsilon))$ metrics with the task floors in Table~\ref{tab:loss-stabilizers} (all metric definitions are in Appendix~\ref{app:eval-protocol}); for EU-DEMO, the post-hoc checkpoint evaluation floors both prediction and reference at $\epsilon=10^{-17}$.

\paragraph{Where the linear error lives on 3D fluence.}
The linear rel.~$L_2$ of the 3D fluence models is dominated by the voxels around the point emitters, not by the emitter voxel alone.
Re-scoring the five \ourNO seeds of \autoref{tab:main-results} on the $100$ held-out scenes with the emitter voxels excluded ($2.0$ voxels per scene on average) removes only $35\%$ of the squared error (mean over scenes; $30\%$ median), whereas excluding the $3{\times}3{\times}3$ core around each emitter ($54$ voxels, $0.02\%$ of the grid) removes $95.0\%$ and lowers the mean rel.~$L_2$ from $2.11\pm1.41$ to $0.270\pm0.104$ (median over scenes $0.301\pm0.030$ to $0.182\pm0.018$).
The emitter voxel itself is a grid convention with no continuum limit (its cell average scales as $h^{-2}$ with the voxel width $h$), but its neighbors are converged field values, so the remaining near-source error is a genuine model error; the mean is inflated by a handful of scenes per seed with rel.~$L_2$ above one, and the log-space metrics, which are unaffected by this concentration, are the ones we rank by.

\paragraph{\loss normalizer.}
For \ourNO targets the \loss normalizes the residual by the prediction itself, with the gradient on the denominator detached:
\begin{equation}
    r_i(\param,\scene) \;=\; \frac{[\SCNO(\param)]_i - \MC_{N,i}(\param,\scene)}{\sg\!\big(\max([\SCNO(\param)]_i,\,\eta)\big)},
    \label{eq:rel-resid}
\end{equation}
where $\sg(\cdot)$ blocks gradient flow through the normalizer.
The denominator floor caps the per-voxel gradient magnitude when $[\SCNO]_i$ underestimates the target by many orders of magnitude.

\paragraph{Per-experiment instantiations.}
Table~\ref{tab:loss-stabilizers} lists the reported log-metric floor $\epsilon$ used per experiment.
All entries use the softplus head and the \loss residual of Equation~\eqref{eq:rel-resid}; no model-side clamp or robust-loss transition is active.

\begin{table}[!htb]
\caption{\textbf{Each dataset has its own $\log_{10}$ floor, and every experiment uses the softplus head with \loss.} $\epsilon$ is the lower bound used when reporting $\log_{10}$ metrics; EU-DEMO numbers are post-hoc checkpoint evaluations with prediction and reference both floored at $10^{-17}$.
All experiments use the softplus head $\SCNO(\param)=\softplus(\NO(\param))$ with no latent clamp.}
\label{tab:loss-stabilizers}
\centering
\footnotesize
\setlength{\tabcolsep}{3pt}
\begin{tabular*}{\textwidth}{@{\extracolsep{\fill}}lll@{}}
    \toprule
    Experiment & Reported floor $\epsilon$ & Loss \\
    \midrule
    Far-field radiance (Sec.~\ref{sec:bias-variance}) & $10^{-6}$  & \ourNO{}\,$+$\, \loss \\
    3D fluence                                         & $10^{-10}$ & \ourNO{}\,$+$\, \loss \\
    Spherical-tokamak                                 & $10^{-10}$ & \ourNO{}\,$+$\, \loss \\
    EU-DEMO 1/16 wedge                                & $10^{-17}$ & \ourNO{}\,$+$\, \loss \\
    \bottomrule
\end{tabular*}
\end{table}

The floor never binds on far-field radiance, whose references stay above $10^{-6}$; on 3D fluence, the spherical tokamak, and EU-DEMO it removes the deepest $1.6$, $2.0$, and $6.1$ orders of magnitude of the mean reference span (\autoref{tab:dynamic-range}).
On EU-DEMO, voxels outside the transport domain and voxels the reference never reaches sit exactly on the $10^{-17}$ floor.
\autoref{app:prel2-analysis} analyzes the stationary points of the \loss residual; the EU-DEMO scores in \autoref{tab:main-results} exclude these voxels and use only the cells the reference measured.

\paragraph{Reference engines.}
Far-field radiance references use Mitsuba~3~\citep{Mitsuba}; 3D fluence references use the same forward MC estimator as its labels, cross-checked against OpenMC; neutron transport uses OpenMC~\citep{OpenMC}.
Each held-out set is independently rendered to convergence; architecture and schedule details are stated in each subsection.

\section{Denominator and Output-Head Ablations}
\label{app:denominator-ablation}

The \loss residual of Equation~\eqref{eq:rel-resid} makes two coupled design choices: the softplus output head and the \emph{detached-prediction} denominator, i.e., the prediction with its gradient stopped.
We isolate both with matched-compute ablations on the 3D fluence and EU-DEMO datasets.

\paragraph{3D fluence: denominator $\times$ activation.}
Table~\ref{tab:spatial-denominator-ablation} crosses the output head (softplus vs.\ raw identity) with four residual denominators (detached prediction, live prediction, plain per-sample relative $L_2$, and the noisy MC target) on the 3D fluence dataset ($M{=}3{\times}10^{5}$, $N{=}4$, $60$ epochs of $5{,}000$ scenes, i.e.\ one pass over the training set, FNO modes $16^{3}$, width $32$; evaluation on $100$ held-out scenes with a converged $1{,}048{,}576$-SPP reference).
The detached-prediction denominator is essential: with a \emph{live} prediction denominator, through which gradients flow, the loss is flat wherever the prediction overshoots (gradient $\propto$ target/pred$^2$), so training either inflates its output (softplus: total flux ${\sim}15\times$ too high and linear rel.~$L_2{\approx}500$) or collapses to a negative constant field (identity).
With the \emph{noisy target} in the denominator, zero-flux voxels in the $N{=}4$ labels are divided by the $10^{-10}$ denominator floor, which amplifies their residuals by up to ten orders of magnitude
The softplus head, in turn, only helps when paired with the pointwise stop-gradient residual (first two rows); paired with a per-sample norm ratio it reproduces the absorbing $\softplus(-\infty)=0$ fixed point.
The plain-FNO variant (identity head, per-sample norm) wins the linear median (it directly optimizes that metric) but loses the entire shadow region in log space.

\begin{table}[!htb]
\caption{\textbf{Only the stop-gradient prediction denominator with the softplus head trains stably on 3D fluence.} All runs share data, architecture, and schedule; only the output head and the residual denominator change.
One seed per row; $100$ held-out scenes against the converged reference; the linear median is over scenes and includes the emitter voxel.
\emph{sg} denotes stop-gradient (detached).}
\label{tab:spatial-denominator-ablation}
\centering
\footnotesize
\setlength{\tabcolsep}{3pt}
\begin{tabular*}{\textwidth}{@{\extracolsep{\fill}}llccc>{\raggedright\arraybackslash}p{0.19\textwidth}@{}}
    \toprule
    Head & Denominator & $\log_{10}$~rel.~$L_2$ $\downarrow$ & $\log_{10}$~SSIM $\uparrow$ & Median rel.~$L_2$ $\downarrow$ & Outcome \\
    \midrule
    softplus (\ourNO) & sg prediction   & $\mathbf{0.050}$ & $\mathbf{0.928}$ & $1.03$ & stable \\ %
    identity          & sg prediction   & $0.188$ & $0.755$ & $0.74$ & $8.7\%$ negative voxels \\ %
    softplus          & per-sample norm & $0.801$ & $0.614$ & $1.000$ & collapsed: $\softplus(-\infty)=0$ \\ %
    identity          & per-sample norm & $0.231$ & $0.667$ & $\mathbf{0.066}$ & shadow region lost \\ %
    softplus          & live prediction & $0.166$ & $0.871$ & $515.8$ & output inflated (flux ${\sim}15\times$) \\ %
    identity          & live prediction & $0.801$ & $0.614$ & $6194$ & collapsed: negative constant \\ %
    softplus          & noisy target    & $0.801$ & $0.614$ & $1.000$ & collapsed at epoch 1: all zero \\ %
    identity          & noisy target    & $0.285$ & $0.724$ & $0.999$ & collapse: uniform constant \\ %
    \bottomrule
\end{tabular*}
\end{table}

\paragraph{EU-DEMO: head $\times$ pointwise residual.}
Prediction normalization is the dominant component on EU-DEMO.
Holding the softplus head fixed, replacing \loss with per-sample relative $L_2$ lowers $\log_{10}$ SSIM by $0.3788$ and raises its seed standard deviation from $0.0016$ to $0.1181$.
Holding \loss fixed, replacing softplus with an identity head lowers SSIM by $0.0845$.
The normalization establishes accuracy and seed stability; softplus supplies an additional SSIM gain and enforces positive flux.

\begin{table}[!htb]
\caption{\textbf{On EU-DEMO, the pointwise stop-gradient residual matters most, and softplus adds a further gain.} Five-seed output-head $\times$ residual ablation on the EU-DEMO continuous-energy training setup.
Entries are mean $\pm$ sample standard deviation over five seeds from independent evaluation of final checkpoints on the $24$ held-out configurations at $140{\times}73{\times}146$, scored only on cells where the converged reference is measured ($45.4\%$ of the mesh).
The softplus--\loss and identity--relative-$L_2$ rows share their runs with Table~\ref{tab:main-results}.}
\label{tab:demo-recipe-ablation}
\centering
\footnotesize
\setlength{\tabcolsep}{3pt}
\begin{tabular*}{\textwidth}{@{\extracolsep{\fill}}llcc@{}}
    \toprule
    Head & Residual & $\log_{10}$~SSIM $\uparrow$ & $\log_{10}$~rel.~$L_2$ $\downarrow$ \\
    \midrule
    softplus (\ourNO) & pointwise sg (\loss)   & $\mathbf{0.7923\pm0.0016}$ & $\mathbf{0.0839\pm0.0006}$ \\ %
    identity          & pointwise sg (\loss)   & $0.7078\pm0.0139$          & $0.0885\pm0.0047$ \\ %
    softplus          & per-sample rel.~$L_2$  & $0.4135\pm0.1181$          & $0.2227\pm0.0232$ \\ %
    identity          & per-sample rel.~$L_2$  & $0.1455\pm0.0066$          & $0.2310\pm0.0050$ \\ %
    \bottomrule
\end{tabular*}
\end{table}

\section{Datasets}
\label{sec:data}

\autoref{tab:data-summary} summarizes the four datasets of the main text; Appendix~\ref{app:refract} adds a fifth.

\begin{table}[!htb]
    \caption{\textbf{The four main-text datasets span two physical regimes and two output types.}
    Each maps varying scene inputs to a transport solution on a fixed output grid; ``Varied inputs'' lists what changes between scenes, and held-out scenes are fresh draws from the same design.}
    \label{tab:data-summary}
    \centering
    \scriptsize
    \setlength{\tabcolsep}{2pt}
    \begin{tabular}{@{}>{\raggedright\arraybackslash}p{0.13\textwidth}>{\raggedright\arraybackslash}p{0.12\textwidth}>{\raggedright\arraybackslash}p{0.2\textwidth}>{\raggedright\arraybackslash}p{0.15\textwidth}>{\raggedright\arraybackslash}p{0.16\textwidth}>{\raggedright\arraybackslash}p{0.17\textwidth}@{}}
        \toprule
        Dataset   & Physics               & Solver                        & Geometry                   & Output grid                  & Varied inputs                  \\
        \midrule
        Far-field radiance & radiative transfer      & MC labels; reference by Mitsuba 3~\citep{Mitsuba} & spherical slab (fixed)     & $40\!\times\!80$ far-field radiance     & $\sigma_t,c,Q$ fields, $g$ \\
        3D fluence         & radiative transfer & MC labels and reference, validated against OpenMC~\citep{OpenMC} & cube $[-1,1]^3$ & $64^3$ Cartesian fluence          & $\sigma_t,c$ fields, $g$, source field \\
        Sph.~tokamak       & neutron transport  & OpenMC~\citep{OpenMC} & parametric (Paramak) & $64^3$ Cartesian flux             & 20 (geometry and source)         \\
        EU-DEMO            & neutron transport & OpenMC~\citep{OpenMC}         & EU-DEMO 1/16 wedge (fixed) & $140\!\times\!73\!\times\!146$ flux & 8 (source only)            \\
        \bottomrule
    \end{tabular}
\end{table}

\subsection{Radiative Transfer in a Spherical Slab}
\label{subsec:data-spherical}
The medium is the unit ball $\{\sPos:\|\sPos\|\le 1\}$.
Inside it, the extinction $\totCS_t$ and single-scattering albedo $c$ depend only on the line-of-sight direction $\hat{\sPos}=\sPos/\|\sPos\|$, defined analytically and tabulated on arccos-uniform spherical grids (equal solid-angle bins in the polar angle $\vartheta$, $\vartheta_i=\arccos(1-2(i+\tfrac12)/n_\vartheta)$ for $n_\vartheta$ polar bins; $40\!\times\!80$ for training inputs).
Anisotropic scattering uses a Henyey--Greenstein phase function with one asymmetry parameter $g\in(-0.99,0.99)$ per scene.
The source is an environment map $\Source(\Dir)$ at infinity; the boundary is a null interface, so light enters without refraction or reflection.
A reference spans only about $3$ orders of magnitude (\autoref{tab:dynamic-range}): this dataset is a controlled testbed for label noise and particle-budget allocation, where converged references are affordable, not a high-dynamic-range case.

\paragraph{Sampling.}
Each scene draws three resolution-independent analytical fields plus one scalar.
The source combines $1$--$4$ spherical Gaussian lobes (angular width in $[0.15,0.5]$, random centers) with $1$--$4$ boxes in polar and azimuthal angle $(\vartheta,\varphi)$, each carrying an intensity drawn from $[0.1,10]$.
Extinction and albedo are piecewise-constant Heaviside fields: with probability $\tfrac12$ a spherical checkerboard ($1$--$3$ polar $\times$ $1$--$6$ azimuthal bins), otherwise $1$--$6$ random $(\vartheta,\varphi)$ boxes, with values drawn from $\totCS_t\in[0.01,10]$ and $c\in[0.01,0.99]$.
Because the fields are analytical, the discontinuous region boundaries are represented exactly at every rendering resolution.
The HG parameter is $g\!\sim\!\mathrm{Unif}(-0.99,0.99)$.

\begin{figure}[!htb]
    \centering
    \includegraphics[width=\linewidth]{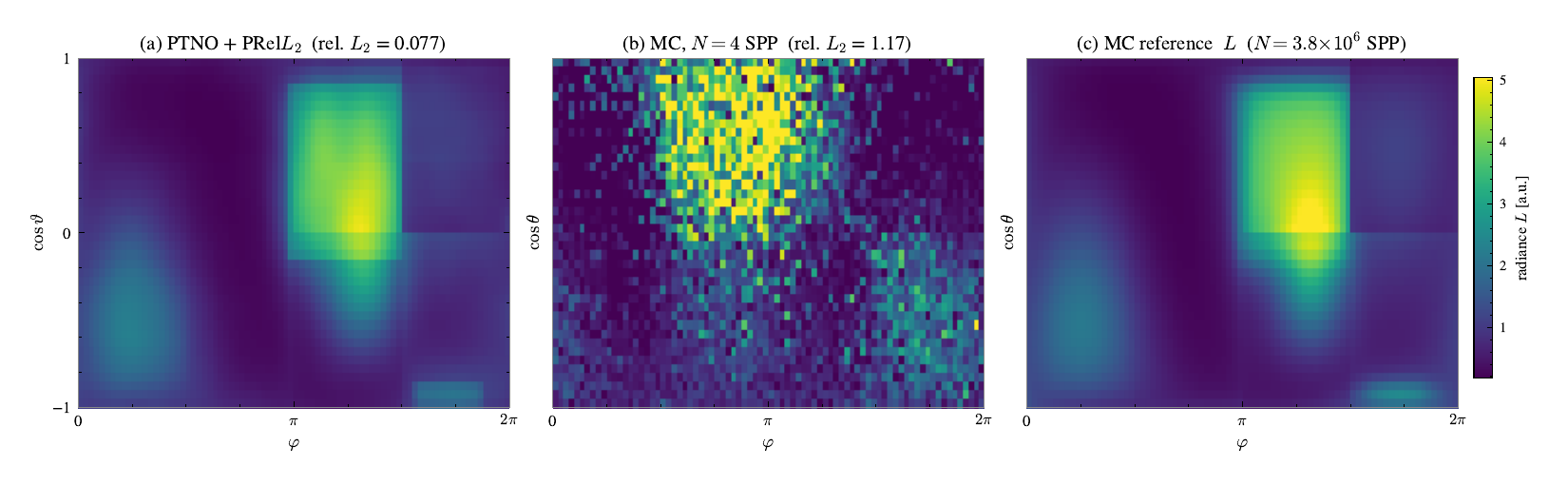}
    \caption{\textbf{\ourNO recovers the converged far-field radiance from the scene inputs, while one training label is mostly noise.} Held-out scene 2 in an equal-solid-angle layout (azimuth $\varphi$ against $\cos\vartheta$) on a shared color scale.
    (a)~The seed-42 \ourNO + \loss checkpoint from Table~\ref{tab:main-results}, using the source, extinction, and albedo tables.
    (b)~A single MC render at the training budget of $N{=}4$ SPP, i.e.\ what one training label looks like.
    (c)~The converged MC reference.
    Panel titles give the rel.~$L_2$ against (c); the cost comparison is in Appendix~\ref{app:cost}.}
    \label{fig:data-spherical}
\end{figure}

\paragraph{Reference.}
Evaluation references use Mitsuba~3's volumetric path tracer with multiple importance sampling~\citep{Mitsuba}, and the training labels are forward MC estimates of the same scenes: a sphere boundary (null BSDF) encloses a heterogeneous medium that reads $(\totCS_t,c)$ from the angular tables, an HG phase function with parameter $g$, and an environment-map emitter carrying the equirectangular source bitmap (resampled to the Mitsuba Y-up convention).
A distant angular sensor records the far-field radiance on the same arccos-uniform $40\!\times\!80$ grid.
Training: $10^6$ scenes with $K\!=\!1$ renders at $4$~SPP (Monte Carlo samples per pixel of the $40\!\times\!80$ sensor) each.
Evaluation: $100$ unseen scenes rendered at native $120\!\times\!240$ with $960\!\times\!1{,}920$ material and environment textures, accumulated adaptively until convergence (per-pixel coefficient of variation~$<\!0.5\%$ and successive-mean rel.~$L_2\!<\!10^{-4}$) or a $10^7$~SPP cap; the $40\!\times\!80$ and $80\!\times\!160$ evaluation grids are conservative equal-solid-angle downsamples of these renders.
An earlier fixed-budget $10^5$~SPP reference on the same $100$ scenes, used for earlier loss-comparison tables, agrees with the converged reference to a median per-scene rel.~$L_2$ of $1.9\%$; re-evaluating the checkpoints that are still available against the converged reference moves the main metrics by $1$--$2\%$ relative.
Figure~\ref{fig:data-spherical} shows one converged scene next to a training-budget render and the \ourNO prediction.

\subsection{3D Fluence}
\label{subsec:data-spatial}
The medium occupies $[-1,1]^3$ in scene units, discretized on a $64^3$ Cartesian grid.
Each scene is built from resolution-independent ingredients: $2$--$5$ large axis-aligned boxes in a near-vacuum background ($\totCS_t=0.01$, $c=0.5$), each box carrying its own extinction $\totCS_t$ and albedo $c$, and $1$--$3$ point emitters at random positions.
About half of the emitters are isotropic; the others carry a random Perlin angular emission profile in a random orientation.
Anisotropic scattering uses an HG phase function with one $g$ per scene.
The boundary is an open vacuum.

\paragraph{Sampling.}
Per scene: box centers $\mathrm{Unif}(-0.6,0.6)^3$ and half-extents $\mathrm{Unif}(0.1,0.7)$ per axis, with $\totCS_t\!\sim\!\mathrm{Unif}(0.01,50)$ and $c\!\sim\!\mathrm{Unif}(0.01,0.99)$ drawn per box and later boxes overwriting earlier ones where they overlap; emitter positions $\mathrm{Unif}(-0.85,0.85)^3$ and intensity $\mathrm{Unif}(1,20)$, with a Perlin angular profile (values in $[0.1,2]$) on each emitter with probability $1/2$; $g\!\sim\!\mathrm{Unif}(-0.95,0.95)$.
The $100$ evaluation scenes realize $g\in[-0.92,0.93]$.

\paragraph{Reference.}
All renders use a forward Monte Carlo collision estimator with double-precision per-voxel accumulators (necessary for HDR), cross-checked against OpenMC~\citep{OpenMC} on the evaluation scenes.
Training: $300{,}000$ scenes with $K{=}2$ independent noisy renders at $4$~SPP each, i.e., $4$ Monte Carlo samples per voxel of the $64^3$ grid ($\sim\!10^6$ particles per render---a deliberately weak supervision regime that exercises the loss family at extreme noise).
Evaluation: a held-out batch of $100$ unseen scenes whose reference flux is re-rendered to $\sim\!10^6$~SPP ($\sim\!2.7\!\times\!10^{11}$ particles per scene, $\sim\!2.6\!\times\!10^{5}$ times one training render).
The reference fluence of a scene spans about $10$ orders of magnitude (\autoref{tab:dynamic-range}); over the $100$ evaluation scenes, $2.3\%$ of reference voxels are exact zeros that no particle reaches even at the reference budget (cast shadows), against $18\%$ in a four-SPP training render (both means over scenes).

\begin{figure}[!htb]
    \centering
    \includegraphics[width=\linewidth]{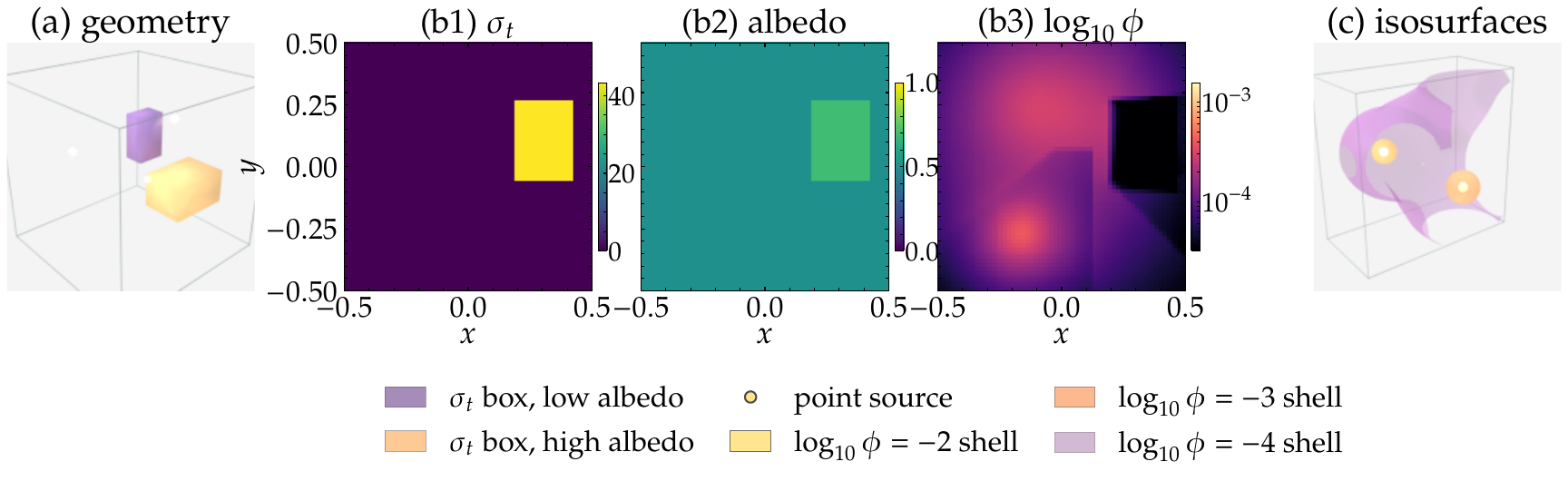}
    \caption{\textbf{In the 3D fluence dataset, contrast blocks cast shadows that span orders of magnitude around the point emitters.} (a)~A representative scene: two $\totCS_t$ blocks rendered as translucent boxes inside the cube (axes in normalized units) (purple / orange tint by $\totCS_t$ magnitude), with the three point sources as bright spheres. (b)~Mid-cut slices ($z\!=\!32$, i.e.\ $z\!=\!0$ in scene units) of the same scene's $\totCS_t$, albedo, and reference fluence (log scale); only the high-$\totCS_t$/high-albedo block intersects this plane; the dark rectangle in the fluence panel is its cast shadow. (c)~Half-cut ($X\!\le\!0$ visible) view of the same scene's log-fluence iso-surfaces at $\{-2,-3,-4\}$, illustrating the multi-order-of-magnitude attenuation around the two emitters in this half.}
    \label{fig:data-spatial}
\end{figure}

\begin{figure}[!htb]
    \centering
  \begingroup%
  \catcode`\_=8\relax%
  \edef\pgfincludename{\detokenize{fig_spatial_samples}}%
  \graphicspath{{fig/\pgfincludename/}{fig/}}%
  \setkeys{Gin}{draft=false}%
  \resizebox{\linewidth}{!}{\input{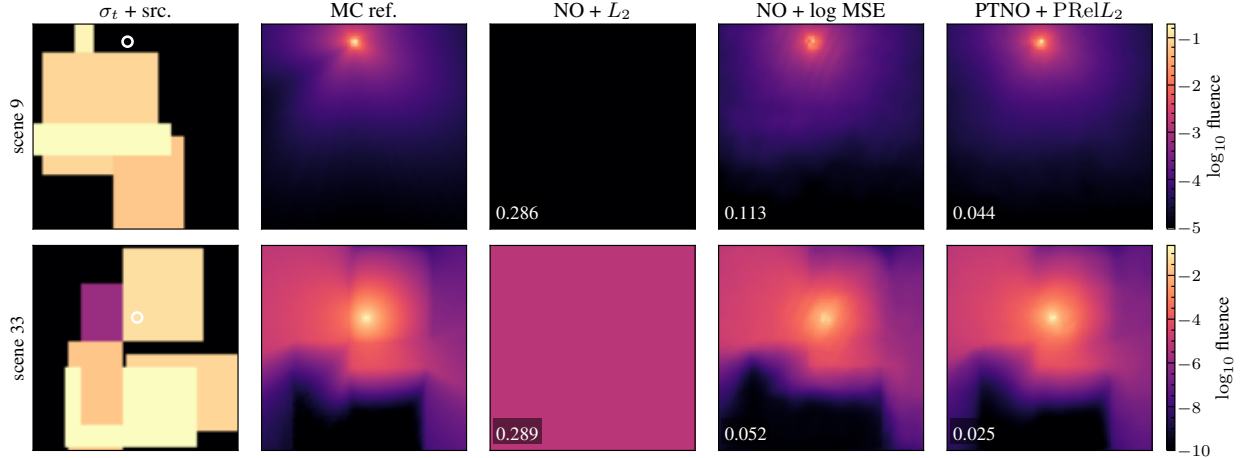}}%
  \endgroup%

    \caption{\textbf{\ourNO + \loss more closely reproduces the source halo and shadowed fluence than the baseline losses.} Predictions from the seed-42 checkpoints in Table~\ref{tab:main-results} on held-out scenes 9 and 33.
    Columns: extinction $\sigma_t$ with source locations marked, MC reference (log scale), and predictions from \ourNO and the baseline losses on a shared row-wise color scale; inset numbers are per-scene $\log_{10}$~rel.~$L_2$.
    The raw-output $L_2$ prediction is nearly spatially constant; log MSE retains the main spatial structure but has larger errors near the source and in dim regions.}
    \label{fig:spatial-samples}
\end{figure}

\subsection{Spherical-Tokamak Neutronics}
\label{subsec:data-sptok}
Each scene is a full toroidal fusion spherical tokamak built parametrically with Paramak using a 6-axis radial build (center-column shield inner/outer radius, blanket thickness, divertor width) plus major and minor radius~\cite{shimwell2021paramak}.
The geometry is configured with a plasma source expressed as toroidal rings, radially weighted as a function of fusion reactivity in an H-mode configuration~\cite{Delaporte-Mathurin_OpenMC_Plasma_Source_2023}.
The CAD assembly is voxelized and tagged with five materials (vacuum, plain-carbon shield, ferritic steel first wall and vessel, Pb--Li breeder blanket).
The bounding box, a cube of side $1{,}800$\,cm, is voxelized on a $64^3$ Cartesian mesh; the per-voxel material id is queried from the OpenMC geometry, giving a volumetric label that conditions the operator.
The prediction target is the raw OpenMC flux-tally mean in each energy group, in neutron-cm per source neutron, integrated over each mesh cell and without conversion to a reactor source rate.

\paragraph{Sampling.}
The 20 scalar parameters in Table~\ref{tab:sptok-sampling} are drawn from Latin-hypercube space-filling designs over the box specified there: 4 paramak radial-build axes, the four D-shape descriptors $(R_0,a_{\mathrm{p}},\kappa,\delta)$ shared between the paramak surface and the H-mode plasma source~\cite{Delaporte-Mathurin_OpenMC_Plasma_Source_2023} (D--T fusion ring), 10 axes for the peaked ion-density and ion-temperature profiles (origin, pedestal/separatrix fractions, peaking factor, pedestal radius fraction, temperature exponent $\beta$) together with a Shafranov shift of the magnetic axis, and a rigid radial offset of the whole source.
The 8 geometry parameters are drawn once per group of $32$ consecutive training configurations, which share one CAD model, while the source parameters are drawn per configuration; the source offset has its own one-dimensional design.

\begin{table}[!htb]
    \centering
    \caption{\textbf{Spherical-tokamak scenes vary 20 geometry and plasma-source parameters.} Sampling box of the 20 scalar parameters, drawn from Latin-hypercube designs.
    The geometry block parameterizes the paramak radial build; D-shape descriptors $(R_0,a_{\mathrm{p}},\kappa,\delta)$ are shared between the paramak surface and the plasma source ring; the profile blocks parameterize the H-mode ion-density and ion-temperature radial profiles.}
    \label{tab:sptok-sampling}
    \scriptsize
    \setlength{\tabcolsep}{4pt}
    \begin{tabular*}{\textwidth}{@{\extracolsep{\fill}}llll@{}}
        \toprule
        Symbol & Range & Unit & Description \\
        \midrule
        \multicolumn{4}{l}{\textit{Paramak radial build (geometry only)}} \\
        $R_{\mathrm{cs,in}}$    & $[20,\,40]$  & cm & center-column shield inner radius \\
        $R_{\mathrm{cs,out}}$   & $[55,\,95]$  & cm & center-column shield outer radius \\
        $t_{\mathrm{blk}}$      & $[40,\,110]$ & cm & blanket thickness \\
        $w_{\mathrm{div}}$      & $[25,\,65]$  & cm & divertor width \\
        \midrule
        \multicolumn{4}{l}{\textit{D-shape descriptors (shared geometry $\leftrightarrow$ source)}} \\
        $R_{0}$                 & $[250,\,500]$ & cm & major radius \\
        $a_{\mathrm{p}}$        & $[60,\,140]$  & cm & minor radius \\
        $\kappa$                & $[1.2,\,2.0]$ & ---  & elongation \\
        $\delta$                & $[0.2,\,0.6]$ & ---  & triangularity \\
        \midrule
        \multicolumn{4}{l}{\textit{Magnetic-axis and source offsets (source only)}} \\
        $\Delta_{\mathrm{Shaf}}$ & $[0,\,20]$   & cm & Shafranov shift of the magnetic axis \\
        $f_{\mathrm{off}}$ & $[-1,\,1]$ & --- & radial source offset, as a fraction of $\min(\text{plasma gap},\,20\,\mathrm{cm})$ \\
        \midrule
        \multicolumn{4}{l}{\textit{Ion-density profile (source only)}} \\
        $n_{i,0}$               & $[8\!\times\!10^{18},\,1.2\!\times\!10^{20}]$ & m\textsuperscript{-3} & on-axis ion density \\
        $f_{n,\mathrm{ped}}$    & $[0.65,\,0.95]$ & --- & pedestal-to-axis density ratio \\
        $f_{n,\mathrm{sep}}$    & $[0.08,\,0.22]$ & --- & separatrix-to-axis density ratio \\
        $\alpha_{n}$            & $[0.8,\,2.0]$  & --- & density peaking factor \\
        $r_{\mathrm{ped}}/a_{\mathrm{p}}$    & $[0.75,\,0.92]$ & --- & pedestal radius fraction \\
        \midrule
        \multicolumn{4}{l}{\textit{Ion-temperature profile (source only)}} \\
        $T_{i,0}$               & $[8,\,25]$    & keV & on-axis ion temperature \\
        $f_{T,\mathrm{ped}}$    & $[0.25,\,0.60]$ & --- & pedestal-to-axis temperature ratio \\
        $f_{T,\mathrm{sep}}$    & $[0.01,\,0.06]$ & --- & separatrix-to-axis temperature ratio \\
        $\alpha_{T}$            & $[1.5,\,5.0]$ & --- & temperature peaking factor \\
        $\beta_{T}$             & $[1.5,\,4.5]$ & --- & temperature profile exponent \\
        \bottomrule
    \end{tabular*}
\end{table}

\paragraph{Training and reference.}
The training pool has $1{,}024{,}000$ configurations, each labeled by one OpenMC fixed-source run of $N{=}10^5$ histories ($10$ batches of $10^4$); training draws $4\times10^5$ of them without repetition ($10^5$ optimizer updates at batch size $4$).
The $50$ held-out configurations carry a $10^9$-history reference pooled from $100$ independent $10^7$-history runs; its residual noise is $0.017$ in $\log_{10}$ relative $L_2$, against $0.045$ for the operator.
The operator is a 3D FNO (modes $24^3$, width $32$, $4$ layers) trained for $10^5$ optimizer updates.

\begin{figure}[!htb]
    \centering
    \includegraphics[width=\linewidth]{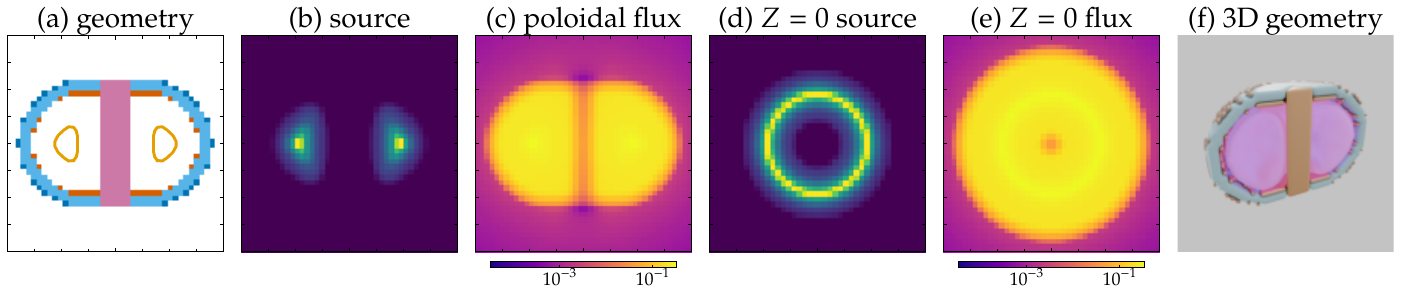}
    \caption{\textbf{Each spherical-tokamak scene couples a parametric radial build to a D-shaped plasma source, and the flux falls by orders of magnitude from the plasma through the shield and blanket.} (a) Paramak radial build for one scene with the D-shaped plasma source outline, major/minor radius, elongation, and triangularity annotated. (b)--(e) Per-scene $64^3$ mesh views: toroidally averaged source $Q$ and group-summed flux (log) on the poloidal plane, then $Z{=}0$ source and flux slices showing the annular plasma ring and inboard/outboard attenuation. (f) Half-cut 3D view: the toroidal source isosurface inside translucent shield/first-wall/blanket/vessel shells.}
    \label{fig:data-sptok}
\end{figure}

\subsection{EU-DEMO 1/16 Sector}
\label{subsec:data-demo}
The fourth dataset uses an OpenMC model over a 1/16 toroidal slice of the EU-DEMO reactor~\citep{flammini2019demo_mcnp_22p5}.
Sector neutronics models of this kind are widely used in fusion analysis because they retain the main blanket, vessel, shielding, and source-transport physics of a reactor-scale configuration while exploiting toroidal symmetry to reduce computational cost.
They are a standard setting for studies of neutron flux distribution, nuclear heating, shielding performance, activation, and diagnostic response in DEMO-like designs.

The geometry of the reactor is modeled using a CSG geometry of about $5{,}300$ surfaces (planes, quadrics, cylinders, tori, and cones)~\cite{DEMO_diagnostics, lu2020serpent_mcnp_demo}.
The surfaces describing the inboard/outboard blanket modules, divertor, vacuum vessel, and shielding, are closed by two reflective planes at $\pm 11.25^{\circ}$ and a vacuum plane at $z={-}2{,}732$\,cm.
Materials use the upstream physics-resolved compositions (water-cooled lithium--lead blanket, Eurofer first wall and vessel, etc.) with TENDL-2019 nuclear cross-section data.
Geometry, materials, and reactor configuration are fixed across the dataset; only the plasma source varies.

\paragraph{Sampling.}
The 8 plasma-source shape parameters in Table~\ref{tab:demo-sampling} are drawn from a Latin-hypercube space-filling design over the upstream ranges of the EU-DEMO parametric plasma source (PPS) configuration.

\begin{table}[!htb]
    \centering
    \caption{\textbf{EU-DEMO scenes vary only the plasma source, through 8 shape parameters.} Sampling box of these parameters, drawn from a Latin-hypercube space-filling design.
    Ranges are the upstream EU-DEMO PPS configuration; the geometry and materials are fixed.}
    \label{tab:demo-sampling}
    \scriptsize
    \setlength{\tabcolsep}{4pt}
    \begin{tabular*}{\textwidth}{@{\extracolsep{\fill}}llll@{}}
        \toprule
        Symbol & Range & Unit & Description \\
        \midrule
        $T$         & $[14.00,\,17.00]$       & keV & plasma ion temperature (D--T thermal broadening) \\
        $\alpha$    & $[1.20,\,1.80]$         & --- & radial peaking factor of the source profile \\
        $R_{0}$     & $[715.04,\,1{,}072.56]$    & cm  & major radius of the plasma ring \\
        $a_{\mathrm{p}}$ & $[230.64,\,345.96]$     & cm  & minor radius of the plasma ring \\
        $\kappa$    & $[1.32,\,1.98]$         & --- & plasma elongation \\
        $\delta$    & $[0.2664,\,0.3996]$     & --- & plasma triangularity \\
        $\Delta R$  & $[-20,\,20]$            & cm  & radial shift of the magnetic axis \\
        $\Delta Z$  & $[-20,\,20]$            & cm  & vertical shift of the magnetic axis \\
        \bottomrule
    \end{tabular*}
\end{table}

\begin{figure}[!htb]
    \centering
    \makebox[\linewidth][c]{\includegraphics[width=\linewidth]{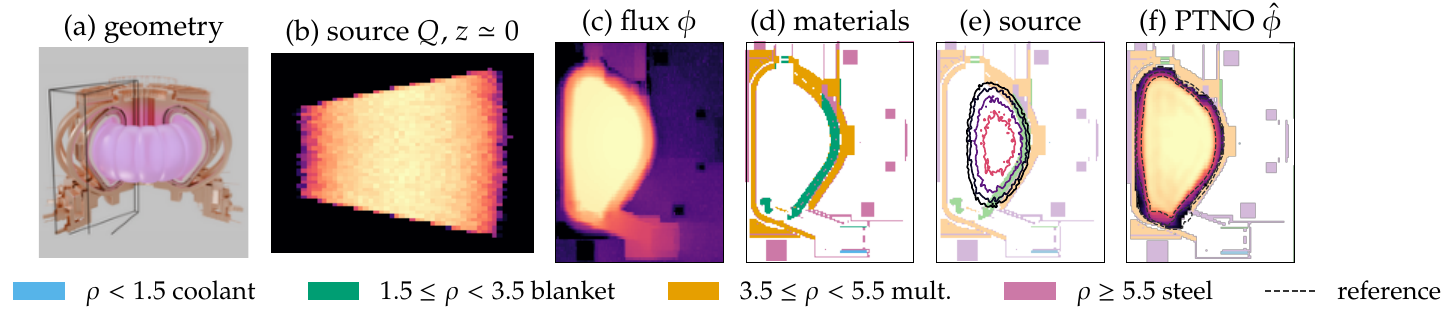}}
    \caption{\textbf{The EU-DEMO reference flux falls by more than nine orders of magnitude over one poloidal slice, and the \ourNO prediction follows it through the inboard blanket toward the divertor.} One held-out validation configuration (the scene of \autoref{fig:intro-demo}).
    \emph{(a)}~Rendering of the DEMO tokamak~\cite{DEMO_diagnostics}; the wireframe brackets the 1/16 wedge whose poloidal cross-section is plotted in the remaining panels.
    \emph{(b)}~The $z\!\simeq\!0$ slice of the plasma source $Q$ on the $140\times73\times146$ training mesh.
    \emph{(c)}~The converged reference flux $\phi$ on the physical $y\!=\!0$ slice, shown over nine orders of magnitude below its peak.
    \emph{(d)}~Material classes on the same slice, binned by mass density as in the legend.
    \emph{(e)}~Iso-contours of the source over the materials; the D-shaped plasma sits inside the vacuum vessel.
    \emph{(f)}~The \ourNO prediction $\hat\phi$ over three orders of magnitude, with dashed reference iso-contours at $10^{-1}$, $10^{-2}$, and $10^{-3}$ of the peak; the predicted field follows the reference through the inboard blanket and toward the divertor.}
    \label{fig:data-demo}
\end{figure}

\paragraph{Reference.}
The neutron source is a parametric D--T plasma source parameterized by the 8 plasma-shape variables above together with the wedge angles.
The emitted neutron energy is modeled as a thermally broadened 14.1\,MeV D--T fusion spectrum, with broadening controlled by the ion temperature.
The source is broadly similar in spirit to the more general confinement-mode-dependent source models described by~\cite{fausser2012tokamak}, but uses a reduced geometric parameterization and a simplified spectral model.
Per scene we (i) voxelize the source onto the flux-tally mesh, so that the operator input and output share a coordinate frame, and (ii) run OpenMC continuous-energy fixed-source transport with neutron-flux mesh tallies on a trimmed $140\times73\times146$ Cartesian mesh.
The prediction target is the OpenMC flux-tally mean divided by the mesh-cell volume, in neutrons per square centimeter per source neutron, without conversion to a reactor source rate.
The $24$ held-out configurations use a reference computed at $700\times365\times730$ with FW-CADIS and MAGIC weight windows, a median of $6{,}130$ core-hours per configuration, and conservatively block-averaged to the $70$, $140$, and $280$ grids used for evaluation.
Figure~\ref{fig:data-demo} shows the wedge geometry and a representative scene.

\paragraph{Training.}
$M{=}10^5$ configurations from the sampling design above, each labeled by one OpenMC run of $N{=}5\times10^5$ histories on the $140\times73\times146$ mesh ($K{=}1$).
The backbone is a 3D FNO (modes $32\times24\times38$, width $32$, $4$ layers, $6.3\times10^7$ parameters) that reads a four-class material-density geometry channel and the 8 source parameters as NeRF-embedded scalars ($37$ input channels).
Every EU-DEMO model in Table~\ref{tab:main-results} trains with AdamW (learning rate $10^{-3}$, no weight decay, gradient clipping at $1.0$) at batch size $1$ with $4$-step gradient accumulation for two passes over the $10^5$ configurations, that is $5\times10^4$ optimizer updates under a cosine schedule that reaches zero at the last update, taking about $2.9$ hours per run on one H200.
Log-space metrics use a floor of $10^{-17}$ for both predictions and references.

\section{Evaluation Protocol}
\label{app:eval-protocol}

Every metric is computed per scene on the physical prediction $\hat\phi$ and reference $\phi$, averaged over the scenes of the held-out set, and reported as the mean $\pm$ sample standard deviation of that average across training seeds, or as the two per-seed values when there are two seeds; rows that report a median over scenes say so.
A scene is scored on a set of cells $\mathcal{C}$ with weights $w_i$, $\sum_{i\in\mathcal{C}}w_i=1$, given by the cell volume on Cartesian grids and by the solid angle on the far-field sphere.
The structured grids have equal-measure cells (uniform Cartesian voxels and arccos-uniform angular bins), so their weights are uniform.
The tetrahedral extension in Appendix~\ref{app:tetra-status} instead uses the volume of each tetrahedron.

For a field $u$, which is either $\phi$ itself (linear metrics) or $\log_{10}\max(\phi,\epsilon)$ ($\log_{10}$ metrics, with the same floor $\epsilon$ applied to prediction and reference), we report
\begin{align}
  \mathrm{rel}L_2(\hat u,u) &= \Big(\textstyle\sum_{i\in\mathcal{C}} w_i(\hat u_i-u_i)^2 \big/ \sum_{i\in\mathcal{C}} w_i u_i^2\Big)^{1/2}, \\
  \mathrm{PSNR}(\hat u,u) &= 10\log_{10}\!\Big(R^2 \big/ \textstyle\sum_{i\in\mathcal{C}} w_i(\hat u_i-u_i)^2\Big), \\
  \mathrm{SSIM}(\hat u,u) &= \textstyle\sum_{i\in\mathcal{C}} w_i\,\dfrac{(2\mu_{\hat u,i}\mu_{u,i}+C_1)(2\sigma_{\hat u u,i}+C_2)}{(\mu_{\hat u,i}^2+\mu_{u,i}^2+C_1)(\sigma_{\hat u,i}^2+\sigma_{u,i}^2+C_2)}.
\end{align}
Relative $L_2$ is a ratio of sums over the scene, not a mean of per-cell ratios, and the \%~RMSE of some appendix tables is $100$ times the linear relative $L_2$.
For structured-grid SSIM, $\mu$, $\sigma^2$, and $\sigma_{\hat u u}$ are local means, variances, and covariance under a separable Gaussian window with standard deviation $1.5$ cells truncated to $11$ taps per axis (2D on the sphere, 3D on volumes), with reflective padding at the domain boundary, $C_1=(0.01R)^2$, and $C_2=(0.03R)^2$.
The tetrahedral extension instead uses volume-weighted neighborhoods with a physical scale of $100$\,cm (Appendix~\ref{app:tetra-status}).
The data range $R$ is the span of the scene's own reference: $\max u-\min u$, which for the $\log_{10}$ metrics is $\log_{10}\max\phi-\log_{10}\max(\min\phi,\epsilon)$; predictions are never rescaled independently of the reference.
On the multi-group spherical-tokamak flux, every metric is computed per energy group and averaged over the groups whose reference is not identically zero.
The reporting rule is the final checkpoint of a fixed training schedule, with no checkpoint selection on a validation or reference set.
The far-field reruns (200 epochs, $10^5$ updates), EU-DEMO reruns, tetrahedral extension, and interface models follow this rule.
The 3D fluence main-table replacement is still running, and the original spherical-tokamak training corpus is unavailable locally for an exact rerun; their historical rows are explicitly provisional because checkpoints were selected on the same configurations used for evaluation.
Earlier single-seed diagnostics and qualitative figures retain their documented historical checkpoints until replaced.
Far-field hyperparameters, including HPO candidates and promotions, were chosen using the same 100 evaluation scenes; the new runs remove checkpoint selection but do not provide an independent test of hyperparameter selection.

\autoref{tab:eval-protocol} lists the settings that differ between datasets; within a dataset they are identical for every model compared.
Some tables report linear metrics with the voxels around point emitters removed, or score EU-DEMO only on the cells its reference measured; each such table says so in its caption.

\begin{table}[!htb]
\caption{\textbf{Evaluation settings for the five structured-grid datasets.} $\epsilon$ is the floor of the $\log_{10}$ metrics; these grids have equal-measure cells, so the measure weights are uniform; $R$ is the reference's own span for PSNR and SSIM.
The SSIM window is the same Gaussian ($\sigma=1.5$ cells, $11$ taps per axis) on these grids; tetrahedral settings are in Appendix~\ref{app:tetra-status}.}
\label{tab:eval-protocol}
\centering
\footnotesize
\setlength{\tabcolsep}{3pt}
\begin{tabular*}{\textwidth}{@{\extracolsep{\fill}}llll@{}}
  \toprule
  Dataset & $\epsilon$ & Measure and grid & Data range $R$ \\
  \midrule
  Far-field radiance & $10^{-6}$ & solid angle, $40{\times}80$ & span of the scene's reference \\
  3D fluence & $10^{-10}$ & volume, $64^3$ & span of the scene's reference \\
  Spherical tokamak & $10^{-10}$ & volume, $64^3$, per energy group & span of the scene's reference, per group \\ %
  EU-DEMO wedge & $10^{-17}$ & volume, $140{\times}73{\times}146$ & span of the scene's reference \\ %
  Interfaces (App.~\ref{app:refract}) & $3.28\times10^{-7}$ & volume, $64^3$ & span of the scene's reference \\ %
  \bottomrule
\end{tabular*}
\end{table}

\graphicspath{{./fig/debias/}}

\section{Explicit Debiasers for Log-Space Training}
\label{app:exp2-debias}

\begin{figure}[H]
    \centering
  \begingroup%
  \catcode`\_=8\relax%
  \edef\pgfincludename{\detokenize{fig_bias_variance}}%
  \graphicspath{{fig/\pgfincludename/}{fig/}}%
  \setkeys{Gin}{draft=false}%
  \resizebox{\linewidth}{!}{\input{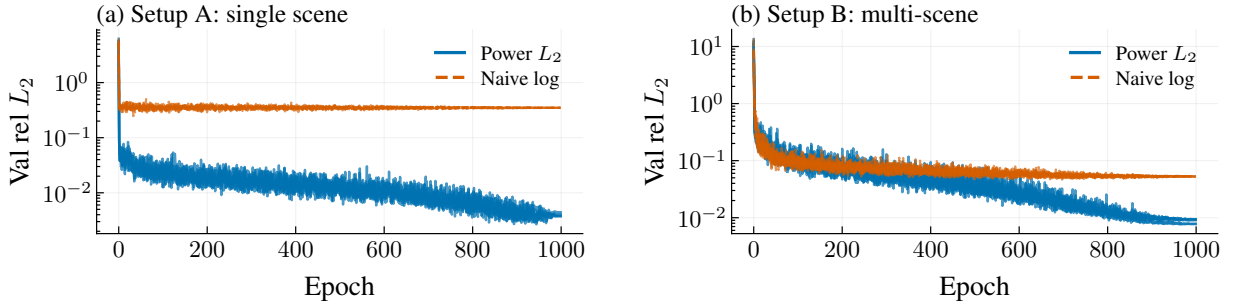}}%
  \endgroup%

    \caption{\textbf{Log MSE converges to a higher error floor than training in physical space.} Validation linear rel.~$L_2$ against training epoch for (a) Setup~A, one scene with $4$-SPP labels, and (b) Setup~B, $M{=}10^6$ training scenes with $80$-SPP labels; ``Naive log'' in the legend is log MSE, and ``Power $L_2$'' is the earlier physical-space $L_2$ loss on a $10^{\NO}$ head described in the setup above, not the softplus head with \loss.}
    \label{fig:bias-variance}
\end{figure}

\begin{figure}[H]
    \centering
  \begingroup%
  \catcode`\_=8\relax%
  \edef\pgfincludename{\detokenize{fig_jensen}}%
  \graphicspath{{fig/\pgfincludename/}{fig/}}%
  \setkeys{Gin}{draft=false}%
  \resizebox{\linewidth}{!}{\input{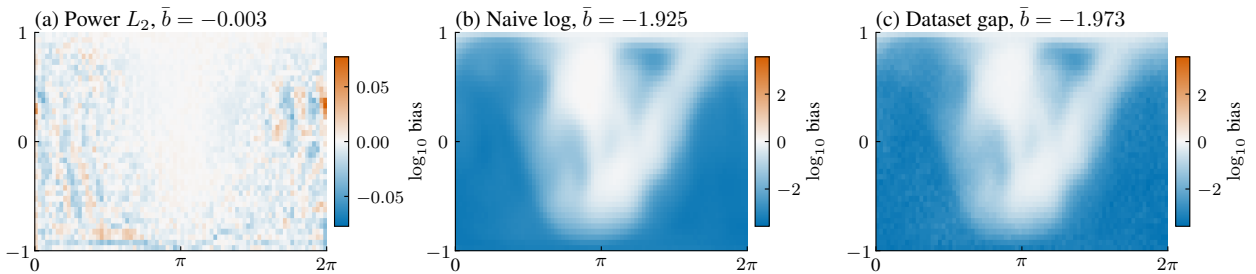}}%
  \endgroup%

    \caption{\textbf{Log MSE learns the biased target; training in physical space does not.}
    Setup~A. Per-pixel mean residual $\hat b(\sPos)$ is $\log_{10}$ prediction minus $\log_{10}\sol(\param,\sPos)$; $\bar{b}$ is the spatial mean.
    (a)~Physical-space $L_2$ on a $10^{\NO}$ head, labeled ``Power $L_2$'' (an earlier version of our method; $\bar{b}{\approx}0$).
    (b)~log MSE ($\bar{b}{\approx}{-}1.9$).
    (c)~Empirical Jensen bias estimated on the train set, matching panel (b).}
    \label{fig:jensen}
\end{figure}

\paragraph{Setup.}
The single-seed and earlier-model figures below retain their historical checkpoints.
Tables~\ref{tab:exp1} and~\ref{tab:inverse-mc-v2} use the new fixed-schedule final checkpoints; the remaining diagnostics await replacement.
We confirm the $\log_{10}$-target Jensen bias under two configurations: Setup~A (single scene, $M{=}1$, 5 seeds, labels of one $4$-SPP render) and Setup~B (multi-scene, $M{=}10^{6}$ train / $M{=}100$ test, 6 seeds, labels averaging $20$ four-SPP renders, i.e.\ $80$~SPP), both for $10^{5}$ optimization steps.
Figures~\ref{fig:bias-variance} and~\ref{fig:jensen} show that log MSE converges to a strictly higher validation rel.~$L_2$ than a model trained in physical space (an earlier version of our method: a physical-space $L_2$ loss on a $10^{\NO}$ output head, without the softplus head and the pointwise normalization of \loss), and that an NO trained against $\log_{10}\!\MC_N(a,\xi)$ systematically under-predicts the solution by the empirical Jensen offset.

\paragraph{Stratified loss comparison.}
Table~\ref{tab:exp1} stratifies \%~RMSE and $\log_{10}$~SSIM by reference brightness across the $100$ held-out far-field scenes, for the five-seed far-field models of \autoref{tab:main-results}.
Log MSE underperforms \ourNO + \loss in every bin; $L_2$ on raw NO output loses contrast in dim regions ($\log_{10}$~SSIM $0.24$ in the dimmest bin and at most $0.78$ in any bin); \ourNO + \loss keeps $\log_{10}$~SSIM between $0.89$ and $0.94$ across all four brightness bins.
This table does not separate the contributions of the softplus head and the \loss residual to the dim-bin improvement; Appendix~\ref{app:denominator-ablation} does.

\begin{table}[ht]
\caption{\textbf{\ourNO + \loss is the only loss that is accurate in every brightness bin.}
Far-field radiance, $100$ held-out scenes against the converged reference; columns stratify scenes by their mean reference radiance ($8$, $16$, $42$, and $32$ scenes), and the last column covers all $100$ scenes (two lie outside the bin edges) and reproduces \autoref{tab:main-results}.
Mean $\pm$ standard deviation over the five seeds of \autoref{tab:main-results}; bolded values are best per metric and bin.}
\label{tab:exp1}
\centering
\setlength{\tabcolsep}{2pt}
\scriptsize
\begin{tabular*}{\textwidth}{@{\extracolsep{\fill}}lllccccc@{}}
    \toprule
    Metric & Arch. & Loss
        & $0.0016$--$0.013$
        & $0.013$--$0.10$
        & $0.10$--$0.85$
        & $0.85$--$6.9$
        & All \\
    \midrule
        \multirow{3}{*}{\%~RMSE $\downarrow$}
        & NO & $L_2$ & $64.0\pm6.9$ & $18.5\pm1.5$ & $\mathbf{8.2\pm0.2}$ & $\mathbf{5.6\pm0.1}$ & $\mathbf{15.3\pm1.3}$ \\
        & NO & log MSE & $99.9\pm0.0$ & $90.6\pm0.2$ & $58.2\pm0.2$ & $29.7\pm0.2$ & $57.6\pm0.2$ \\
        & \ourNO & \loss (ours) & $\mathbf{29.1\pm2.7}$ & $\mathbf{15.7\pm1.8}$ & $15.4\pm0.5$ & $14.4\pm0.6$ & $16.4\pm0.6$ \\
    \midrule
    \multirow{3}{*}{$\log_{10}$~SSIM $\uparrow$}
        & NO & $L_2$ & $0.241\pm0.079$ & $0.490\pm0.046$ & $0.699\pm0.009$ & $0.781\pm0.009$ & $0.652\pm0.020$ \\
        & NO & log MSE & $0.514\pm0.001$ & $0.509\pm0.001$ & $0.441\pm0.001$ & $0.471\pm0.002$ & $0.470\pm0.001$ \\
        & \ourNO & \loss (ours) & $\mathbf{0.935\pm0.003}$ & $\mathbf{0.945\pm0.006}$ & $\mathbf{0.925\pm0.004}$ & $\mathbf{0.888\pm0.005}$ & $\mathbf{0.917\pm0.004}$ \\
    \bottomrule
\end{tabular*}
\end{table}

\paragraph{Explicit log-target debiasers.}
The log MSE row in Table~\ref{tab:exp2} suffers Jensen bias because $\E\!\left[\log_{10}\MC_N\right]\neq\log_{10}\E\!\left[\MC_N\right]$ at finite $N$.
A natural fix is to construct an explicitly debiased or lower-bias $\log_{10}$-target estimator from $K$ independent renders and feed that into a standard log-space MSE.
We compare three such estimators against \loss trained with the exponential output head $10^{\NO}$ of an earlier model version:
\begin{itemize}
    \item \emph{Log of mean} ($K{=}10$): the $\log_{10}$ of the average of $10$ available $4$-SPP renders, equivalent in expectation to one $40$-SPP render.
    \item \emph{Second-moment correction} ($K{=}10$): $\log_{10}\bar Y_K + \widehat\sigma^2/(2K\bar Y_K^2\ln 10)$ where the same $K$ renders are used to estimate the mean and variance.
    \item \emph{Taylor fallback} ($5{+}5$): a Taylor--Russian-roulette log estimator with a second-moment fallback on high-coefficient-of-variation pixels.
\end{itemize}

\paragraph{Implementation of the log-target debiasers.}
All log-target debiasers are applied independently at every output pixel to the stored stack of repeated $4$-SPP renders for the same scene.
Let $Y_1,\ldots,Y_K$ denote the selected independent renders, let $\bar Y_K=K^{-1}\sum_iY_i$, and let $\epsilon=10^{-6}$ be the same floor used by the far-field evaluation.
The log-of-mean target is
\[
    \widehat z_{\mathrm{mean}}
    = \log_{10}\!\left(\max(\bar Y_{10},\epsilon)\right).
\]
The second-moment row uses the delta-method correction
\[
    \widehat z_{\mathrm{2nd}}
    = \frac{1}{\ln 10}\left[
        \ln\!\left(\max(\bar Y_{10},\epsilon)\right)
        + \frac{\widehat\sigma^2}{2K\max(\bar Y_{10},\epsilon)^2}
    \right],
\]
where $\widehat\sigma^2$ is the unbiased sample variance of the same $K{=}10$ renders.
The Taylor fallback row splits the selected renders into $5$ anchor renders and $5$ correction renders.
The anchor group forms \(\hat\mu=\max(\frac{1}{5}\sum_{i=1}^{5}Y_i,\epsilon)\).
For correction render $Y_{5+j}$, $j=1,\ldots,5$, define $r_j=(Y_{5+j}-\hat\mu)/\hat\mu$.
The estimator uses product terms from the Taylor series of the log of the pixel mean $\mu=\E[Y_i]$
\[
    \ln \mu
    = \ln \hat\mu
      + \sum_{m\ge 1}\frac{(-1)^{m+1}}{m}
        \left(\frac{\mu-\hat\mu}{\hat\mu}\right)^m,
\]
with $\prod_{j=1}^{m}r_j$ estimating the $m$th power term.
We average $32$ Russian-roulette paths, i.e., randomly truncated series whose surviving terms are reweighted by their survival probability, with continuation probability $q_m=\min(\mathrm{CV}\,m/(m+1),0.95)$, where CV is the coefficient of variation estimated from the selected renders.
Pixels with $\mathrm{CV}\ge1$ or non-finite roulette output fall back to the second-moment estimator above.
These procedures debias the transformed target as much as possible under the $K{=}10$ render budget, but the resulting training objective is still ordinary log-space MSE; the $K$ extra renders are amortized over the training set, not the stochastic gradient.

Table~\ref{tab:exp2-debias} reports metrics averaged over the $100$ held-out scenes against the naive baseline at $N{=}4$; each debiaser cuts the naive-log \%~RMSE by $50$--$59\%$, but none closes the gap to \loss, shown here with the exponential $10^{\NO}$ output head of an earlier model version rather than the softplus head of \ourNO.

\begin{table}[!htb]
\caption{\textbf{Even with explicit Jensen-bias correction, log-target debiasers leave a substantial residual gap.} All methods are trained at $N{=}4$ on the far-field radiance dataset; explicit debiasing methods consume up to $K{=}10$ renders for target estimation.
One run per row, scored on the $100$ held-out scenes against the converged reference; \%~RMSE, PSNR, and SSIM are linear, the other two columns $\log_{10}$.}
\label{tab:exp2-debias}
\centering
\scriptsize
\setlength{\tabcolsep}{2pt}
\begin{tabular*}{\textwidth}{@{\extracolsep{\fill}}llccccc@{}}
    \toprule
    Arch. & Loss / target & \%~RMSE $\downarrow$ & $\log_{10}$ rel.~$L_2$ $\downarrow$ & PSNR $\uparrow$ & SSIM $\uparrow$ & $\log_{10}$ SSIM $\uparrow$ \\
    \midrule
    NO   & log MSE                       & 56.23          & 1.357          & 19.23          & 0.561          & 0.474          \\ %
    NO   & Log-mean MSE ($K{=}10$)       & 25.44          & 0.489          & 27.29          & 0.848          & 0.787          \\ %
    NO   & 2nd-moment MSE ($K{=}10$)     & 23.21          & 0.456          & 28.18          & 0.861          & 0.797          \\ %
    NO   & Taylor-fallback MSE ($5{+}5$) & 28.39          & 0.516          & 25.10          & 0.829          & 0.735          \\ %
    NO, $10^{\NO}$ head & \loss                & \textbf{17.17} & \textbf{0.083} & \textbf{29.53} & \textbf{0.942} & \textbf{0.914} \\ %
    \bottomrule
\end{tabular*}
\end{table}

Each explicit debiaser cuts the naive-log \%~RMSE by $50$--$59\%$, but none achieves performance comparable to training the model with our proposed loss, even though they require extra renders.
The gap is widest in $\log_{10}$ rel.~$L_2$, where ours reaches $0.083$ versus $0.456$ for the next best (2nd-moment MSE, $K{=}10$).

\begin{figure}[!htb]
    \centering
  \begingroup%
  \catcode`\_=8\relax%
  \edef\pgfincludename{\detokenize{fig_exp2_debias_log_cases}}%
  \graphicspath{{fig/\pgfincludename/}{fig/}}%
  \setkeys{Gin}{draft=false}%
  \resizebox{\linewidth}{!}{\input{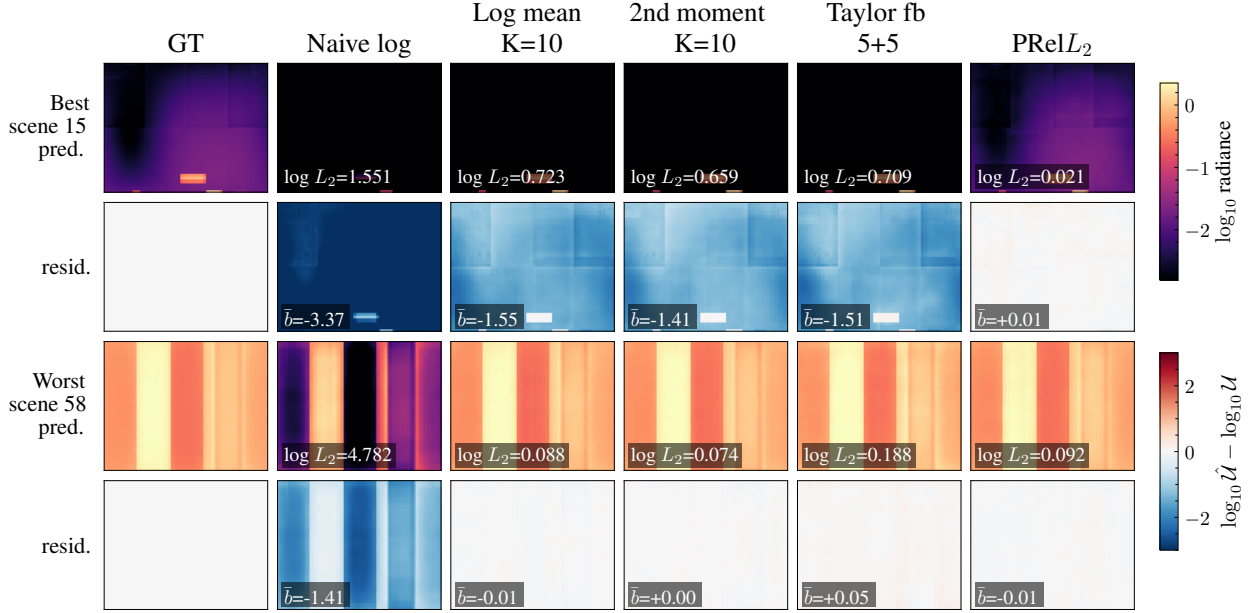}}%
  \endgroup%

    \caption{\textbf{The explicit debiasers reduce the log-MSE offset but keep structured errors; \loss keeps the residual centered.} Best- and worst-case predictions and residuals per debiasing strategy on the $100$ held-out far-field scenes, using the earlier $10^{\NO}$-head model for the \loss column as in Table~\ref{tab:exp2-debias}.
    Rows are selected by $\log_{10}$ rel.~$L_2$: the best block is the lowest-error scene under our loss, and the worst block is the highest-error scene under \emph{log MSE}.
    Prediction rows use one shared $\log_{10}$ radiance scale; residual rows show the $\log_{10}$ prediction minus $\log_{10}\sol(\param)$ with one symmetric $\pm3$ orders-of-magnitude scale.
    Log MSE under-predicts the bright lobes uniformly; the explicit debiasers reduce the offset but retain structured errors; \loss keeps the residual centered near zero without large-area bias.}
    \label{fig:exp2-debias-log-cases}
\end{figure}

\section{Bias Correction for Target-Denominator rel\texorpdfstring{$L_2$}{L2}}
\label{app:inverse-mc}

\begin{figure}[!htb]
    \centering
  \begingroup%
  \catcode`\_=8\relax%
  \edef\pgfincludename{\detokenize{fig_inverse_mc_bias}}%
  \graphicspath{{fig/\pgfincludename/}{fig/}}%
  \setkeys{Gin}{draft=false}%
  \resizebox{\linewidth}{!}{\input{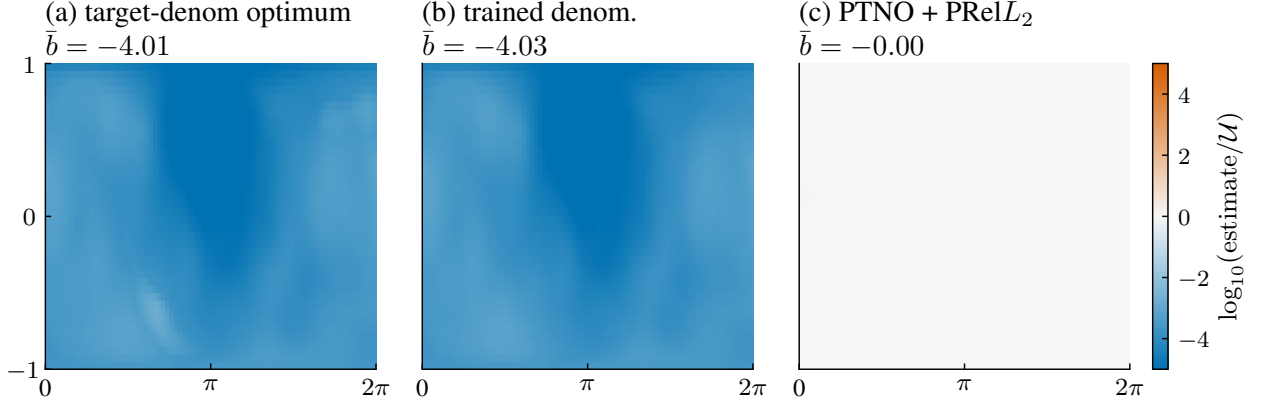}}%
  \endgroup%

    \caption{\textbf{With the noisy label in the denominator, the relative $L_2$ loss collapses the prediction by several orders of magnitude; \ourNO + \loss does not.} One far-field scene with high dynamic range; both models are trained on this scene alone ($M{=}1$) and scored against its converged reference.
    Each panel shows $\log_{10}$ estimate minus $\log_{10}\sol(\param,\sPos)$.
    (a)~The exact stationary point of the baseline rel$L_2$ objective, $((B{-}Y)/(Y{+}\eta))^2$, uses $1/(Y{+}\eta)^2$ weights and collapses the prediction by several orders of magnitude.
    (b)~A trained baseline rel$L_2$ NO converges to the same collapsed field (rel.~$L_2{\approx}1.000$).
    (c)~The matched \ourNO + \loss run is centered at the physical solution (rel.~$L_2{=}0.0067$).
    }
    \label{fig:inverse-mc-bias}
\end{figure}

Figure~\ref{fig:inverse-mc-bias} and Table~\ref{tab:inverse-mc-v2} study how Jensen bias affects relative-$L_2$ training on the far-field radiance dataset.
The baseline rel$L_2$ objective uses noisy MC labels in the denominator, so the nonlinear reciprocal-square map moves the empirical minimizer away from the physical solution.
We compare that baseline with two rel$L_2$ bias-correction strategies that estimate the inverse-square scale from extra independent renders, and with our pointwise relative loss.
The bias-corrected rel$L_2$ variants reduce the collapse but still land far worse than \ourNO + \loss.
This appendix documents the full definitions of the corresponding losses.

\begin{table}[!htb]
\centering
\small
\caption{\textbf{Bias correction improves the label-denominator relative $L_2$ loss but leaves it at least $5\times$ worse than \ourNO.} We compare four loss variants on the far-field radiance dataset at $(M,N,K)=(10^6,4,10)$ and evaluate on the $100$ held-out scenes against the converged reference.
The variants are: (1) the baseline relative ($L_2$) loss, which uses MC labels in the denominator; (2--3) relative ($L_2$) with bias-correction strategies described in \autoref{app:inverse-mc}; and (4) \ourNO with our prediction-normalized pointwise relative loss.
Bias correction improves over the biased baseline but remains at least $5\times$ worse than our method in relative $L_2$.
Mean $\pm$ standard deviation over three seeds.}
\label{tab:inverse-mc-v2}
\setlength{\tabcolsep}{4pt}
\begin{tabular}{llccc}
\toprule
Objective & Bias Correction & rel.~$L_2\downarrow$ & $\log_{10}$ rel.~$L_2\downarrow$ & $\log_{10}$ SSIM$\uparrow$ \\
\midrule
rel$L_2$ & -- & $1.0000\pm0.0000$ & $5.0226\pm0.0000$ & $0.1347\pm0.0000$ \\ %
rel$L_2$ & 2nd-order approx. Taylor & $0.7230\pm0.0011$ & $0.6571\pm0.0009$ & $0.4393\pm0.0052$ \\ %
rel$L_2$ & Taylor & $0.7188\pm0.0028$ & $0.6486\pm0.0028$ & $0.4413\pm0.0025$ \\ %
 \loss (ours) & -- & $\mathbf{0.1304\pm0.0031}$ & $\mathbf{0.0624\pm0.0008}$ & $\mathbf{0.9454\pm0.0019}$ \\ %
\bottomrule
\end{tabular}
\end{table}

\paragraph{Why the label denominator cannot be repaired.}
The weak labels put a point mass at zero, so the inverse-square weight of the label-denominator loss has no finite expectation (\autoref{tab:zero-mass}).
About $15\%$ of the pixels of a ten-render mean are exactly zero, and their weights reach ${\sim}10^{12}$ against ${\sim}2\times10^2$ for a typical (median) pixel of the same label (${\sim}2\times10^4$ for a single render).
The loss then reduces to reproducing zeros where the label is zero, the optimizer drives the whole field to the softplus floor, and the vanishing gradient there prevents recovery, which is the first-epoch collapse behind the baseline row of \autoref{tab:inverse-mc-v2} and panel~(b) of \autoref{fig:inverse-mc-bias}.
Any repair, such as a label floor or a mask on zero pixels, changes the target rather than removing the bias.

\begin{table}[!htb]
\caption{\textbf{Weak labels have a point mass at zero, so the label-denominator weight $1/(Y+\eta)^2$ has no finite mean.} $100$ training scenes of the far-field radiance corpus, $\eta=10^{-6}$; fractions are means over scenes (medians in parentheses), and the maximum pixel weight is each scene's maximum averaged over scenes.}
\label{tab:zero-mass}
\centering
\footnotesize
\begin{tabular}{lccc}
    \toprule
    Label & Pixels exactly zero & Pixels below $\eta$ & Max pixel weight \\
    \midrule
    One 4-SPP render              & $0.326$ ($0.140$) & $0.392$ & $9.6\times10^{11}$ \\
    Mean of ten renders (40 SPP)  & $0.148$ ($0.009$) & $0.201$ & $8.0\times10^{11}$ \\
    \bottomrule
\end{tabular}
\end{table}

\paragraph{Implementation of the rel$L_2$ bias corrections.}
\ourNO + \loss uses a $10$-render target mean \(\bar Y_{10}\), i.e., a $40$-SPP effective target, but its denominator is the prediction:
\[
    \ell_{\mathrm{sg}}
    = \left(\frac{B_\theta-\bar Y_{10}}
                  {\max(\sg(B_\theta),\eta)}\right)^2,
    \qquad B_\theta=\softplus(\NO(\param)).
\]
Conditioned on the prediction, the denominator is a constant and the stochastic gradient remains centered at the physical solution.
The baseline rel$L_2$ objective instead puts the same noisy target in the scale,
\[
    \ell_{\mathrm{tgt}}
    = \left(\frac{B-\bar Y_{10}}{\sg(\bar Y_{10})+\eta}\right)^2.
\]
For a scalar prediction $B$, its stationary point is
\[
    B^*
    =
    \frac{\E\!\left[\bar Y_{10}/(\bar Y_{10}+\eta)^2\right]}
         {\E\!\left[1/(\bar Y_{10}+\eta)^2\right]},
\]
which is an inverse-MC-weighted target rather than $\E[\bar Y_{10}]$.

The two rel$L_2$ bias corrections try to repair this by keeping the residual render independent from the denominator estimate.
They train
\[
    \ell_{\mathrm{iw}}
    = \widehat W\,(B_\theta-Y_r)^2,
\]
where $Y_r$ is one selected $4$-SPP residual render and $\widehat W$ is computed from disjoint renders.
If $\widehat W$ were a stable low-variance estimate of $(\E[Y]+\eta)^{-2}$, independence would recover the correct fixed point.
The second-order bias correction uses $K{=}10$ anchor renders with
\[
    \bar A_K=\frac{1}{K}\sum_{i=1}^{K}A_i,
    \qquad
    \hat\mu=\max(\bar A_K,\eta),
    \qquad
    \widehat\sigma_A^2
    =
    \frac{1}{K-1}\sum_{i=1}^{K}(A_i-\bar A_K)^2 .
\]
For the weight \(w(x)=(x+\eta)^{-2}\), with \(\mu=\E[A_i]\) and \(\sigma^2=\Var(A_i)\), the delta method gives
\[
    \E[w(\bar A_K)]
    \approx
    w(\mu)+\frac{1}{2}w''(\mu)\frac{\sigma^2}{K}
    =
    (\mu+\eta)^{-2}
    +
    \frac{3\sigma^2}{K(\mu+\eta)^4},
\]
so the inverse-square analog of the second-moment log correction subtracts this leading plug-in bias:
\[
    \widehat W_2
    =
    \left[
        (\hat\mu+\eta)^{-2}
        -
        \frac{3\widehat\sigma_A^2}{K(\hat\mu+\eta)^4}
    \right]_+ .
\]
The Taylor bias correction splits its $10$ renders into $5$ anchor renders $A_i$ and $5$ correction renders $A'_j$.
The anchors give a stable expansion point $\hat\mu=\max(\frac{1}{5}\sum_i A_i,\eta)$, and each correction render contributes a mean-zero relative residual $r_j=(A'_j-\hat\mu)/(\hat\mu+\eta)$, which is an unbiased estimator of $(\mu-\hat\mu)/(\hat\mu+\eta)$.
Expanding the inverse-square weight $(\mu+\eta)^{-2}$ around $\hat\mu$ gives
\[
    (\mu+\eta)^{-2}
    =
    (\hat\mu+\eta)^{-2}
    \sum_{m\ge0}(-1)^m(m+1)
    \left(\frac{\mu-\hat\mu}{\hat\mu+\eta}\right)^m .
\]
Because the $r_j$ are independent, the product $\prod_{j=1}^m r_j$ is an unbiased estimator of the $m$-th power, so plugging products of fresh residuals into successive series terms preserves unbiasedness.
We truncate at $m{=}5$ (one product per correction render) to obtain the plug-in weight
\[
    \widehat W_{\mathrm{T}}
    =
    (\hat\mu+\eta)^{-2}
    \left[
        1+\sum_{m=1}^{5}(-1)^m(m+1)\prod_{j=1}^{m}r_j
    \right].
\]
Pixels with $\max_j |r_j|>0.95$ fall back to the second-order weight; remaining non-finite values are zeroed and the final weights are clamped nonnegative.
The raw Taylor--Russian-roulette diagnostic uses the same inverse-power series with $8$ roulette paths and continuation $q_m=\min(\overline{|r|}\,m/(m+1),0.95)$, but it produced non-finite weights on the first batch and is therefore not reported as a successful training run.

As with the log-target debiasers in the previous section, the $K$ extra renders are amortized over the training set, not the stochastic gradient.
The bias corrections reduce the most direct denominator-target coupling, but the corrected rel$L_2$ objectives still concentrate optimization on unstable inverse-scale weights and do not recover the prediction-normalized \loss solution.
Per-run numbers are tabulated in Table~\ref{tab:inverse-mc-v2}.

\begin{figure}[!htb]
    \centering
  \begingroup%
  \catcode`\_=8\relax%
  \edef\pgfincludename{\detokenize{fig_inverse_mc_cases}}%
  \graphicspath{{fig/\pgfincludename/}{fig/}}%
  \setkeys{Gin}{draft=false}%
  \resizebox{\linewidth}{!}{\input{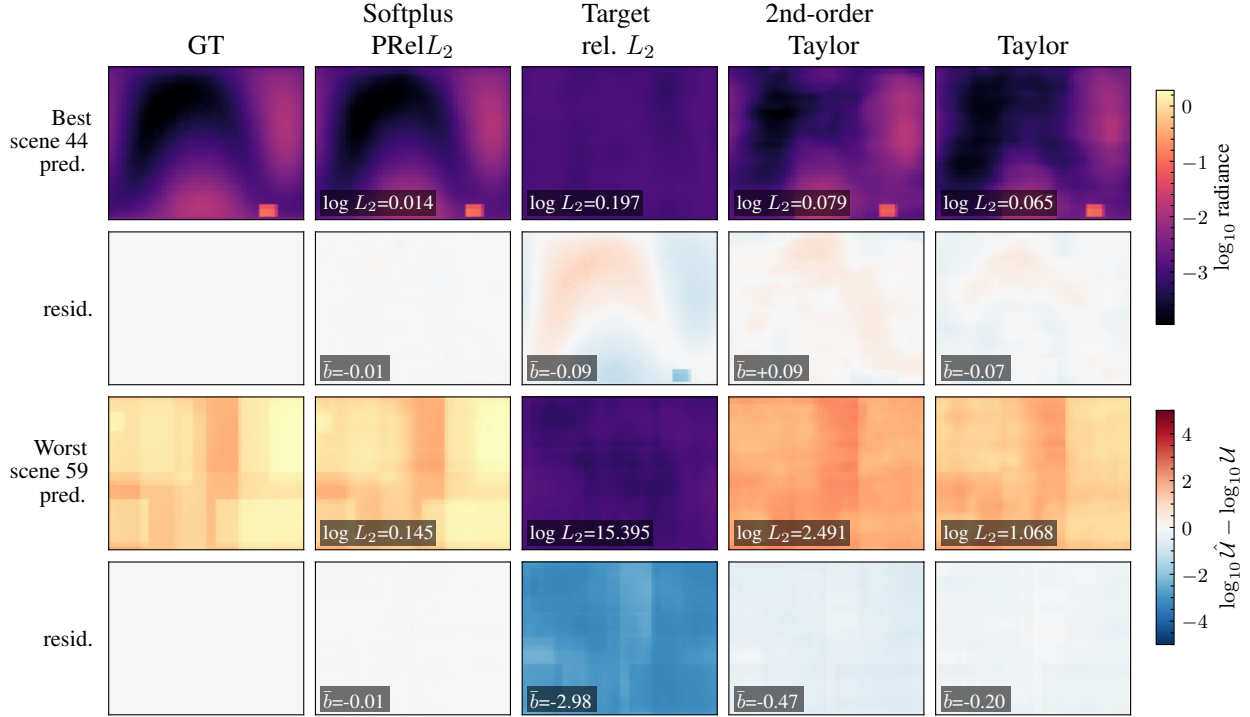}}%
  \endgroup%

    \caption{\textbf{The label-denominator and bias-corrected rel$L_2$ variants stay biased downward on the hard scene, while \loss keeps the residual centered.} Best- and worst-case far-field predictions for the rel$L_2$ bias-correction study in Table~\ref{tab:inverse-mc-v2}.
    The best block is the lowest $\log_{10}$ rel.~$L_2$ scene under \ourNO + \loss, and the worst block is the highest-error scene under baseline rel$L_2$.
    Prediction rows use one shared $\log_{10}$ radiance scale; residual rows use one symmetric $\pm5$ orders-of-magnitude scale.
    The baseline and bias-corrected rel$L_2$ variants remain biased downward on the hard scene, whereas the prediction-normalized \loss keeps the residual centered.}
    \label{fig:inverse-mc-cases}
\end{figure}

\section{Stationary Points and Dynamics of \loss}
\label{app:prel2-analysis}

\loss has no stationary point other than the population mean of the label wherever the output Jacobian has full row rank, and its expected dynamics decrease a potential monotonically.
The minimizer-equivalence argument of \autoref{sec:method} covers only the unweighted squared loss, so we analyze \loss, which normalizes each residual by the stop-gradient prediction, directly.
Fix an input $a$ and let $B_\theta=\SCNO(a)$ denote the network prediction, $U=\expect{\xi}{\MC_N(a,\xi)\mid a}$ the population mean of the noisy target, $J=\partial B_\theta/\partial\theta$ the output Jacobian, and $D(B_\theta)$ the positive diagonal matrix of stop-gradient normalization factors, with entries $\max(B_{\theta,i},\eta)^2$.

\paragraph{Fixed points.}
Because the normalization is detached, the expected parameter gradient is, up to a positive constant,
\begin{equation}
    \mathbb{E}_\xi[\nabla_\theta\ell\mid a] = J^\top D(B_\theta)^{-1}(B_\theta - U).
    \label{eq:prel2-expected-grad}
\end{equation}
Any stationary point therefore satisfies $J^\top D(B_\theta)^{-1}(B_\theta - U)=0$.
Since $D(B_\theta)$ is positive definite, the normalization introduces no additional zero directions; under the standard local assumption that $J$ has full row rank ($JJ^\top\succ0$), stationarity forces $B_\theta=U$.
\loss therefore introduces no new biased output-space fixed point.
Rank-deficient Jacobians can create parameter-space stationary points, a limitation shared with the ordinary squared loss.

\paragraph{Dynamics.}
Define the potential
\begin{equation*}
    \Phi(B_\theta) = \sum_i \int_{U_i}^{B_{\theta,i}} \frac{s - U_i}{\max(s,\eta)^2}\,\dif s,
\end{equation*}
where $\max(s,\eta)^2>0$ is the prediction-dependent normalization factor at voxel $i$.
Its output-space gradient is $\nabla_B\Phi(B_\theta)=D(B_\theta)^{-1}(B_\theta-U)$, which matches the expected \loss gradient with respect to the prediction.
The corresponding dynamics are obtained through the network Jacobian, $\dot\theta=-J^\top\nabla_B\Phi$, and the potential evolves as
\begin{equation*}
    \frac{\dif\Phi}{\dif t} = -\nabla_B\Phi^\top JJ^\top\nabla_B\Phi \le 0 .
\end{equation*}
The expected dynamics therefore decrease $\Phi$ monotonically.
The prediction-dependent denominator may change the effective learning rate and the convergence speed of different voxels, but it does not reverse the update direction or create an additional output-space minimum.
Under $JJ^\top\succ0$, the only point at which the dynamics stop is $B_\theta=U$; when $U_i=0$ exactly, softplus approaches this solution as a boundary limit rather than attaining it at a finite latent value.

\paragraph{Scope.}
Monotonic decrease of the potential alone does not guarantee convergence to $B_\theta=U$.
When $J$ is rank-deficient, the normalized residual may lie in the null space of $J^\top$, so additional stationary points cannot be excluded.
This limitation arises from optimizing an output-space objective through a potentially rank-deficient neural-network parameterization rather than from \loss specifically: standard squared-error training has the stationary condition $J^\top(B_\theta-U)=0$.
The analysis concerns continuous-time, full-batch population-gradient dynamics and does not directly establish convergence for finite-step stochastic optimization or Adam.

\paragraph{Empirical test across initializations.}
Five independent initializations of \ourNO converge to the same unbiased solution, whereas log MSE converges to a shared biased one (\autoref{tab:fixed-point}).
We split each seed's $\log_{10}$ error field $d_s=\log_{10}B_s-\log_{10}U$ into the part shared by all seeds, $\bar d$, and the seed-specific remainder, and report the signed mean of $d_s$ as a multiplicative offset.
Log MSE is the positive control: $99.9\%$ of its squared error is shared across seeds, and every seed predicts $2.2\%$ of the reference, ${\approx}46\times$ too low.
\ourNO recovers the reference to within $0.1\%$ on average.
Its remaining error is also mostly shared, but it carries no sign and shrinks with scene coverage: across the label-budget allocations of \autoref{tab:label-allocation}, the signed offset stays within $\pm0.07$ orders of magnitude while the RMS $\log_{10}$ error falls from $0.77$ to $0.13$.
A fixed point would not move with $M$; approximation error does.
The offset also holds across brightness: \ourNO stays within a factor of $0.972$--$1.034$ in every bin of \autoref{tab:exp1}, while log MSE is $617\times$ too low in the dimmest bin.

\begin{table}[!htb]
\caption{\textbf{\ourNO seeds agree on an unbiased solution; log-MSE seeds agree on a biased one.} Far-field radiance, $(M,N,K)=(10^6,4,1)$, five seeds, $100$ held-out scenes against the converged reference.
Offset is $10^{\mathrm{mean}(d_s)}$, mean $\pm$ standard deviation across seeds.}
\label{tab:fixed-point}
\centering
\footnotesize
\begin{tabular*}{\textwidth}{@{\extracolsep{\fill}}lcccc@{}}
    \toprule
    Loss & RMS $\log_{10}$ error & Shared RMS & Seed-specific RMS & Prediction / reference \\
    \midrule
    NO + $L_2$            & $1.1032$ & $0.7982$ & $0.7615$ & $0.525\pm0.060$ \\
    NO + log MSE          & $2.1337$ & $2.1325$ & $0.0721$ & $0.0215\pm0.0000$ \\
    \ourNO + \loss (ours) & $\mathbf{0.1518}$ & $\mathbf{0.1394}$ & $\mathbf{0.0602}$ & $\mathbf{0.999\pm0.005}$ \\
    \bottomrule
\end{tabular*}
\end{table}

\section{Controlled Ablations and Label-Budget Allocation}
\label{app:controlled-ablation}

The five-seed loss comparison on all four datasets is \autoref{tab:main-results}.
On far-field radiance, the linear-$L_2$ loss wins the bright-region-dominated \%~RMSE ($15.27\pm1.32$ against $16.43\pm0.55$ for \ourNO) at the cost of a $0.26$ lower $\log_{10}$ SSIM, which is lost in the dim regions.

\subsection{Allocating a Fixed Label Budget}

Scene coverage controls accuracy until the training set reaches roughly $5\times10^5$ configurations, consistent with walk-on-spheres operator learning and Monte Carlo surrogate models~\citep{viswanath2026operatorlearningusingweak, pratt2026moment}; \autoref{tab:m-scaling} shows that how much coverage helps depends on the loss.
Table~\ref{tab:label-allocation} fixes the raw label budget at $M(N/4)K=10^6$ four-SPP render equivalents and the training budget at $10^5$ optimizer updates ($200$ epochs of $5{,}000$ samples), and changes only how the renders are allocated.
For $N\ge8$, repeated four-SPP renders are averaged into one target; the $K>1$ rows cycle those same raw renders across visits.

\begin{table}[!htb]
\caption{\textbf{At a fixed label budget, spreading the renders over more scenes helps until about $5\times10^5$ scenes.} Far-field radiance.
The first block increases scene coverage while reducing the effective samples per scene.
The second block cycles repeated four-SPP renders instead of pre-averaging them.
Entries are $\log_{10}$ SSIM on the $100$ held-out scenes at $40{\times}80$ against the converged reference, mean $\pm$ standard deviation over the listed seeds; entries without an uncertainty are single-seed measurements.}
\label{tab:label-allocation}
\centering
\footnotesize
\setlength{\tabcolsep}{5pt}
\begin{tabular}{rrrrc}
    \toprule
    Scenes $M$ & Effective $N$ & Renders $K$ & Seeds & $\log_{10}$ SSIM $\uparrow$ \\
    \midrule
    $125$       & $32{,}000$ & $1$ & $3$ & $0.2716\pm0.0098$ \\
    $1{,}000$   & $4{,}000$  & $1$ & $3$ & $0.4829\pm0.0043$ \\
    $8{,}000$   & $500$      & $1$ & $3$ & $0.7581\pm0.0019$ \\
    $31{,}250$  & $128$      & $1$ & $3$ & $0.8450\pm0.0026$ \\
    $62{,}500$  & $64$       & $1$ & $3$ & $0.8716\pm0.0018$ \\
    $125{,}000$ & $32$       & $1$ & $1$ & $0.9009$ \\
    $250{,}000$ & $16$       & $1$ & $3$ & $0.9191\pm0.0011$ \\
    $500{,}000$ & $8$        & $1$ & $3$ & $\mathbf{0.9234\pm0.0030}$ \\
    $1{,}000{,}000$ & $4$    & $1$ & $5$ & $0.9165\pm0.0040$ \\
    \midrule
    $31{,}250$  & $4$        & $32$ & $3$ & $0.9096\pm0.0049$ \\
    $125{,}000$ & $4$        & $8$  & $3$ & $0.9143\pm0.0019$ \\
    $500{,}000$ & $4$        & $2$  & $1$ & $0.9196$ \\
    \bottomrule
\end{tabular}
\end{table}

Cycling raises $\log_{10}$ SSIM by $0.0646$ at $M=31{,}250$ and by $0.0134$ at $M=125{,}000$; the difference is $-0.0038$ at $M=500{,}000$.
Repeated labels help before the coverage plateau and cease to matter once the corpus is broad enough.

Broad noisy supervision beats the converged-label control with one twenty-fifth of the particle budget.
Training \ourNO on $10^6$ scenes at four SPP reaches $0.9165\pm0.0040$ with a total budget of $4\times10^6$ samples, while $10^4$ scenes at $10^4$ SPP reach $0.8893\pm0.0071$ with $10^8$ samples, so spreading the particles over more scenes gains $0.0272$ in $\log_{10}$ SSIM.
The $10^4$-SPP control labels are close to converged: from the sample variance of $32$ independent four-SPP renders per scene, a $10^4$-SPP label has a relative $L_2$ error of $1.9\%$ (median $1.2\%$), against $94\%$ for a single four-SPP label.

\ourNO also wins on converged labels, so \loss is not only a fix for label noise (\autoref{tab:ctrl-2x2}).
At $M{=}10^4$, replacing linear $L_2$ with \ourNO raises $\log_{10}$ SSIM by $0.26$ on $10^4$-SPP labels and by $0.20$ on four-SPP labels.
Converged labels help both losses at this small $M$; \autoref{tab:label-allocation} shows that the same particles buy more when spread over more scenes.

\begin{table}[!htb]
\caption{\textbf{\ourNO + \loss beats linear $L_2$ on both near-converged and noisy labels.} Loss $\times$ label quality at $M{=}10^4$, far-field radiance.
The two label columns share the same $10^4$ scenes.
Entries are $\log_{10}$ SSIM on the $100$ held-out scenes, mean $\pm$ standard deviation over three seeds.}
\label{tab:ctrl-2x2}
\centering
\footnotesize
\begin{tabular}{lcc}
    \toprule
    Loss & $N{=}10^4$ labels & $N{=}4$ labels \\
    \midrule
    NO + $L_2$            & $0.6257\pm0.0060$ & $0.3468\pm0.0069$ \\
    \ourNO + \loss (ours) & $\mathbf{0.8893\pm0.0071}$ & $\mathbf{0.5478\pm0.0043}$ \\
    \bottomrule
\end{tabular}
\end{table}

\subsection{More Scenes Cannot Remove a Bias}

\ourNO keeps improving with scene count, while the biased losses flatten (\autoref{tab:m-scaling}).
Holding $N{=}4$ and $K{=}1$ and growing $M$ from $10^5$ to $10^6$ raises \ourNO by $0.14$ in $\log_{10}$ SSIM, against $0.02$ for linear $L_2$ and $0.01$ for log MSE.
Extra data reduces variance, but the minimizer of a biased objective does not move toward the solution.

\begin{table}[!htb]
\caption{\textbf{With more scenes, \ourNO keeps improving while the biased losses flatten.} Scene scaling at fixed label quality, far-field radiance, $(M,N,K)=(M,4,1)$.
Entries are $\log_{10}$ SSIM on the $100$ held-out scenes, mean $\pm$ standard deviation over three seeds (five at $M{=}10^6$).}
\label{tab:m-scaling}
\centering
\footnotesize
\begin{tabular}{rccc}
    \toprule
    $M$ & NO + $L_2$ & NO + log MSE & \ourNO + \loss \\
    \midrule
    $10^4$ & $0.3468\pm0.0069$ & $0.2391\pm0.0034$ & $\mathbf{0.5478\pm0.0043}$ \\
    $10^5$ & $0.6316\pm0.0055$ & $0.4562\pm0.0013$ & $\mathbf{0.7815\pm0.0028}$ \\
    $10^6$ & $0.6524\pm0.0196$ & $0.4699\pm0.0007$ & $\mathbf{0.9165\pm0.0040}$ \\
    \bottomrule
\end{tabular}
\end{table}

\subsection{Splitting a Per-Scene Budget between \texorpdfstring{$N$ and $K$}{N and K}}

At a fixed per-scene budget, cycling several cheap labels beats averaging them into one (\autoref{tab:nk-split}).
We fix $M{=}31{,}250$ and $NK{=}128$ and vary the split: each scene stores $K$ renders of $N$ SPP, and training presents one of them per visit.
Moving from one $128$-SPP label to eight $16$-SPP labels raises $\log_{10}$ SSIM from $0.8450$ to $0.9111$.
Accuracy saturates around $K{=}8$--$16$: $K{=}16$ gives the best score, a further gain of $0.0070$, while $K{=}32$ falls back to $0.9096$.
A pre-averaged label fixes one noise realization that the network can fit, whereas cycling presents a fresh unbiased draw at every visit.

\begin{table}[!htb]
\caption{\textbf{At a fixed per-scene budget, cycling several cheap labels beats one averaged label, up to about $K=8$--$16$.} Per-scene $N$--$K$ split at $M{=}31{,}250$ and $NK{=}128$, far-field radiance.
Entries are $\log_{10}$ SSIM on the $100$ held-out scenes, mean $\pm$ standard deviation over three seeds.}
\label{tab:nk-split}
\centering
\footnotesize
\begin{tabular}{rrc}
    \toprule
    $N$ & $K$ & $\log_{10}$ SSIM $\uparrow$ \\
    \midrule
    $128$ & $1$  & $0.8450\pm0.0026$ \\
    $64$  & $2$  & $0.8683\pm0.0021$ \\
    $32$  & $4$  & $0.8922\pm0.0024$ \\
    $16$  & $8$  & $0.9111\pm0.0022$ \\
    $8$   & $16$ & $\mathbf{0.9181\pm0.0033}$ \\
    $4$   & $32$ & $0.9096\pm0.0049$ \\
    \bottomrule
\end{tabular}
\end{table}

\subsection{Finite-Sample View of the Allocation}
\label{app:finite-sample}

This subsection states which terms scale with $M$, $N$, and $K$, and what is proved for each loss.
Scene $a_i$, $i=1,\ldots,M$, carries $K$ independent renders $Y_{i,j}=\sol(a_i)+\error_{i,j}$, each averaging $N$ Monte Carlo samples, with $\mathbb E[\error_{i,j}\mid a_i]=0$ and per-cell variance $\sigma^2/N$, where $\sigma^2$ is the single-sample variance.

\paragraph{Renders enter through their mean.}
For any per-cell weight $w$ that does not depend on the labels ($w=1$ for the squared loss, $w=\DENO(\SCNO(a_i))^{-2}$ for \loss with the normalizer detached),
\begin{equation*}
    \frac{1}{K}\sum_{j=1}^{K} w\,(B-Y_{i,j})^2 = w\,(B-\bar Y_i)^2 + \frac{w}{K}\sum_{j=1}^{K}(Y_{i,j}-\bar Y_i)^2,
    \qquad \bar Y_i=\frac{1}{K}\sum_{j=1}^{K}Y_{i,j},
\end{equation*}
and the last term does not depend on the prediction $B$.
The empirical objective and its full-batch gradient therefore see the $K$ renders only through $\bar Y_i$, whose noise has variance $\sigma^2/(NK)$: $K$ renders of $N$ samples act as one render of $NK$ samples.
Stochastic training that presents one render per visit is not covered by this identity; \autoref{tab:nk-split} shows that at small $M$ it beats pre-averaging at equal $NK$.

\paragraph{Scenes versus Monte Carlo precision.}
For the unweighted squared loss with these unbiased labels, the risk against the converged solution is the clean objective $\LossC$ of \autoref{sec:method} up to a constant, and for a realizable class $\mathcal F$ of effective dimension $d_{\mathrm{eff}}$ the standard least-squares decomposition gives
\begin{equation*}
    \mathbb{E}\,\mathcal{R}(\hat\theta)-\inf_{\theta\in\Theta}\mathcal{R}(\theta)\;\lesssim\;\mathcal{E}_M(\mathcal{F})+\frac{\sigma^2\,d_{\mathrm{eff}}}{MNK},
\end{equation*}
where $\mathcal E_M(\mathcal F)$ is the excess risk the same estimator would reach from $M$ converged labels.
$\mathcal E_M$ depends on $M$ and $\mathcal F$ but not on $N$ or $K$; the label noise contributes only through $1/(MNK)$, so scene coverage reduces the risk at any label precision, and there is no term in $1/N$ or $1/K$ alone.

\paragraph{Nonlinear label transforms.}
If the loss compares the prediction with a transformed label $\log_{10}Y$, the population minimizer is $\mathbb E[\log_{10}Y]$, and the delta method gives $\mathbb E[\log_{10}Y]=\log_{10}\sol-\sigma^2/(2n\,\sol^2\ln10)+O(n^{-2})$ per cell, with $n=N$ when each render is transformed and cycled and $n=NK$ when the $K$ renders are averaged first.
The fitted prediction is shifted by $O(1/N)$ (respectively $O(1/(NK))$), which adds a floor of order $1/N^2$ (respectively $1/(NK)^2$) to the risk against $\log_{10}\sol$; no number of scenes removes it, and cycling more renders does not reduce it.
The label-denominator relative loss has the analogous shift, and with a point mass of labels at zero its weight has no finite mean (Appendix~\ref{app:inverse-mc}).

\paragraph{What is shown for \loss.}
\autoref{app:prel2-analysis} shows that, in expectation over the Monte Carlo noise, the \loss gradient has no output-space stationary point other than the label mean $\sol(a)$ wherever the output Jacobian has full row rank, so the Jensen shift above does not arise.
The averaging identity also holds for \loss at fixed normalizers.
The variance bound above is proved only for the unweighted squared loss; because the \loss weights depend on the prediction, we do not claim it for \loss.

\section{Cost Accounting and Speedup}
\label{app:cost}

The EU-DEMO accuracy crossings below use the final checkpoints.
The other tasks retain their historical accuracy crossings and timings pending the corresponding reruns; their checkpoint limitations are recorded in Appendix~\ref{app:eval-protocol}.

The surrogate repays its offline cost after $12$--$197$ queries, and at matched accuracy it is $10^3$--$10^5\times$ cheaper than MC on neutronics, while on radiative transfer MC costs $0.8$--$11\times$ the operator.
\autoref{tab:speedup} summarizes the comparison; the paragraphs below give the offline cost, the same-device comparison against converged MC, and the matched-accuracy measurement.

\begin{table}[!htb]
\caption{\textbf{At matched accuracy, MC costs $10^3$--$10^5\times$ the operator on neutronics and $0.8$--$11\times$ on radiative transfer.} Both methods run on the same device (one GPU for radiative transfer, one 24-thread CPU for neutronics).
Matched-accuracy entries are per-scene medians of the MC cost needed to reach the operator's error, divided by the operator's cost; EU-DEMO MC uses FW-CADIS weight windows with their generation cost included. ${}^{\ast}$MC does not reach the operator's $\log_{10}$ SSIM within the measured budget on most scenes; the spherical-tokamak entry extrapolates per-scene error curves by at most one order of magnitude, and the EU-DEMO entry is a lower bound.}
\label{tab:speedup}
\centering
\footnotesize
\begin{tabular*}{\textwidth}{@{\extracolsep{\fill}}lcccc@{}}
    \toprule
    & \multicolumn{3}{c}{MC cost $/$ operator cost} & \\
    \cmidrule(lr){2-4}
    Dataset & Converged MC & Match $\log_{10}$ SSIM & Match $\log_{10}$ rel.~$L_2$ & Break-even queries \\
    \midrule
    Far-field radiance & $1.8\times10^{5}$ & $11$ & $5.5$ & $12$ \\
    3D fluence         & $1.8\times10^{4}$ & $2.1$ & $0.8$ & $87$ \\
    Sph.~tokamak       & $9.9\times10^{4}$ & $6.4\times10^{4}\,{}^{\ast}$ & $9.4\times10^{3}$ & $197$ \\ %
    EU-DEMO            & $1.2\times10^{4}$ & $>1.6\times10^{4}\,{}^{\ast}$ & $1.6\times10^{3}$ & $12$ \\
    \bottomrule
\end{tabular*}
\end{table}

\paragraph{Offline cost and break-even.}
Converged labels for our training sets would cost $5\times10^3$--$10^5\times$ more than the noisy labels we train on (\autoref{tab:offline-cost}, ``Converged / noisy'').
Break-even divides the offline cost, label generation plus training, by the cost of one converged solve minus one inference.
For the neutronics tasks, GPU training time is small next to CPU label generation and is listed separately rather than converted to core-hours.
Labeling the EU-DEMO training set at converged quality would take ${\approx}6\times10^8$ core-hours.

\begin{table}[!htb]
\caption{\textbf{Offline cost is repaid after tens to a few hundred queries.} Label costs were measured on H200 GPUs (radiative transfer), in 8-core tasks on Xeon Gold 6130 CPUs (spherical tokamak), and in 8-core tasks on Xeon Ice Lake and Skylake nodes (EU-DEMO); training costs are historical five-run means for the same recipes on one H200 GPU, retained for cost accounting rather than the runtimes of the replacement runs. The spherical-tokamak converged cost is stated in the 8-core configuration of its training labels; the $10^9$-history reference itself ran as 4-core tasks at $305.5$ core-hours per scene.
Costs are measured from the scheduler accounting; the EU-DEMO converged cost is the mean over the $24$ reference scenes.}
\label{tab:offline-cost}
\centering
\scriptsize
\setlength{\tabcolsep}{3pt}
\begin{tabular*}{\textwidth}{@{\extracolsep{\fill}}lccccc@{}}
    \toprule
    Dataset & Noisy-label set & Training & Converged / scene & Converged / noisy & Break-even \\
    \midrule
    Far-field radiance & $3.79$ GPU-h ($10^6$)             & $0.75$ GPU-h & $0.372$ GPU-h     & $9.8\times10^{4}$ & $12$ \\
    3D fluence         & $1.97$ GPU-h ($3{\times}10^5$)      & $7.98$ GPU-h & $0.115$ GPU-h     & $1.75\times10^{4}$ & $87$ \\
    Sph.~tokamak       & $70{,}300$ core-h ($10^6$) & $1.81$ GPU-h & $356.8$ core-h & $5.2\times10^{3}$ & $197$ \\
    EU-DEMO            & $71{,}607$ core-h ($10^5$)        & $2.88$ GPU-h & $6{,}069$ core-h  & $8.5\times10^{3}$ & $12$ \\
    \bottomrule
\end{tabular*}
\end{table}

\paragraph{Same device, converged MC.}
Reproducing a converged reference costs MC $1.2\times10^4$--$1.8\times10^5\times$ the operator's inference when both run on the same device.
\autoref{fig:paper-setup-compare} times the operator on a GPU and MC on a 24-thread CPU, because OpenMC has no GPU implementation, and that ratio folds a hardware factor of $10$--$60\times$ into the comparison; the numbers here remove it.
On the same device and the reference's own output grid, converged MC costs $1.8\times10^5\times$ (far-field radiance, $607$\,s for $2^{20}$~SPP on the $40{\times}80$ output grid, which reproduces the reference to $\log_{10}$ rel.~$L_2$ $0.0008$, against $3.4$\,ms) and $1.8\times10^4\times$ (3D fluence, $2{,}636$\,s against $144$\,ms) the operator's inference, both methods timed on one RTX~5070 Laptop GPU.
On one 24-thread CPU (Xeon Gold 6130 for the spherical tokamak, Xeon Platinum 8352Y for EU-DEMO), the per-scene median is $9.9\times10^4\times$ for the spherical tokamak ($10^9$ histories against a $0.47$\,s forward pass) and $1.2\times10^4\times$ for EU-DEMO (the full weight-window and transport chain, median $6{,}130$ core-hours, against a $79$\,s forward pass on the $700\times365\times730$ reference grid).

\paragraph{Matched accuracy.}
MC reaches the operator's accuracy cheaply on radiative transfer and only at great cost on neutronics.
For every held-out scene we sweep the MC budget on the same device, score each budget against an independent reference with the metric used for the operator, and record the cost at which MC first matches the operator's error; we report the median over scenes.
On far-field radiance ($100$ scenes), MC matches $\log_{10}$ SSIM at $11.3\times$ the operator's cost (interquartile range $4.5$--$34$) and $\log_{10}$ relative $L_2$ at $5.5\times$ ($3.4$--$10.5$).
On 3D fluence ($100$ scenes) the factors are $2.1\times$ ($1.0$--$9.6$) and $0.8\times$ ($0.44$--$2.2$): MC matches the norm error within the operator's own inference time on half of the scenes.
On the spherical tokamak ($50$ scenes), we split each scene's $100$ independent $10^7$-history runs into a $5\times10^8$-history reference and a sequence of MC budgets that grows to $5\times10^8$ histories.
MC matches $\log_{10}$ relative $L_2$ at a median $9.4\times10^3\times$ (interquartile range $4.9\times10^3$--$2.3\times10^4$); for $\log_{10}$ SSIM, $32$ of $50$ scenes remain unmatched at the largest budget (${\approx}4.7\times10^4\times$), and extrapolating each scene's error curve by at most one order of magnitude gives a median of $6.4\times10^4\times$.

On EU-DEMO, MC matches the operator's norm error only with a weight window, and even with one it misses the operator's $\log_{10}$ SSIM on every one of the $24$ scenes within the measured budgets (\autoref{tab:demo-matched}).
We run three MC variants on $24$ scenes and count the weight-window generation cost where it applies.
Without a window, MC does not match the operator within $10^7$ histories on any scene.
A cheap FW-CADIS window lets MC match $\log_{10}$ relative $L_2$ at $1.6\times10^3\times$ the operator's cost, but all $24$ scenes still miss its $\log_{10}$ SSIM.
The reference's own FW-CADIS$\to$MAGIC window also reaches the norm target, but building it takes ${\approx}2{,}000$ core-hours per scene, which lifts the end-to-end factor to $2.2\times10^5\times$.
On every task, matching the dim-region structure measured by $\log_{10}$ SSIM costs MC more than matching the norm.

\begin{table}[!htb]
\caption{\textbf{EU-DEMO matched-accuracy factors depend on the weight window given to MC.} Per-scene medians over $24$ scenes of MC cost divided by operator cost, both on one Xeon Platinum 8352Y. ``Transport'' counts transport only; ``+ window'' adds the window generation cost. $>$ marks a lower bound where MC stays short of the operator.
All entries are scored only on cells where the reference is measured ($45.4\%$ of the mesh).
The MAGIC SSIM factors are extrapolated; the measured end-to-end lower bound is $2.6\times10^5$.}
\label{tab:demo-matched}
\centering
\scriptsize
\setlength{\tabcolsep}{3pt}
\begin{tabular*}{\textwidth}{@{\extracolsep{\fill}}lccccc@{}}
    \toprule
    & & \multicolumn{2}{c}{Match $\log_{10}$ rel.~$L_2$} & \multicolumn{2}{c}{Match $\log_{10}$ SSIM} \\
    \cmidrule(lr){3-4}\cmidrule(lr){5-6}
    MC variant & Window / scene & Transport & + window & Transport & + window \\
    \midrule
    Analog                 & ---                  & $>1.6\times10^{3}$ & --- & no crossing & --- \\
    FW-CADIS               & $10.2$ core-h        & $5.4\times10^{2}$ & $1.57\times10^{3}$ & unmatched & $>1.6\times10^{4}$ \\
    FW-CADIS$\to$MAGIC     & $2{,}009$ core-h     & $6.8\times10^{2}$ & $2.17\times10^{5}$ & $9.1\times10^{5}$ & $1.1\times10^{6}$ \\
    \bottomrule
\end{tabular*}
\end{table}

For the MAGIC SSIM estimate, we extrapolate each scene's error curve through its last two measured budgets by at most $1.5$ orders of magnitude in particle count.
Twenty scenes cross within this range; the remaining four are counted as lying above this range when computing the median over all $24$ scenes.

\section{Scope and In-Distribution Boundary}
\label{app:scope}

\paragraph{Which transport problems suit \ourNO.}
\ourNO targets transport problems where low-order models fail and converged MC is expensive.
In optically thick media with smooth coefficients, near-isotropic scattering, and clear scale separation, diffusion or low-order moment equations are accurate and a learned surrogate offers little.
Our settings violate these conditions in different ways.
The radiative-transfer datasets sample the Henyey--Greenstein asymmetry over nearly its whole range together with piecewise or shape-wise varying coefficients, so much of the training distribution is strongly anisotropic and has sharp interfaces.
The neutronics settings are deep-shielding problems whose reference flux spans about $10$ (spherical tokamak) and $17$ (EU-DEMO) orders of magnitude (\autoref{tab:dynamic-range}), hard enough that the EU-DEMO reference itself needs FW-CADIS and MAGIC weight windows.
The cost accounting of \autoref{app:cost} gives the practical criterion: the surrogate pays off most where the dim tail of the solution is expensive for MC to converge.

\paragraph{What the training designs vary.}
Every held-out configuration is a fresh draw from the same design as the training set (\autoref{tab:id-boundary}), so all reported accuracy is interpolation within a parameterized family.
The axes behave differently at the boundary.
Coefficient ranges ($\totCS_t$, $c$, $g$) and source parameters form a soft boundary: the solution depends smoothly on them, so a query just outside the sampled range is a mild extrapolation, which we do not evaluate.
Geometry families form a hard boundary: within a family, geometry is parameterized and covered by the design, but a different machine, a new port, or a component absent from training is out of distribution.
For EU-DEMO the family is a single fixed geometry.

\begin{table}[!htb]
\caption{\textbf{Each dataset varies coefficients, sources, or geometry within a fixed family}; held-out sets are fresh draws from the same design.}
\label{tab:id-boundary}
\centering
\scriptsize
\setlength{\tabcolsep}{3pt}
\begin{tabular*}{\textwidth}{@{\extracolsep{\fill}}>{\raggedright\arraybackslash}p{0.15\textwidth}>{\raggedright\arraybackslash}p{0.6\textwidth}>{\raggedright\arraybackslash}p{0.2\textwidth}@{}}
    \toprule
    Dataset & Varied per scene & Held fixed \\
    \midrule
    Far-field radiance & Environment map at infinity ($1$--$4$ Gaussian lobes and $1$--$4$ angular boxes, intensities in $[0.1,10]$); piecewise-constant angular $\totCS_t\in[0.01,10]$ and $c\in[0.01,0.99]$; $g\sim\mathrm{Unif}(-0.99,0.99)$ & Unit-ball geometry \\
    3D fluence & $2$--$5$ axis-aligned boxes with $\totCS_t\in[0.01,50]$ and $c\in[0.01,0.99]$ in a near-vacuum background; $1$--$3$ point emitters with intensity in $[1,20]$, half with a Perlin angular profile; $g\sim\mathrm{Unif}(-0.95,0.95)$ & Cube $[-1,1]^3$, vacuum boundary \\
    Sph.~tokamak & All $20$ axes of \autoref{tab:sptok-sampling}: radial build, D-shape shared by geometry and plasma source, H-mode density and temperature profiles, Shafranov shift, radial source offset & Material compositions, tally mesh \\
    EU-DEMO & The $8$ plasma-source axes of \autoref{tab:demo-sampling} & Entire reactor geometry and materials \\
    \bottomrule
\end{tabular*}
\end{table}

\paragraph{Physical realism of the radiative-transfer designs.}
The radiative-transfer designs are broader than any single application; the 3D fluence design carries the high-dynamic-range case on the radiative-transfer side, reproducing the dynamic range of the spherical tokamak.
Extinction, albedo, and phase-function ranges are the physically admissible ranges used in participating-media simulation and rendering; cloud albedo, for example, lies close to $1$, inside our range.
The spatially varying fields combine localized structures, multiple scales, and sharp interfaces rather than one empirical distribution.
\autoref{tab:dynamic-range} shows that a synthetic 3D fluence reference spans about $10$ orders of magnitude, as many as a spherical-tokamak reference and fewer than EU-DEMO (about $17$), while far-field radiance spans only about $3$.
Far-field radiance is therefore not a high-dynamic-range stress test; we use it for the label-noise and allocation studies because it is the one setting where converged references for thousands of training scenes are affordable.

\begin{table}[!htb]
\caption{\textbf{3D fluence and spherical-tokamak references span about $10$ orders of magnitude, EU-DEMO about $17$, and far-field radiance about $3$.} Per scene, the span is $\log_{10}$ of the reference maximum over its smallest positive value; ``above $\epsilon$'' replaces that smallest value by the dataset's $\log_{10}$ floor $\epsilon$ when it is lower, which is the span the $\log_{10}$ metrics see (their data range, Appendix~\ref{app:eval-protocol}). Mean $\pm$ standard deviation over the held-out evaluation scenes, on the field as scored: far-field radiance on the $40{\times}80$ grid, the spherical tokamak per energy group (over the $350$ scene--group pairs), and EU-DEMO only on cells where the reference is measured.}
\label{tab:dynamic-range}
\centering
\footnotesize
\setlength{\tabcolsep}{4pt}
\begin{tabular*}{\textwidth}{@{\extracolsep{\fill}}lccccc@{}}
    \toprule
    Dataset & Scenes & Span & $\epsilon$ & Span above $\epsilon$ & Exact zeros \\
    \midrule
    Far-field radiance & $100$ & $3.16\pm1.27$  & $10^{-6}$  & $3.16\pm1.27$  & $0.0\%$ \\
    3D fluence         & $100$ & $10.48\pm1.69$ & $10^{-10}$ & $8.85\pm0.85$  & $2.3\%$ \\
    Sph.~tokamak       & $50$  & $10.16\pm0.99$ & $10^{-10}$ & $8.21\pm0.88$  & $41.6\%$ \\
    EU-DEMO            & $24$  & $16.97\pm0.26$ & $10^{-17}$ & $10.90\pm0.04$ & $0.0\%$ \\
    \bottomrule
\end{tabular*}
\end{table}

\paragraph{Measured data.}
Dense, full-field measurements are not available for either domain.
No operating fusion power plant provides energy-resolved neutron fields under reactor conditions, and existing detectors give sparse readings with instrument-specific energy responses.
Light-transport datasets provide images or sparse radiometry rather than full transport fields with the geometry, material, and illumination specifications needed for pointwise validation.
Our evaluation is therefore simulation-based, as is standard in shielding and reactor design.

\section{Backbone Comparison and Hyperparameter Search}
\label{app:backbone-hpo}

FNO ranks first under both matched parameter count and matched search budget on the far-field radiance task with four-SPP labels.
All checkpoints are evaluated on the $40{\times}80$ training grid against the same $100$-scene converged reference.

\paragraph{Matched parameter count.}
We resize eight backbone families to a $9.4$--$13.3$ million parameter band and train each family with \ourNO{} $+$ \loss and log MSE for three seeds.
The comparison includes FNO, SFNO~\citep{bonev2023sfno}, U-NO~\citep{rahman2023uno}, CNO~\citep{raonic2023cno}, U-Net~\citep{ronneberger2015unet}, DeepONet~\citep{lu2021deeponet}, Transolver~\citep{wu2024transolver}, and GNOT~\citep{hao2023gnot}.

\begin{table}[!htb]
\caption{\textbf{The \ourNO objective improves every backbone, and FNO ranks first at matched parameter count.} Far-field radiance with four-SPP labels.
Accuracy entries are $\log_{10}$ SSIM on the $100$ held-out scenes at $40{\times}80$, mean $\pm$ sample standard deviation across three seeds, except GNOT with \loss (one completed seed; two diverged).
Latency is measured with the same single-sample inference benchmark for every family.
The gain is the \ourNO{} $+$ \loss SSIM minus the matched log-MSE SSIM.}
\label{tab:backbone-params}
\centering
\scriptsize
\setlength{\tabcolsep}{2.5pt}
\begin{tabular*}{\textwidth}{@{\extracolsep{\fill}}lrrccc@{}}
    \toprule
    Backbone & Params (M) & Latency (ms) & \ourNO{} $+$ \loss & Log MSE & Gain \\
    \midrule
    FNO        & 10.390 &  3.07 & $\mathbf{0.9220\pm0.0026}$ & $0.4733\pm0.0020$ & $+0.4487$ \\
    SFNO       &  9.445 &  4.38 & $0.9076\pm0.0002$ & $0.4691\pm0.0007$ & $+0.4385$ \\
    U-NO       & 12.633 &  3.56 & $0.8498\pm0.0016$ & $0.4525\pm0.0011$ & $+0.3973$ \\
    U-Net      & 13.258 &  1.60 & $0.8328\pm0.0037$ & $0.4338\pm0.0028$ & $+0.3990$ \\
    CNO        & 12.477 &  3.73 & $0.7976\pm0.0027$ & $0.4213\pm0.0022$ & $+0.3763$ \\
    Transolver & 11.206 & 10.54 & $0.4055\pm0.0058$ & $0.2191\pm0.0032$ & $+0.1865$ \\
    GNOT       & 10.940 & 15.37 & $0.3724$ & $0.2538\pm0.0024$ & --- \\
    DeepONet   & 12.064 &  0.49 & $0.2957\pm0.0058$ & $0.1964\pm0.0032$ & $+0.0993$ \\
    \bottomrule
\end{tabular*}
\end{table}

The \ourNO objective improves mean SSIM in all seven families with three completed seeds in both arms: gains are $0.3763$--$0.4487$ for the five grid-based families and $0.0993$--$0.1865$ for DeepONet and Transolver.
GNOT has only one completed \loss seed, so no matched three-seed gain is reported.
The objective therefore transfers across backbones, while attainable accuracy remains backbone-limited.

\paragraph{Transolver and GNOT.}
Transolver and GNOT trail the grid-based operators under the shared training recipe on this task.
Both were built for irregular meshes and general geometries, whereas this task maps fields on a regular $40{\times}80$ angular grid, where spectral and convolutional operators carry the matching inductive bias; we train both from the authors' reference implementations.
Their deficit is the same under log MSE, $0.2191$ and $0.2538$ against $0.4213$--$0.4733$ for the five grid-based families, so it is not caused by \loss.
Transolver has a seed standard deviation of $0.0058$, compared with at most $0.0037$ for the grid-based families.
Two of the three GNOT runs with \loss diverge before the schedule ends; the sole completed seed scores $0.3724$ and cannot establish seed stability.
Neither family entered the matched search of \autoref{tab:backbone-hpo}, so their configurations are untuned.
The comparison therefore ranks backbones for regular-grid transport fields under one training recipe and does not test these two architectures on the irregular meshes they target.

\paragraph{Matched search budget.}
Each of six families receives six $40$-epoch candidates.
We promote the top two candidates per family to $200$ epochs and retrain the selected configurations and evaluate their final checkpoints for three seeds.
Candidate selection and promotion used these same evaluation scenes; an independent selection set was not available.
Table~\ref{tab:backbone-hpo} reports the stronger promoted configuration from each family.

\begin{table}[!htb]
\caption{\textbf{Under a matched search budget, FNO still ranks first.} Far-field radiance with four-SPP labels.
Every family receives the same candidate and promotion budget; the configuration gives the Fourier modes per axis, the channel width, and the number of layers.
Entries are mean $\pm$ sample standard deviation across three seeds after $200$ epochs, on the $100$ held-out scenes at $40{\times}80$.}
\label{tab:backbone-hpo}
\centering
\footnotesize
\setlength{\tabcolsep}{3pt}
\begin{tabular*}{\textwidth}{@{\extracolsep{\fill}}llcc@{}}
    \toprule
    Backbone & Selected configuration & $\log_{10}$ rel.~$L_2$ $\downarrow$ & $\log_{10}$ SSIM $\uparrow$ \\
    \midrule
    FNO      & $10$ modes, width $128$, $6$ layers & $\mathbf{0.0750\pm0.0021}$ & $\mathbf{0.9277\pm0.0042}$ \\
    SFNO     & $16$ modes, width $128$, $6$ layers & $0.1100\pm0.0003$ & $0.9037\pm0.0005$ \\
    U-NO     & $16$ modes, width $32$, $6$ layers  & $0.1233\pm0.0007$ & $0.8482\pm0.0017$ \\
    U-Net    & width $96$, $4$ layers             & $0.1326\pm0.0015$ & $0.8439\pm0.0018$ \\
    CNO      & width $64$, $5$ layers             & $0.1395\pm0.0007$ & $0.8265\pm0.0005$ \\
    DeepONet & branch $256$, trunk $128$, basis $128$    & $0.5526\pm0.0104$ & $0.3229\pm0.0062$ \\
    \bottomrule
\end{tabular*}
\end{table}

The two comparisons produce the same ordering among their six shared families: FNO, SFNO, U-NO, U-Net, CNO, then DeepONet.
Capacity explains little of the objective's advantage: increasing the FNO from $2.77$ million to $10.39$ million parameters raises SSIM from $0.9165\pm0.0040$ to $0.9220\pm0.0026$, a gain of $0.0055$, against the matched FNO loss gain of $0.4487$ in Table~\ref{tab:backbone-params}.

\newpage

\section{Cross-Resolution Evaluation and Multi-Resolution Training}
\label{app:supres}

Single-resolution training supports native FNO inference on nearby meshes.
On EU-DEMO, accuracy holds from $0.5\times$ to $1\times$ the training resolution and degrades moderately at $2\times$; on far-field radiance, a native forward at $2\times$ outperforms low-resolution prediction followed by upsampling.

\paragraph{Far-field radiance.}
We train at $40{\times}80$ and evaluate one earlier single-seed checkpoint, trained on $128$-SPP labels, on $100$ held-out scenes.
At $80{\times}160$, native inference reaches $\log_{10}$~SSIM $0.931$, compared with $0.901$ for block-replicating the $40{\times}80$ prediction.
The native forward also lowers linear rel.~$L_2$ from $0.165$ to $0.139$.
The five-seed checkpoints of \autoref{tab:main-results} reach $\log_{10}$ SSIM $0.8969\pm0.0030$ at $120{\times}240$, three times the training resolution per axis, against $0.9165\pm0.0040$ on the training grid.

\paragraph{EU-DEMO.}
We train at $140{\times}73{\times}146$ and evaluate the five main-table seeds at $0.5\times$, $1\times$, and $2\times$ resolution, scoring only cells where the reference is measured ($42.6\%$, $45.4\%$, and $46.6\%$ of the mesh).
$\log_{10}$~SSIM is $0.7895\pm0.0040$ at $0.5\times$, $0.7923\pm0.0016$ at the training mesh, and $0.7181\pm0.0016$ at $2\times$; the supports differ slightly between grids.
At $5\times$ ($700{\times}365{\times}730$, the reference's native mesh, $49.1\%$ measured) $\log_{10}$ rel.~$L_2$ stays at $0.0844\pm0.0004$, while $\log_{10}$ SSIM falls to $0.5441\pm0.0006$: its $11$-cell window spans one fifth of the physical extent per axis that it spans at the training mesh, so the two SSIM values measure structure at different physical scales.

\begin{table}[!htb]
\caption{\textbf{Without retraining, accuracy holds on nearby meshes and degrades with larger resolution steps.} Cross-resolution evaluation.
The far-field block compares a native forward with block upsampling for one earlier single-seed checkpoint trained on $128$-SPP labels, not comparable to \autoref{tab:main-results}.
EU-DEMO entries are mean $\pm$ sample standard deviation over the five main-table seeds on the $24$ held-out configurations, scored only on cells where the reference is measured.}
\label{tab:demo}
\centering
\footnotesize
\setlength{\tabcolsep}{3pt}
\begin{tabular*}{\textwidth}{@{\extracolsep{\fill}}lccc@{}}
    \toprule
    Mesh and inference path & rel.~$L_2$ $\downarrow$ & $\log_{10}$ rel.~$L_2$ $\downarrow$ & $\log_{10}$ SSIM $\uparrow$ \\
    \midrule
    \multicolumn{4}{@{}l}{\textbf{Far-field radiance}} \\ %
    $40{\times}80$ (training mesh)                    & $\mathbf{0.120}$ & $\mathbf{0.056}$ & $\mathbf{0.952}$ \\
    $80{\times}160$ (native forward)                  & $0.139$          & $0.069$          & $0.931$ \\
    $80{\times}160$ ($40{\times}80$ then upsample)   & $0.165$          & $0.094$          & $0.901$ \\
    \midrule
    \multicolumn{4}{@{}l}{\textbf{EU-DEMO}} \\ %
    $70{\times}37{\times}73$ ($0.5\times$ native)      & $0.1203\pm0.0088$ & $0.0854\pm0.0007$ & $0.7895\pm0.0040$ \\
    $140{\times}73{\times}146$ (training mesh)         & $\mathbf{0.1085\pm0.0068}$ & $0.0839\pm0.0006$ & $\mathbf{0.7923\pm0.0016}$ \\
    $280{\times}146{\times}292$ ($2\times$ native)    & $0.1216\pm0.0057$ & $\mathbf{0.0836\pm0.0005}$ & $0.7181\pm0.0016$ \\
    $700{\times}365{\times}730$ ($5\times$ native)   & $0.1259\pm0.0058$ & $0.0844\pm0.0004$ & $0.5441\pm0.0006$ \\ %
    \bottomrule
\end{tabular*}
\end{table}

\paragraph{Mixed-resolution training at a tenfold grid step.}
Adding $100$ high-resolution training scenes reduces the active-voxel log MAE (the mean absolute $\log_{10}$ error over voxels whose reference flux is above the floor) of a direct forward pass on the $700{\times}550{\times}800$ native mesh from $5.36$ to $2.36$.
Linear rel.~$L_2$ remains near one, so the added data reduces the severity of the native-scale collapse without establishing tenfold resolution invariance.

\begin{table}[!htb]
\caption{\textbf{Adding a few high-resolution scenes reduces, but does not remove, the collapse of a direct forward pass at a tenfold grid step.} Mixed-resolution diagnostic on an earlier EU-DEMO setup, one run per row.
The baseline uses $50{,}000$ scenes at $70{\times}55{\times}80$; the mixed run adds $100$ scenes at $350{\times}275{\times}400$.
Native-direct metrics use three paired scenes, and low-resolution prediction followed by upsampling uses ten.}
\label{tab:demo-multires-training}
\centering
\footnotesize
\setlength{\tabcolsep}{3pt}
\begin{tabular*}{\textwidth}{@{\extracolsep{\fill}}lcccc@{}}
    \toprule
    Training corpus & Train-grid rel.~$L_2$ & Native-direct rel.~$L_2$ & Active log MAE & Lowres-up rel.~$L_2$ \\
    \midrule
    Low mesh only       & $0.0173$ & $1.0000$ & $5.36$ & $0.1627$ \\ %
    Low $+$ high meshes & $0.0207$ & $0.9998$ & $2.36$ & $0.1617$ \\ %
    \bottomrule
\end{tabular*}
\end{table}

The mixed run reached epoch $16$ of $20$ under a six-hour limit.
Its low-resolution accuracy matches the single-resolution baseline, and its direct native-scale prediction remains outside the training amplitude range.
Nearby-mesh inference is reliable; a tenfold grid step requires an explicitly resolution-consistent representation.

\section{Source Superposition, Tetrahedral Transport, and Label Validation}
\label{app:tetra-status}

\subsection{Source Superposition and Geometry Generalization}

Fixed-geometry transport is linear in the source, so a nonlinear surrogate must be compared with source-superposition baselines.
On EU-DEMO, a nonnegative rank-$64$ Green operator reaches log relative $L_2$ $0.0703$ and log SSIM $0.7958$, against $0.0840$ and $0.7920$ for the main-table PTNO recipe on the same three seeds.
Both use final checkpoints, the same code, H200 hardware, labels, loss, and $50{,}000$ updates; their source representations differ (a voxelized field versus eight parameters).
The Green operator's linear relative $L_2$ is worse, $0.2529$ against $0.1085$, and its linear PSNR is $28.27$ versus $33.83$\,dB.
Thus fixed-geometry DEMO does not establish that a nonlinear operator outperforms a source-linear model in every metric.

A separate spherical-tokamak experiment trains from scratch on $51{,}200$ configurations ($1{,}600$ geometries, $32$ sources each, $10^5$ histories per label).
It evaluates $50$ unseen geometries against $10^9$-history references and $16$ held-out sources on training geometries against $4\times10^8$-history references.
With two seeds and final checkpoints, PTNO reaches log relative $L_2$ $0.0474$ on unseen geometries, compared with $0.3160$ for the strongest tested geometry-interpolation baseline, and $0.0570$ on held-out sources, compared with $0.2974$ for same-geometry nonnegative least-squares superposition.
These are $6.7\times$ and $5.2\times$ lower errors, respectively, under the stated noisy-label budget.
The superposition baseline combines only a few noisy training labels, whereas the learned operator pools information across the corpus.
The advantage is consistent with cross-sample denoising under this label budget; it does not establish nonlinear source response within a fixed geometry.
On seen geometries, superposition has higher linear PSNR ($33.98$ versus $29.93$\,dB).

\begin{table}[!htb]
\caption{\textbf{The spherical-tokamak surrogate improves log-space accuracy over noisy-label superposition.} Six volume-weighted metrics, averaged over energy groups and scenes; PTNO is the mean of two final-checkpoint seeds. G denotes unseen geometries and S held-out sources on training geometries. This $51{,}200$-scene corpus is separate from the historical main-table experiment.}
\label{tab:sptok-superposition}
\centering
\scriptsize
\setlength{\tabcolsep}{3pt}
\begin{tabular}{llrrrrrr}
\toprule
Split & Method & Log rel.~$L_2$ & Log SSIM & Log PSNR & Rel.~$L_2$ & SSIM & PSNR \\
\midrule
G & PTNO & $.0474$ & $.8550$ & $29.57$ & $.2202$ & $.9740$ & $30.68$ \\
G & Interpolation & $.3160$ & $.4312$ & $12.10$ & $.3908$ & $.9275$ & $26.34$ \\
S & PTNO & $.0570$ & $.8429$ & $28.32$ & $.2581$ & $.9680$ & $29.93$ \\
S & NNLS superposition & $.2974$ & $.4752$ & $13.20$ & $.3324$ & $.9616$ & $33.98$ \\
\bottomrule
\end{tabular}
\end{table}

Source-superposition augmentation is neutral within the two-seed spread on this corpus and was removed from the spherical-tokamak recipe.
A geometry-conditioned rank-$64$ Green head, which remains linear in the source, improves log relative $L_2$ by approximately $2\%$ on unseen geometries and $5\%$ on held-out sources against its matched L40S control, but increases linear relative $L_2$ from $0.21$--$0.25$ to $0.62$--$0.68$.
It remains an experimental alternative and does not replace the main recipe.

\begin{table}[!htb]
\caption{\textbf{A learned source-linear Green operator improves log-space accuracy at a cost in linear-space error.} Final-checkpoint means; three seeds for DEMO on H200, two for each spherical-tokamak split on L40S. Each Green row is compared only with its paired control. DEMO uses measured support and floor $10^{-17}$; spherical tokamak uses floor $10^{-10}$.}
\label{tab:green-six-metrics}
\centering
\scriptsize
\setlength{\tabcolsep}{3pt}
\begin{tabular}{llrrrrrr}
\toprule
Task & Model & Log rel.~$L_2$ & Log SSIM & Log PSNR & Rel.~$L_2$ & SSIM & PSNR \\
\midrule
DEMO & PTNO & $.0840$ & $.7920$ & $19.78$ & $.1085$ & $.9834$ & $33.83$ \\
DEMO & Green & $.0703$ & $.7958$ & $21.32$ & $.2529$ & $.9900$ & $28.27$ \\
Sptok G & PTNO & $.0473$ & $.8551$ & $29.60$ & $.2100$ & $.9742$ & $30.89$ \\
Sptok G & Green & $.0463$ & $.8600$ & $29.94$ & $.6212$ & $.9642$ & $25.27$ \\
Sptok S & PTNO & $.0570$ & $.8428$ & $28.35$ & $.2454$ & $.9684$ & $30.15$ \\
Sptok S & Green & $.0542$ & $.8491$ & $29.14$ & $.6832$ & $.9563$ & $24.87$ \\
\bottomrule
\end{tabular}
\end{table}

\subsection{EU-DEMO on an Unstructured Mesh}

A tetrahedral extension predicts cell-average neutron flux on a fixed EU-DEMO mesh with $42{,}729$ cells.
OpenMC scores track-length flux directly on an unstructured MOAB mesh; dividing the cell-integrated tally by cell volume gives the training target.
The model is a GINO with volume-integral encoders and decoders around a $24{\times}8{\times}32$ latent FNO, width $16$, four layers, and modes $(8,4,8)$.
Its $13$ inputs are one source-density field transformed by $\operatorname{asinh}(q/10^{-7})$, eight source parameters, and the three cell coordinates plus cell volume in fixed length units.
Training uses a softplus head of scale $10^{-8}$ and \loss with $\eta=10^{-17}$, batch size $2$, learning rate $10^{-3}$, and a fixed $20$-epoch, $10^5$-update cosine schedule.
Every reported tetrahedral model is its final checkpoint; no checkpoint is selected on a validation or reference set.

Two training-label corpora each contain $10^4$ source configurations: analog transport uses $5\times10^4$ histories per source, whereas weight-window transport uses $2\times10^5$.
The reference contains $128$ held-out sources, each pooling $16$ independent runs of $3.6\times10^7$ histories ($5.76\times10^8$ total).
The weight-window-trained model has substantially higher log-space accuracy (Table~\ref{tab:tetra-last}), but the fourfold history difference means that this comparison does not isolate variance reduction at equal simulation budget.
All six metrics use cell-volume weights; SSIM uses a physical neighborhood scale of $100$\,cm and the reference's own data range.

\begin{table}[!htb]
\caption{\textbf{Weight-window labels improve the tetrahedral surrogate under the recorded, unequal history budgets.} Per-seed values for seeds $42$/$43$, averaged over $128$ held-out sources; final checkpoints. Log-space metrics use floor $10^{-20}$.}
\label{tab:tetra-last}
\centering
\small
\begin{tabular}{lcc}
\toprule
Metric & Weight-window labels & Analog labels \\
\midrule
$\log_{10}$ rel.~$L_2$ $\downarrow$ & $0.0400/0.0396$ & $0.2976/0.2982$ \\
$\log_{10}$ SSIM $\uparrow$ & $0.7750/0.7780$ & $0.4408/0.4416$ \\
$\log_{10}$ PSNR (dB) $\uparrow$ & $26.93/27.01$ & $9.24/9.22$ \\
Linear rel.~$L_2$ $\downarrow$ & $0.2620/0.2582$ & $0.3075/0.2872$ \\
Linear SSIM $\uparrow$ & $0.9890/0.9892$ & $0.9881/0.9889$ \\
Linear PSNR (dB) $\uparrow$ & $30.26/30.41$ & $28.90/29.57$ \\
\bottomrule
\end{tabular}
\end{table}

At the structured DEMO floor of $10^{-17}$, the same weight-window checkpoints reach log-space SSIM $0.8683/0.8711$ and relative $L_2$ $0.0230/0.0228$.
The floor materially changes the apparent accuracy, so results from different floors must not be compared as if they used one evaluation protocol.

\subsection{FNG Shutdown-Dose Benchmark}
\label{app:fng-label-check}

An independent experimental check tests the neutron-transport, activation, and delayed-photon stages of our shutdown-dose label generator.
We reproduce the first FNG shutdown-dose campaign using the openly archived geometry, source, irradiation history, and measurements of \citet{peterson2024r2sdata}.
This is a check of the Monte Carlo label pipeline, not a measurement-based evaluation of a trained neural operator.
The calculation uses OpenMC's cell-based rigorous two-step workflow with $65$ activation cells, TENDL-2019 neutron data, an activation chain based on TENDL-2019/ENDF/B-VIII.0, and ENDF/B-VIII.0 photon data.
Each neutron calculation and each cooling-time photon calculation uses $2\times10^7$ histories.

\begin{table}[!htb]
\caption{\textbf{The label pipeline reproduces the measured FNG dose rate within 20\% at the five reference cooling times.} $C/E$ is calculated over measured dose rate. Uncertainties shown are photon Monte Carlo standard errors conditional on the sampled neutron reaction rates; they exclude neutron-sampling and measurement uncertainty.}
\label{tab:fng-ce}
\centering
\small
\begin{tabular}{rcc}
\toprule
Cooling time (days) & Measured dose rate ($\mu$Sv/h) & $C/E$ \\
\midrule
$1$ & $2.46$ & $0.850\pm0.021$ \\
$7$ & $0.699$ & $1.061\pm0.026$ \\
$15$ & $0.495$ & $1.185\pm0.029$ \\
$30$ & $0.416$ & $1.180\pm0.029$ \\
$60$ & $0.316$ & $1.200\pm0.031$ \\
\bottomrule
\end{tabular}
\end{table}

The six nickel-foil $^{58}$Ni$(n,p)$ reaction-rate ratios range from $0.97$ to $1.12$; the $(n,2n)$ ratios are lower, $0.76$--$0.93$.
A preregistered requirement to reproduce the archived calculation within $5\%$ fails at $7$--$30$ days, with differences of $14$--$16\%$.
Replaying the archived nuclide inventories through our photon stage reproduces its dose rates within $2.3\%$; the remaining discrepancy is localized to short-lived activation products introduced by numerical roundoff in the archived inventory, which our reachable-subspace decay solve excludes.
We report this deviation rather than counting the original replication threshold as met.
The benchmark uses cell-based activation and an earlier nuclear-data registry, whereas the current production corpus uses mesh-based activation and updated steel compositions.
It therefore does not establish the accuracy of the production mesh discretization or of a neural operator on the FNG geometry.

\section{Radiative Transfer with Refractive and Reflective Interfaces}
\label{app:refract}

A fifth dataset adds glass, mirror, and matte objects to the 3D fluence task of Appendix~\ref{subsec:data-spatial}.
Light that hits a glass object is bent into the geometric shadow behind it and partly reflected, a mirror folds the light field back on itself, and a matte object casts a soft shadow, so the target field is no longer a smooth function of the medium alone (\autoref{fig:setup-refract}).
The \ourNO recipe still learns this task from labels traced with $64$ Monte Carlo samples per voxel (SPP).
We score a model by the SSIM between the $\log_{10}$ of its predicted 3D fluence volume and the $\log_{10}$ of a high-SPP Monte Carlo reference, computed on the volume itself, not on rendered images.
On a held-out test set of $224$ scenes, scored only once after every design choice had been fixed, the final model reaches a $\log_{10}$~SSIM of $0.961$.
For scale, two independent Monte Carlo solutions of a different $224$-scene set from the same generator score $0.997$ against each other; this gives an empirical scale for reference noise, rather than a strict upper bound on model accuracy.

\begin{figure}[!htb]
    \centering
    \includegraphics[width=\linewidth]{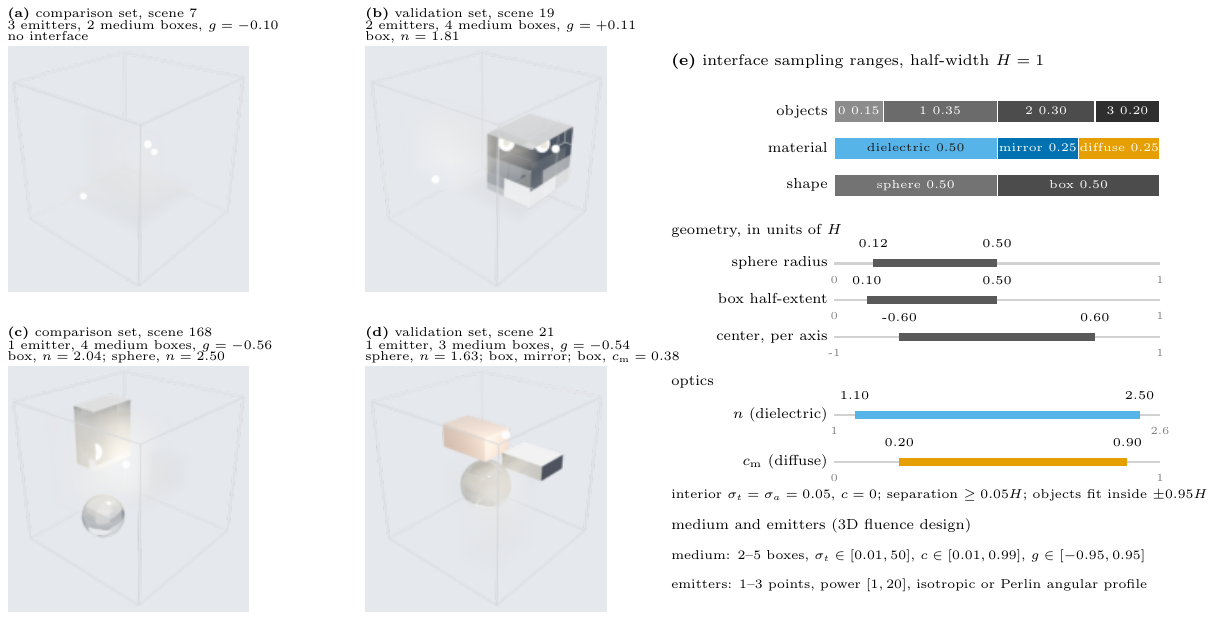}
    \input{fig/fig_setup_refract/caption.tex}
\end{figure}

\subsection{Scenes, Labels, and Reference}
\label{app:refract-scenes}

Each scene is a 3D fluence scene (background $\totCS_t$ and $c$ with $2$--$5$ contrast solids, $1$--$3$ point emitters, one Henyey--Greenstein $g$) plus zero to three solid objects: spheres or axis-aligned boxes made of a dielectric with refractive index $n\in[1.1,2.5]$, an ideal mirror, or a Lambertian (matte) surface with albedo $c_{\mathrm{m}}\in[0.2,0.9]$ (\autoref{tab:refract-sampling}).
Objects are opaque to the medium: the participating medium exists only outside them, and their interiors are a weak absorber (absorption coefficient $\totCS_a=0.05$, no scattering), because a lossless glass box traps most interior directions and its interior fluence has no converged value.
At an interface a photon is reflected or transmitted with the polarized Fresnel probability and refracted by Snell's law, including total internal reflection; mirrors reflect with unit probability, and matte surfaces absorb a fraction $1-c_{\mathrm{m}}$ and re-emit the rest in a cosine-weighted direction.

\begin{table}[!htb]
\caption{\textbf{Each interface scene adds up to three dielectric, mirror, or matte spheres or boxes.} The medium, emitters, and $g$ follow the 3D fluence design of \autoref{tab:id-boundary}; $H=1$ is the half-width of the cube. Realized values are counted over the $10^5$ training scenes; the shape and center rows give the proposal. $n$ spans water ($1.33$) to diamond ($2.42$).}
\label{tab:refract-sampling}
\centering
\scriptsize
\setlength{\tabcolsep}{4pt}
\begin{tabular}{@{}>{\raggedright\arraybackslash}p{0.18\textwidth}>{\raggedright\arraybackslash}p{0.11\textwidth}>{\raggedright\arraybackslash}p{0.30\textwidth}>{\raggedright\arraybackslash}p{0.32\textwidth}@{}}
    \toprule
    Quantity & Distribution & Range & Note \\
    \midrule
    Number of objects & categorical & $P(0,1,2,3)=(0.15,0.35,0.30,0.20)$ & realized $(0.152,0.351,0.298,0.199)$ \\
    Shape & uniform & sphere $0.5$ / box $0.5$ & box faces are axis-aligned \\
    Sphere radius & uniform & $[0.12,0.50]\,H$ & \\
    Box half-extent (per axis) & uniform & $[0.10,0.50]\,H$ & \\
    Center (per axis) & uniform & $[-0.6,0.6]\,H$ & must fit inside $\pm0.95H$ \\
    Material & categorical & dielectric $0.50$ / mirror $0.25$ / matte $0.25$ & realized $0.499/0.250/0.251$ of $154{,}412$ objects \\
    Refractive index $n$ & uniform & $[1.1,2.5]$ & dielectric only; each $0.1$-wide bin holds $6.9$--$7.3\%$ of draws \\
    Matte albedo $c_{\mathrm{m}}$ & uniform & $[0.2,0.9]$ & matte only (mirror reflectivity is $1$); each $0.05$-wide bin holds $6.9$--$7.4\%$ of draws \\
    Interior & fixed & $\totCS_t=\totCS_a=0.05$, $c=0$ & no medium inside objects \\
    Separation & constraint & $\ge 0.05H$ object--object and object--emitter & \\
    \bottomrule
\end{tabular}
\end{table}

Labels come from the Monte Carlo collision estimator of Appendix~\ref{subsec:data-spatial} extended with analytic ray--sphere and ray--box intersections; the target is the per-voxel track-length fluence (the summed path length of photons inside a voxel, an unbiased estimate of fluence) on a $64^3$ grid, divided by voxel volume and by total emitted power.
The estimator was checked against Geant4~\citep{geant4,geant4_2016} optical photons on the same analytic scenes: the two codes share neither the estimator nor the interface code, their difference falls as $N^{-0.45}$ over $2.5\times10^5$--$2\times10^6$ Geant4 photons with no plateau, and seven physics components isolated one at a time agree to $0.04$--$0.21\%$.
Training uses the first $10^5$ scenes of a $10^6$-scene training set, each traced once at $64$~SPP, that is, $64$ photon paths per voxel of the $64^3$ grid.
Four held-out sets are used, all drawn from the same scene generator and disjoint from training and from each other:
a $50$-scene set traced to $786{,}432$~SPP for monitoring during training;
a $224$-scene \emph{stratified comparison set} ($7$ interface compositions $\times$ $32$ scenes: no object, one matte, one mirror, one dielectric with $n<1.5$, one with $n\ge2.0$, two dielectrics, three dielectrics) traced to $32{,}768$~SPP, on which we rank the variants;
and two further $224$-scene sets with the same compositions and SPP but fresh random seeds, a \emph{selection set} that scores each candidate once and is used only to choose among the final candidates, and a \emph{test set} that scores each model once, after that choice.
The label ceiling is one independent Monte Carlo solution of the stratified comparison set scored against another with the same metrics: $\log_{10}$~SSIM $0.9968$ and $\log_{10}$ rel.~$L_2$ $0.0113$.
The final model remains about $0.03$ below this empirical ceiling; this gap is larger than the reference-noise scale, although it does not certify every small difference between models.

\subsection{Inputs, Model, and Protocol}
\label{app:refract-protocol}

The operator sees $23$ input channels, all computed from the scene parameters at the target resolution.
The first $21$ channels comprise one emission grid, nine degree-$0$--$2$ source spherical-harmonic fields, extinction and albedo, four surface fields (refractive index, mirror mask, matte albedo, and signed distance), four polynomial embeddings of $g$, and one asinh-encoded occluded direct-flux field.
The remaining two tell the operator where the interfaces send light before any scattering: $D_0$, the direct flux with every interface treated as opaque, and $R_{K6}$, the extra energy delivered by up to six reflections or refractions; both are computed by a cheap deterministic ray tracer, encoded on a fixed $\log_{10}$ scale, and also fed directly to the readout layer.
Writing $d=\log_{10}\max(D_0,10^{-8})$, the final two channels are $(d+4)/4$ and $[\log_{10}\max(D_0+R_{K6},10^{-8})-d]/4$, without fitted normalization.
These channels are inputs computed from the scene, not labels: they contain no volume scattering or diffuse-surface bounce.
The backbone is a 3D FNO (modes $32^3$, $8$ layers, width $64$ for the final model) with a softplus head and \loss with $\eta=2\times10^{-6}$ on the detached prediction.
Every variant follows one protocol: the first $10^5$ scenes seen once, $25{,}000$ AdamW updates at batch $4$ with cosine decay from $3\times10^{-3}$ to $10^{-7}$, weight decay $10^{-3}$, gradient clipping at $1$, seeds $42$ and $43$, and the checkpoint at the end of training with no validation-based selection.
The longer runs repeat the same $10^5$ scenes two or four times with the schedule stretched accordingly, so they are compared only with each other.
Metrics follow \autoref{app:eval-protocol}, with floor $\epsilon=3.28\times10^{-7}$ in the $\log_{10}$ metrics.
Linear relative $L_2$ is not reported for this dataset: for a point source the cell average in the voxel that contains the emitter scales as $h^{-2}$ with the voxel width $h$ and has no continuum limit, and that single voxel carries $98$--$100\%$ of every model's squared error.
We therefore rank in $\log_{10}$ space and use the peak ratio, the predicted maximum over the reference maximum in each scene, to check that the source core is reproduced.

\subsection{Results}
\label{app:refract-results}

\ourNO learns this task from $64$-SPP labels to within $0.035$ of the label ceiling in $\log_{10}$~SSIM (\autoref{tab:refract-main}).
The final model (width $64$, $8$ layers, $10^5$ updates) scores $\log_{10}$~SSIM $0.9647$/$0.9652$ on the stratified comparison set for seeds $42$/$43$.
We chose its seed-$43$ checkpoint on the selection set; on the test set, scored once after that choice, it reaches $\log_{10}$~SSIM $0.9611$ (median over scenes $0.9762$, 5th percentile $0.8760$), $\log_{10}$ rel.~$L_2$ $0.0720$, and $\log_{10}$~PSNR $38.1$~dB.
Across the seven interface compositions of the stratified comparison set, $\log_{10}$~SSIM ranges from $0.954$/$0.955$ (two dielectrics) to $0.974$/$0.974$ (no object), and the worst single scene in every composition stays above $0.82$.
The source core is reproduced: the median over scenes of the peak ratio is $1.00$/$1.04$.
The remaining error sits in the dim regions behind glass objects, where the model over-predicts by $0.05$--$0.07$ orders of magnitude at the median.

Training in the physical space matters here as on the other four datasets.
With everything else fixed (same inputs, corpus, schedule, and seeds, at the $25{,}000$-update budget), replacing \ourNO's softplus head and \loss with a $\log_{10}$ output trained by MSE lowers $\log_{10}$~SSIM from $0.9503$/$0.9490$ to $0.9391$/$0.9364$.
\autoref{tab:refract-loss-lineup} repeats this comparison at the final model's budget of $10^5$ updates, again changing only the output head and the loss.
The $\log_{10}$-output MSE model scores $0.018$/$0.019$ below \ourNO in $\log_{10}$~SSIM on the stratified comparison set and $0.017$/$0.018$ below on the test set, and an identity head trained with linear $L_2$ collapses to a near-zero field.
In the linear source core the $\log_{10}$-output model is closer to the reference than \ourNO: its mean peak ratio is $0.985$/$1.016$ against $1.799$/$1.627$, and its core error is lower.
These means keep every scene, and the \ourNO seed-$42$ core error of $1.155$ is carried by one three-dielectric scene (median over scenes $0.068$); the two seeds do not settle whether \ourNO systematically overshoots the peak.
Neither component works alone at this budget.
With \loss but an identity head the model collapses to a near-zero field like the linear-$L_2$ model, and with a softplus head but a per-sample relative $L_2$ loss it outputs, in every scene, a constant at the head's lower bound.

\begin{table}[!htb]
\caption{\textbf{\ourNO reaches within $0.035$ of the label ceiling on the interface dataset.} $\log_{10}$ metrics, mean over the $224$ scenes of the stratified comparison set, for seeds $42$/$43$ with end-of-training checkpoints. The selection-set column scores each model once on a separate $224$-scene held-out set (the $25{,}000$-update rows on an earlier, identically generated selection set). Peak ratio is the median over scenes of the predicted over the reference maximum. The label ceiling scores one independent Monte Carlo solution of the stratified comparison set against another. The last two rows share one input construction, corpus, and schedule at $25{,}000$ updates and differ only in the output head and loss.}
\label{tab:refract-main}
\centering
\footnotesize
\setlength{\tabcolsep}{1.5pt}
\begin{tabular*}{\textwidth}{@{\extracolsep{\fill}}>{\raggedright\arraybackslash}p{0.29\textwidth}cccc@{}}
    \toprule
    Model & $\log_{10}$~SSIM $\uparrow$ & $\log_{10}$~rel.~$L_2\downarrow$ & Selection-set $\log_{10}$~SSIM $\uparrow$ & Peak ratio \\
    \midrule
    \ourNO, final ($10^5$ updates, width $64$) & $\mathbf{0.9647}$ / $\mathbf{0.9652}$ & $\mathbf{0.0699}$ / $\mathbf{0.0682}$ & $0.9613$ / $0.9624$ & $1.00$ / $1.04$ \\
    Label ceiling (second MC solution) & $0.9968$ & $0.0113$ & & $1.00$ \\
    \midrule
    \ourNO (softplus + \loss), $25{,}000$ updates & $0.9503$ / $0.9490$ & $0.0928$ / $0.0942$ & $0.9479$ / $0.9471$ & $1.47$ / $1.36$ \\
    NO, $\log_{10}$ output + MSE, $25{,}000$ updates & $0.9391$ / $0.9364$ & & & \\
    \bottomrule
\end{tabular*}
\end{table}

\begin{table}[!htb]
\caption{\textbf{At equal budget, the softplus head with \loss outperforms plain losses on the interface dataset.} Every row uses the final model's construction, corpus, schedule, and $10^5$ updates and changes only the output head and loss; seeds $42$/$43$, end-of-training checkpoints, means over $224$ scenes. The \ourNO row rescores the final model of \autoref{tab:refract-main} with the evaluator used for the other rows (differences below $10^{-4}$). Core rel.~$L_2$ is the linear relative $L_2$ with each emitter voxel and its $26$ neighbors removed; peak ratio is the mean over scenes of the predicted over the reference maximum.}
\label{tab:refract-loss-lineup}
\centering
\scriptsize
\setlength{\tabcolsep}{1pt}
\begin{tabular*}{\textwidth}{@{\extracolsep{\fill}}>{\raggedright\arraybackslash}p{0.2\textwidth}cccccc@{}}
\toprule
& \multicolumn{2}{c}{Stratified comparison set} & \multicolumn{2}{c}{Test set} & \multicolumn{2}{c}{Linear, comparison set} \\
\cmidrule(lr){2-3}\cmidrule(lr){4-5}\cmidrule(l){6-7}
Output head + loss & $\log_{10}$~SSIM $\uparrow$ & $\log_{10}$~rel.~$L_2\downarrow$ & $\log_{10}$~SSIM $\uparrow$ & $\log_{10}$~rel.~$L_2\downarrow$ & Core rel.~$L_2\downarrow$ & Peak ratio \\
\midrule
\ourNO{} (softplus + \loss) & $\mathbf{0.9647}$ / $\mathbf{0.9652}$ & $\mathbf{0.0699}$ / $\mathbf{0.0682}$ & $\mathbf{0.9598}$ / $\mathbf{0.9611}$ & $\mathbf{0.0735}$ / $\mathbf{0.0720}$ & $1.155$ / $0.086$ & $1.799$ / $1.627$ \\  %
NO, $\log_{10}$ output + MSE & $0.9467$ / $0.9462$ & $0.1289$ / $0.1289$ & $0.9430$ / $0.9430$ & $0.1286$ / $0.1287$ & $0.078$ / $0.084$ & $0.985$ / $1.016$ \\  %
NO, identity head + linear $L_2$ & $0.2251$ / $0.2354$ & $0.8929$ / $0.8957$ & $0.2101$ / $0.2198$ & $0.9458$ / $0.9502$ & $0.964$ / $0.962$ & $0.001$ / $0.001$ \\  %
Identity head + \loss & $0.2919$ / $0.2726$ & $0.7056$ / $0.7952$ & $0.2653$ / $0.2454$ & $0.7203$ / $0.8271$ & $0.950$ / $0.950$ & $0.001$ / $0.001$ \\  %
Softplus + per-sample rel.~$L_2$ & $0.1620$ / $0.1620$ & $3.6552$ / $3.6552$ & $0.1658$ / $0.1658$ & $3.7943$ / $3.7943$ & $1.000$ / $1.000$ & $0.000$ / $0.000$ \\  %
\bottomrule
\end{tabular*}
\end{table}

\subsection{Limitations}
\label{app:refract-limits}

All numbers in this section are two seeds of one from-scratch recipe with end-of-training checkpoints, not the five seeds of \autoref{tab:main-results}.
The loss comparison of \autoref{tab:refract-main} was run at the $25{,}000$-update budget; \autoref{tab:refract-loss-lineup} repeats it at $10^5$ updates.
The training corpus is fixed at $10^5$ scenes because the two interface input channels are precomputed and stored for each training scene, so the dim-region bias reported here is the current state, not a floor.
Linear relative $L_2$ and cross-resolution evaluation are not reported, because the cell average in the voxel that contains a point emitter scales as $h^{-2}$ and has no continuum limit.
The estimator and Geant4 share the same simplification of total internal reflection, so their agreement does not test that approximation.

\end{document}